\documentclass[acmtog]{acmart}

\acmSubmissionID{papers\_1147}

\usepackage[ruled,linesnumbered]{algorithm2e} 

\usepackage{multirow} 
\usepackage{subfigure}
\usepackage{wrapfig}

\setcopyright{acmlicensed}

\begin{document}


\title{Learning-based Seam Correspondence Reconstruction in Sewing Patterns}


\author{Zhendong Wang}
\authornote{Corresponding author}
\email{wang.zhendong.619@gmail.com}
\affiliation{%
	\institution{Style3D Research}
	\city{Hangzhou}
	\country{China}}

\author{Jintong Wang}
\authornote{Work done during an internship at Style3D Research.}
\affiliation{%
	\institution{Zhejiang Sci-Tech University}
	\city{Hangzhou}
	\country{China}}
\affiliation{%
	\institution{Style3D Research}
	\city{Hangzhou}
	\country{China}}
\email{2023220603066@mails.zstu.edu.cn}

\author{Chen Liu}
\affiliation{%
	\institution{Style3D Research}
	\city{Hangzhou}
	\country{China}}
\email{eric.liu@linctex.com}

\author{Yao Jin}
\affiliation{%
	\institution{Zhejiang Sci-Tech University}
	\city{Hangzhou}
	\country{China}}
\email{jinyao@zstu.edu.cn}

\author{Ligang Liu}
\affiliation{%
	\institution{University of Science and Technology of China}
	\city{Hefei}
	\country{China}}
\email{lgliu@ustc.edu.cn}

\author{Huamin Wang}
\affiliation{%
	\institution{Style3D Research}
	\city{Hangzhou}
	\country{China}}
\email{wanghmin@gmail.com}


\begin{abstract}
    Digital sewing patterns typically consist of disjoint 2D panels without explicit stitch annotations, making downstream 3D modeling reliant on labor-intensive expert specification. In this paper, we present a graph-based learning framework that reconstructs two-level stitching information—coarse panel connectivity and fine-grained seam correspondence—from 2D panel geometry alone. At the coarse level, panel connectivity is inferred by predicting panel semantics associated with anatomical body regions, enforcing consistency with body structure and garment design conventions. 
    Based on the reconstructed panel graph, fine-grained seam correspondences between panel pairs are inferred by learning latent edge representations that jointly encode local seam geometry and global garment context through graph message passing. The resulting edge embeddings are subsequently decoded into detailed seam correspondences.
    Our method supports complex sewing-pattern topologies, including many-to-one correspondences, intra-panel seams, and curved seams. Experiments demonstrate high stitching accuracy and strong generalization across garment styles.
\end{abstract}



\begin{teaserfigure}
    \centering
    \subfigure[Dress pattern]{\includegraphics[height=0.205\linewidth]{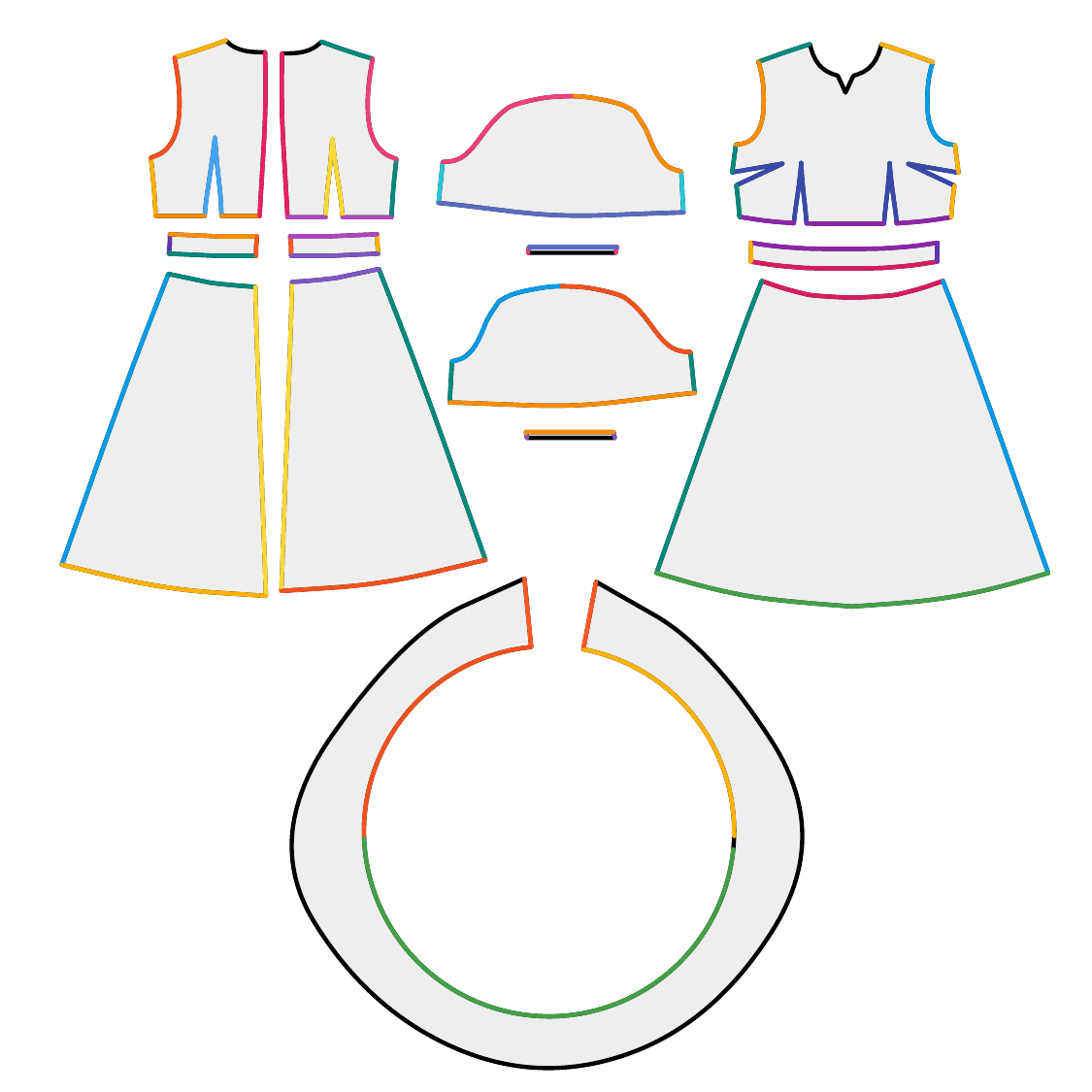}}
    \subfigure[Dress]{\includegraphics[height=0.205\linewidth]{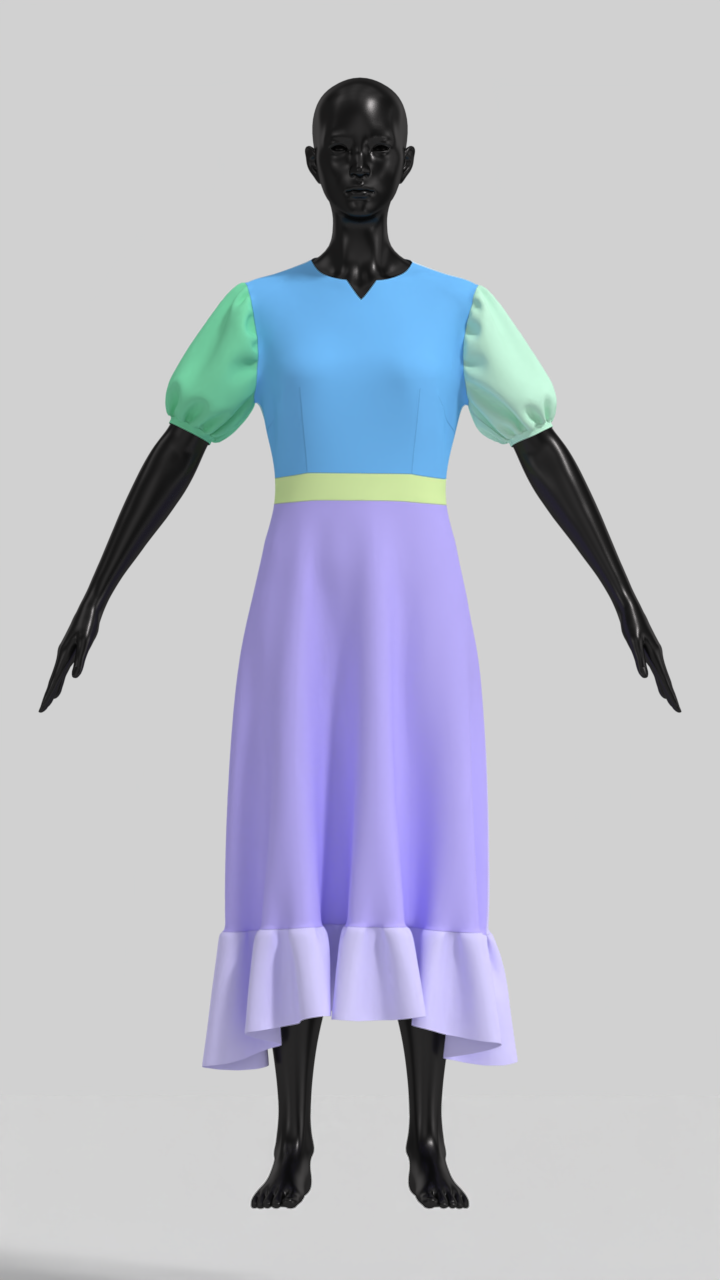}}
    \subfigure[Top pattern]{\includegraphics[height=0.205\linewidth]{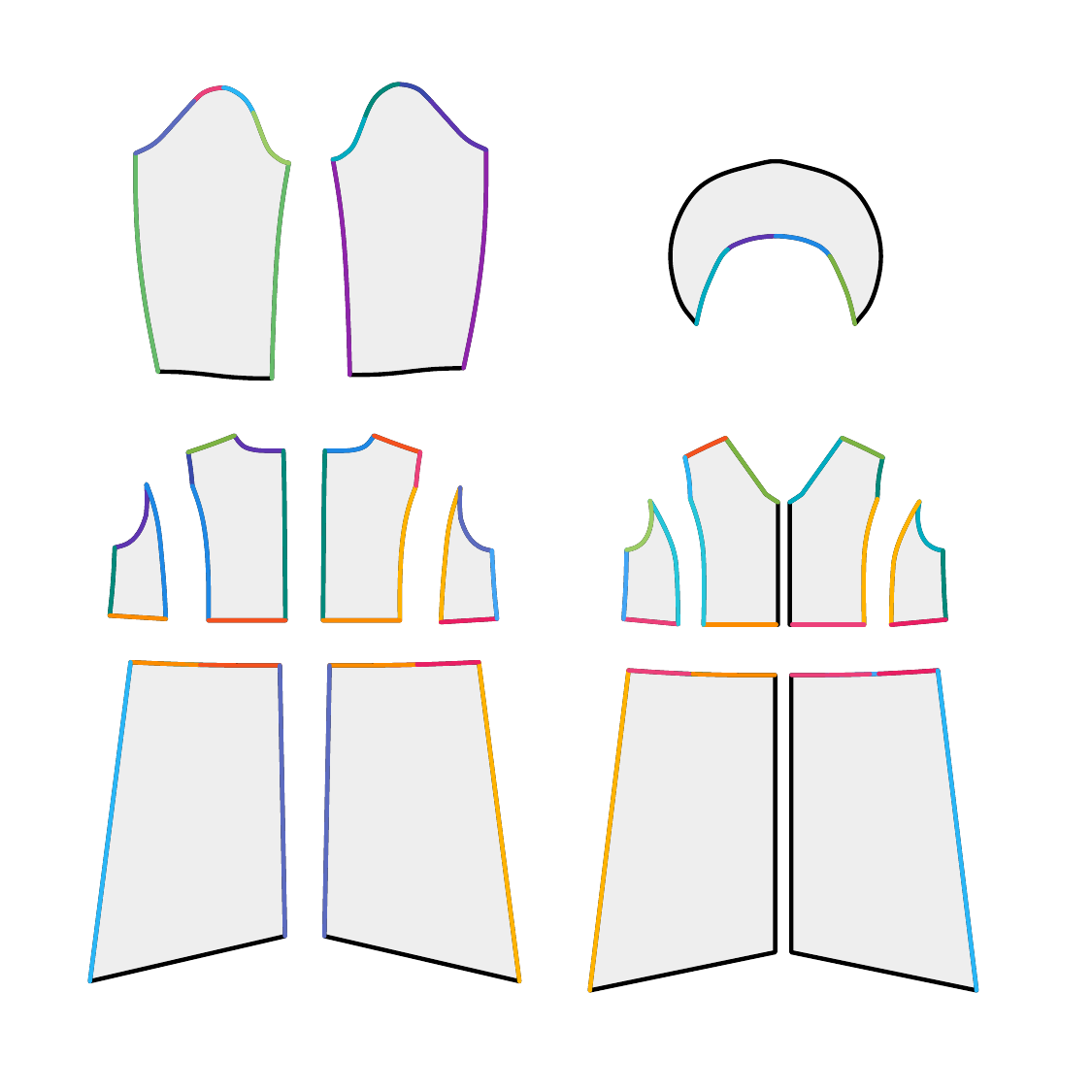}}
    \subfigure[Top]{\includegraphics[height=0.205\linewidth]{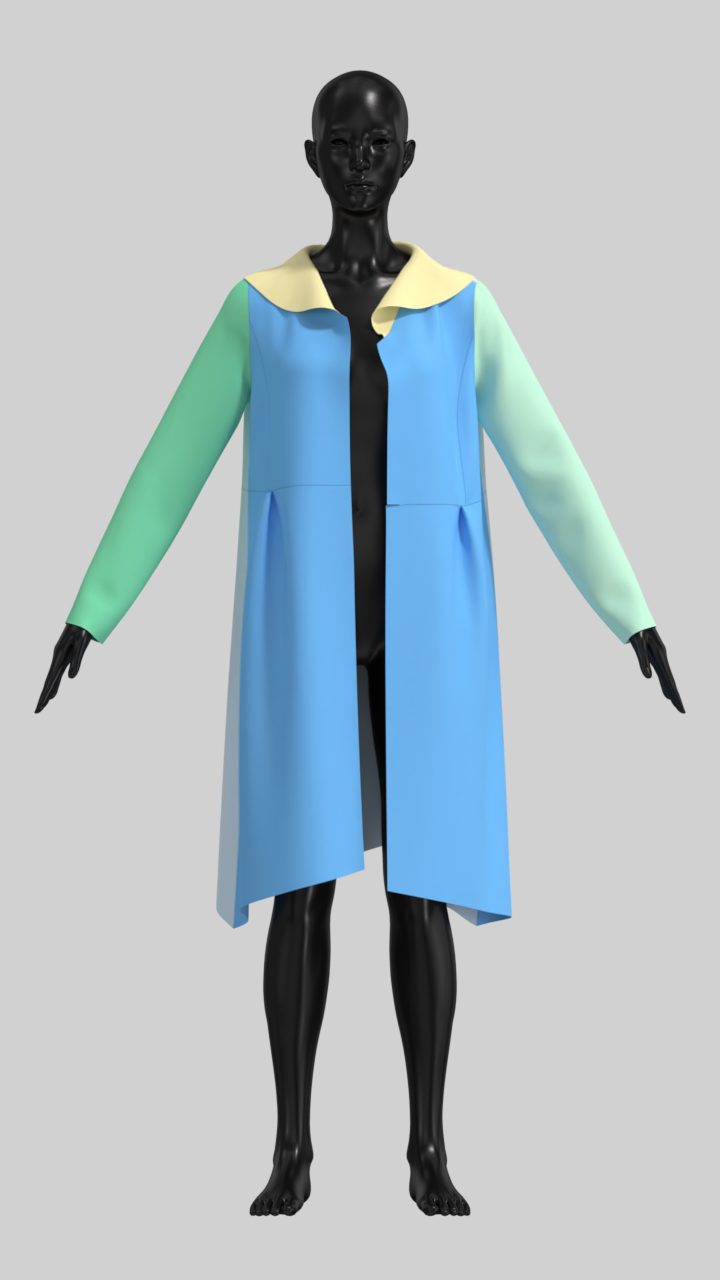}}
    \subfigure[Hoodie pattern]{\includegraphics[height=0.205\linewidth]{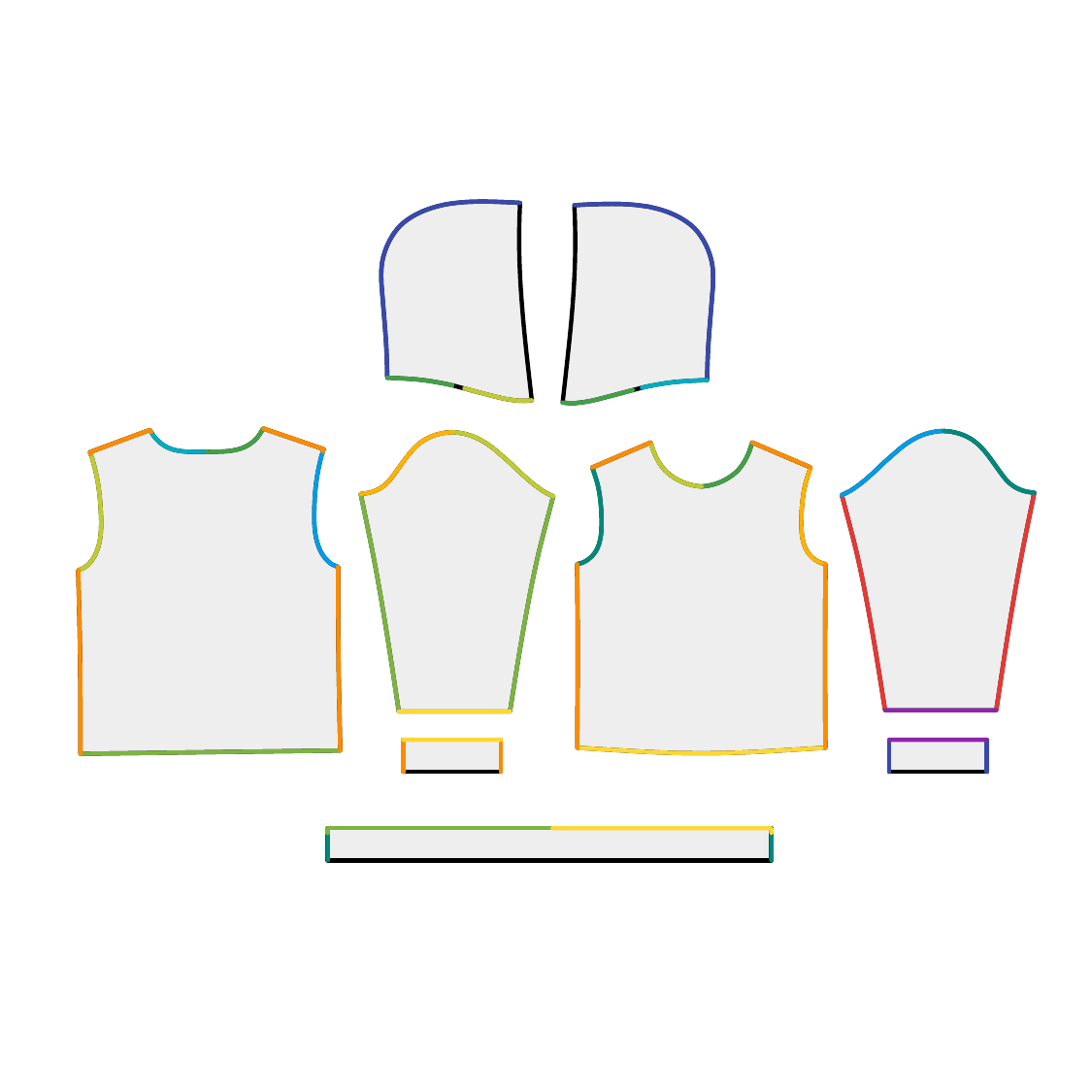}}
    \subfigure[Hoodie]{\includegraphics[height=0.205\linewidth]{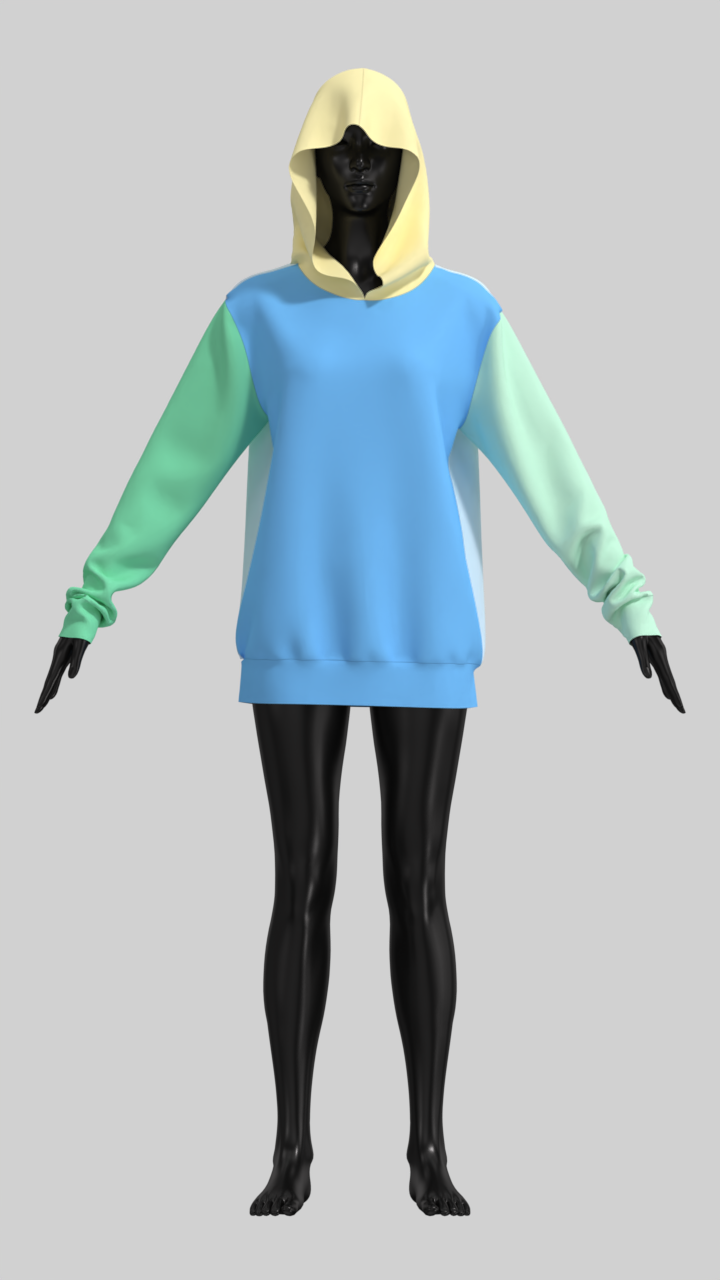}}
    \vspace{-0.2in}
    \caption{Given only the 2D panel geometries of (a) a dress, (c) a top, and (e) a hoodie, our method infers complete seam correspondences that enable faithful 3D garment assembly (b, d, f), together with accurate panel semantics (indicated by colors) aligned with anatomical body regions and coherent panel connectivity.}
    \label{fig:teaser}
\end{teaserfigure}

\maketitle

\section{Introduction}

Garment design has been extensively digitized, ranging from image-based sketching to the generation of precise sewing patterns, enabling efficient design iteration and modification. With the advent of 3D garment modeling, 2D sewing patterns must be converted into 3D assemblies, requiring explicit and detailed stitch specifications between corresponding panel pairs, which are often implicit or absent in traditional patterns. As a result, designers must manually annotate stitch data, a process that is labor-intensive, time-consuming, and reliant on specialized expertise.

Given a sewing pattern consisting solely of disjoint 2D panel geometries with no explicit stitch annotations, the automatic stitching problem is to infer a complete, structurally valid stitch specification that enables assembly of the corresponding 3D garment. This entails: (1) detecting and pairing mutually compatible edge (sub)segments across panels; (2) classifying the stitch type for each matched pair; and (3) estimating associated parameters (e.g., segment reparameterization, orientation, offset, and multiplicity) required for faithful geometric and topological integration. Conventional workflows depend on expert-in-the-loop annotation within specialized CAD tools, which is labor-intensive, error-prone, and difficult to reproduce at scale. Automating seam inference promises higher throughput and consistency. Existing automated approaches predominantly rely on heuristic matching~\cite{Li:2025:GarmageNet} or learned edge-association models~\cite{Berthouzoz:2013:PSP}. However, they are brittle on complex pattern topology (e.g., branching seams, many-to-one correspondences, darts and pleats), curved and non-isometric edge pairs, and heterogeneous discretizations. They also generalize poorly across styles due to limited geometric context modeling and weak treatment of global garment structure, often yielding incomplete, inconsistent, or geometrically incompatible stitch specifications.

In this paper, we present a novel graph-based two-level feature learning method for automatically reconstructing seam correspondence between panel pairs in sewing patterns, operating solely on 2D panel geometries without stitch annotations or semantic labels. The method consists of two stages: (1) panel connectivity reconstruction, with panel semantics corresponding to anatomical garment regions predicted to constrain inter-panel relationships; and (2) seam correspondence prediction, which models fine-grained stitching between panel pairs via latent edge features that jointly encode local seam geometry and global garment context. Across both stages, we employ a panel-centric graph representation in which nodes denote panels and edges encode potential stitching relationships, enabling efficient reasoning over garment structure, inter-panel connectivity, and seam correspondence. Each panel is represented by fixed-resolution images encoding geometric information, including binary masks, boundary masks, tangent vector fields, and distance fields, which are processed by CNN encoders to extract rich geometric features. By jointly modeling local geometric detail and global garment context through node-based graph attention and message passing, the proposed framework robustly handles complex sewing-pattern topologies, diverse panel geometries, and many-to-one seam correspondences. 

We train and evaluate the proposed framework on a comprehensive sewing-pattern dataset covering diverse garment styles. Our method reconstructs complete stitch specifications for faithful 3D garment assembly while simultaneously predicting accurate panel semantics and coherent panel connectivity consistent with garment design conventions. Experimental results demonstrate strong stitching accuracy and robust generalization across garment categories. This work advances automated sewing-pattern processing and has potential applications in digital fashion design, virtual try-on, and rapid prototyping. 

\section{Related Work}

Because seams are intrinsically defined by sewing patterns and encapsulate information about garment design \cite{lu2017new} and modeling\cite{li2023isp,sarafianos2025garment3dgen,liu20183d}, construction\cite{Shen:2020:GanGarment}, and fabrication\cite{Rodriguez:2022:TrueSeams}, related work spans multiple research areas, including stitching inference, garment semantics, pattern generation from images or 3D assets, and sewing pattern datasets and representations.

\emph{Stitch inference and panel assembly.}
Early approaches model stitching as geometric curve matching with handcrafted compatibility criteria and local topology checks, which often fail on curved seams, many-to-one correspondences, and heterogeneous edge discretizations. Representative methods optimize pairwise alignments or search over seam hypotheses using heuristics and soft constraints \cite{Berthouzoz:2013:PSP,Leake:2023:InStitches,Chen:2022:SPNSM,gurarda2019seam}. Auxiliary techniques such as robust curve or shape reparameterization are commonly employed to score candidate matches but remain sensitive to global context \cite{Wang:2018:RSF,Bartle:2016:PGP}. More recent work formulates stitching or panel connectivity prediction as learning on graphs or sequences, leveraging attention mechanisms to aggregate cross-panel context \cite{Chen:2024:Panelformer}. In contrast, we propose a two-level framework that first induces a semantic prior and then performs latent edge feature generation via message passing on a reconstructed panel graph, improving recall under complex topologies while preserving geometric consistency.

\emph{Garment semantics and programmatic pattern representations.}
A line of work focuses on learning structured garment abstractions—such as semantic decompositions, attribute grammars, or domain-specific codes—to enable controllable generation and reasoning under construction constraints \cite{Korosteleva:2022:NeuralTailor,Korosteleva:2023:GarmentCode,Liu:2023:TGSR,Zhou:2025:Design2GarmentCode,Gui:2025:GenPattern}. While these models effectively capture design regularities and long-range dependencies, they primarily target synthesis or editing tasks rather than explicit seam recovery from arbitrary input patterns. Our approach similarly leverages the insight that semantics provide strong structural priors, but uses predicted panel semantics solely to construct a coarse anatomical graph that constrains and guides downstream edge-level stitching inference.

\begin{figure}[t]
    \centering
    \subfigure[Sewing pattern]{\includegraphics[height=0.4\linewidth]{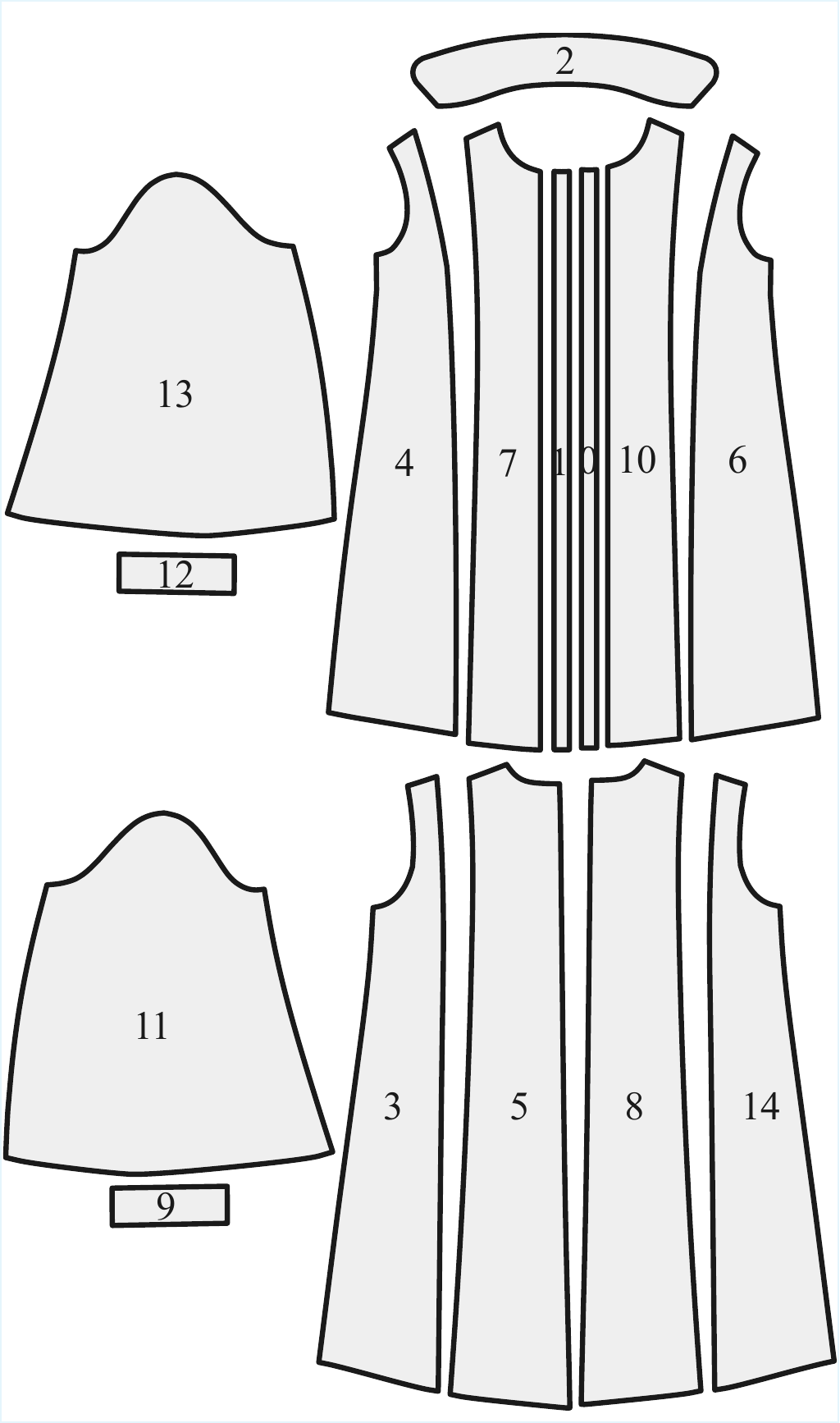}}\hspace{0.01\linewidth}
    \subfigure[Semantics and Connectivity]{\includegraphics[height=0.4\linewidth]{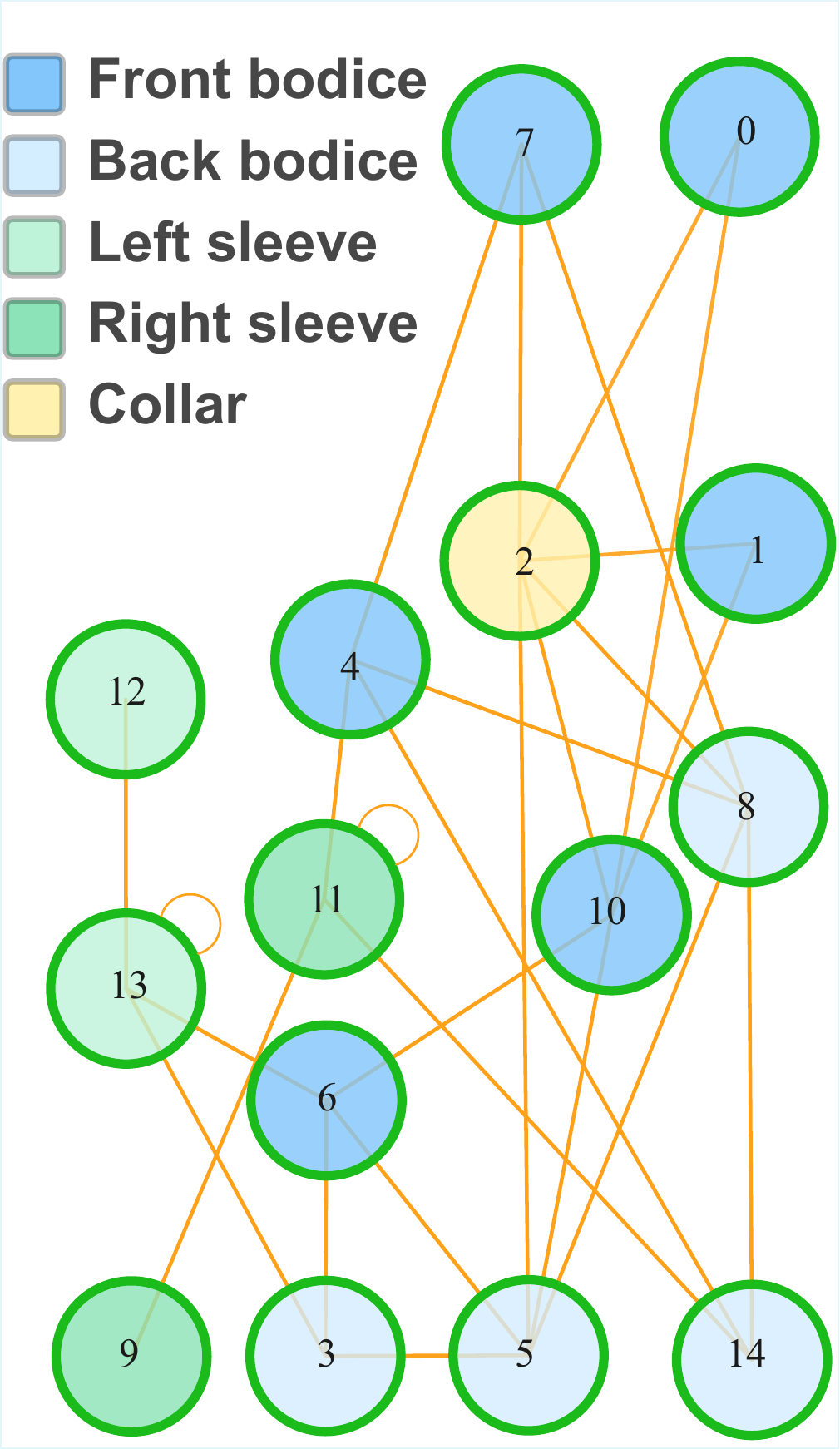}}\hspace{0.01\linewidth}
    \subfigure[Seam correspondence]{\includegraphics[height=0.4\linewidth]{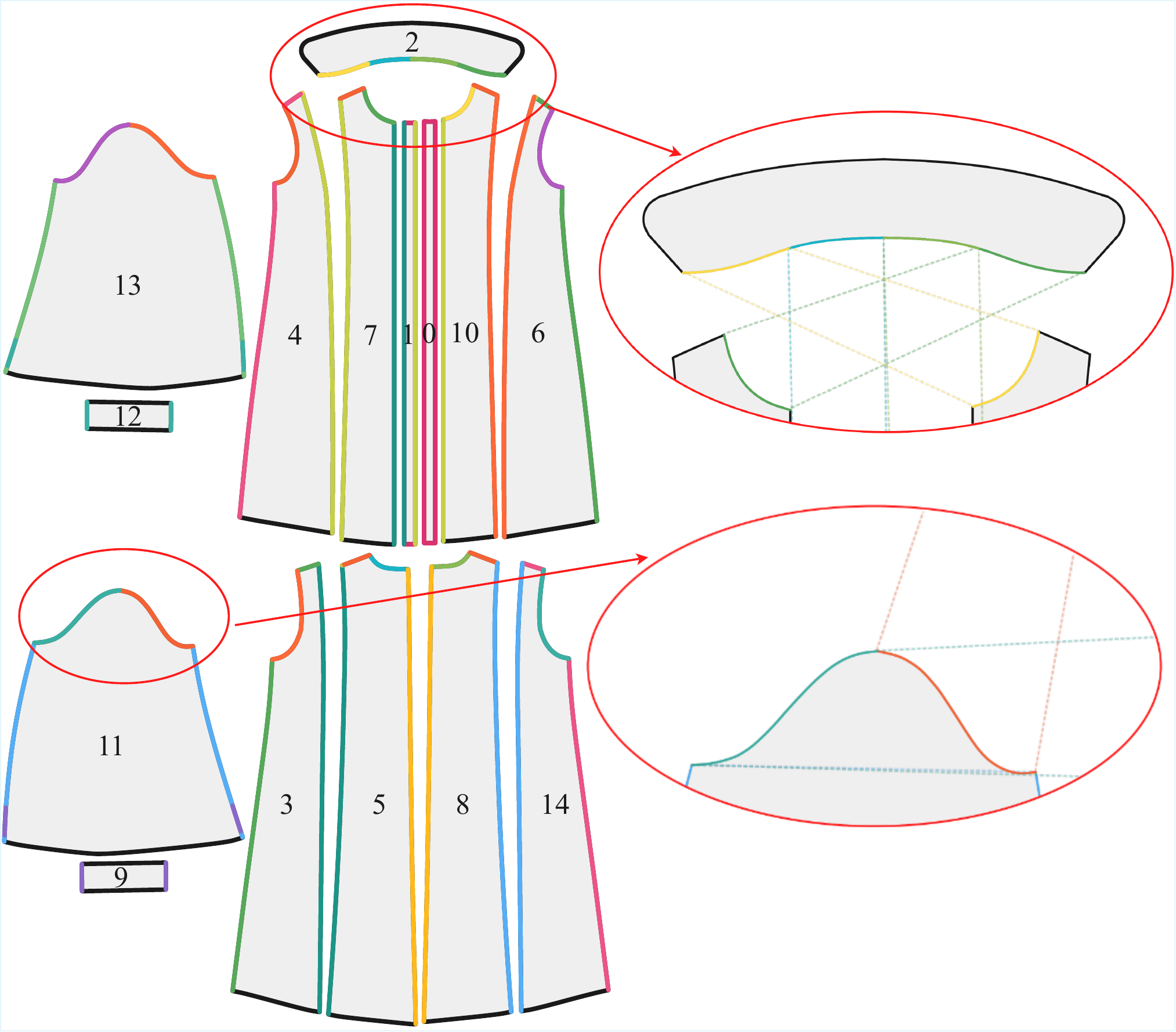}}

    \vspace{-0.1in}
    \caption{Automatic stitching pipeline. (a) Given only 2D panel geometries, our method first predicts (b) panel semantics and reconstructs the panel graph connectivity, then predicts (c) detailed seam correspondences.}
    \label{fig:process}
    \vspace{-0.2in}
\end{figure}

\emph{Sewing pattern generation.} Prior work on inferring sewing patterns from photographs or 3D assets focuses on parsing garment outlines, recovering part layouts, or unwrapping meshes into stitchable panels \cite{Chen:2015:GMDC,Pietroni:2022:CPM,Li:2025:Dress-1-to-3,Tatsukawa:2025:GarmentImage,Liu:2025:MLDM,wang2018learninggarmentshape}. Although some methods implicitly recover stitch relationships as part of their pipelines, their objectives and supervision differ fundamentally from the task of stitching disjoint 2D patterns. In contrast, our setting assumes only panel geometry as input and explicitly targets complete seam specification, including edge correspondence, reparameterization, orientation, offset, and multiplicity.

Public datasets and parsing tools facilitate training and evaluation but rarely provide complete stitch annotations in standard CAD formats, which complicates supervised learning for seam correspondence \cite{He:2024:DressCode,Curtis:2023:DocumentFormat,Qi:2025:Rags2Riches}. To address this gap, we curate a stitched-pattern dataset tailored for evaluating edge-level correspondence and stitch typology, and leverage this corpus to quantify the contributions of semantic priors and two-level feature generation. Compared with heuristic matching methods and single-level learning-based models \cite{Berthouzoz:2013:PSP,Leake:2023:InStitches,Chen:2022:SPNSM,Chen:2024:Panelformer,nakayama2025aipparel}, our approach incorporates garment semantics and aggregates global context into latent edge embeddings. This design enables robust performance on branching seams, curved and non-isometric edges, and many-to-one correspondences, while remaining compatible with standard pattern formats and downstream simulation pipelines.

\section{Problem Statement}

As illustrated in Figure~\ref{fig:process}(a), a sewing pattern consists of a set of disjoint 2D panels, each defined by a non-self-intersecting closed curve that encodes detailed boundary geometry. These panels correspond to individual garment components (e.g., sleeves, torso, collar) and are assembled into a complete 3D garment via seam stitching. In practice, however, sewing patterns typically lack explicit stitch annotations, leaving panel connectivity and seam correspondence to manual specification. Although fabrication-ready patterns may include auxiliary elements such as internal construction lines, grain directions, and notches that facilitate alignment, these increase structural complexity. To ensure tractability, we exclude such auxiliary information and restrict the problem to stitch prediction based solely on panel geometry, assuming no annotations beyond boundary representation.


\begin{figure}[t]
    \centering
    \subfigure[Self-stitch]{\includegraphics[width=0.2\linewidth]{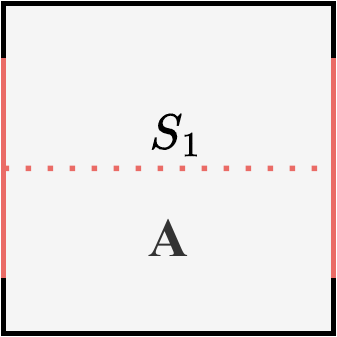}} \hspace{0.03\linewidth}
    \subfigure[One-to-one Single]{\includegraphics[width=0.2\linewidth]{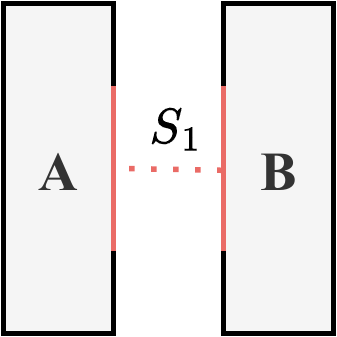}} \hspace{0.03\linewidth}
    \subfigure[One-to-one Multiple]{\includegraphics[width=0.2\linewidth]{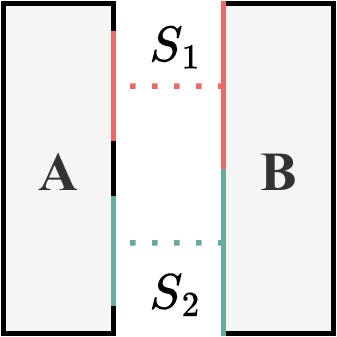}} \hspace{0.03\linewidth}
    \subfigure[Many-to-one]{\includegraphics[width=0.2\linewidth]{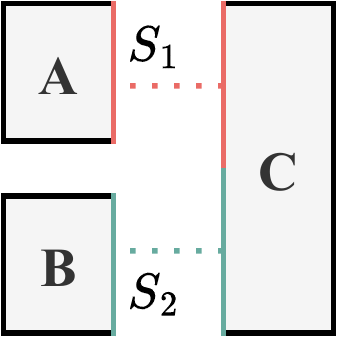}} \hspace{0.03\linewidth}
    \subfigure[Many-to-many]{\includegraphics[width=0.2\linewidth]{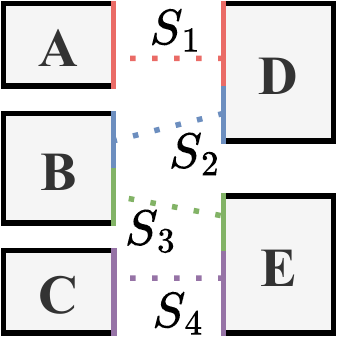}} \hspace{0.03\linewidth}
    \subfigure[Many-to-one overlap]{\includegraphics[width=0.2\linewidth]{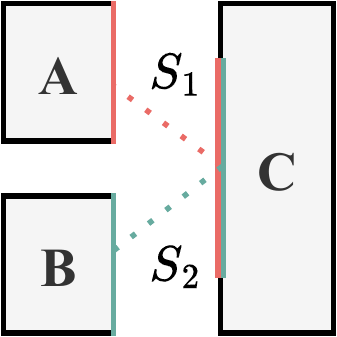}} \hspace{0.03\linewidth}
    \subfigure[Zigzag Self-stitch]{\includegraphics[width=0.2\linewidth]{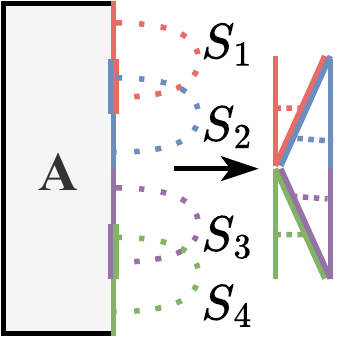}} \hspace{0.03\linewidth}
    \subfigure[Interior attatched]{\includegraphics[width=0.2\linewidth]{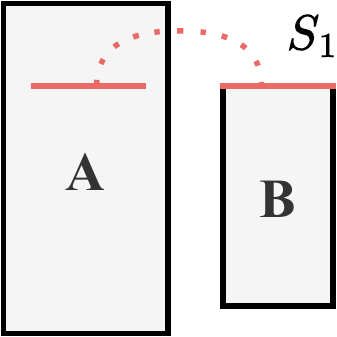}}

    \vspace{-0.1in}
    \caption{We only consider the first five types for seam correspondence prediction and leave the last three types for future work due to their complexity.}
    \label{fig:stitch-types}
    \vspace{-0.2in}
\end{figure}

\subsection{Seam Correspondence}

A stitch specification defines how panels are connected along boundary edges, comprising \emph{boundary edge pairing}, identifying corresponding boundary segments across panels; \emph{orientation}, specifying relative stitching direction; \emph{stitch type}, defining the applied operation (e.g., straight, zigzag, overlock); and \emph{multiplicity}, indicating the number of parallel stitch lines along a seam, and associated parameters such as offset and seam allowance, shown in Figure~\ref{fig:stitch-types}.

Automatically inferring a complete stitch specification is challenging, as it requires joint reasoning over local boundary geometry and the global garment structure. To ensure tractability, we restrict the problem to predicting boundary edge pairings-the core component for garment assembly-and assume stitch orientation can be deterministically derived from the geometry of paired edges. More complex attributes, including stitch types, multiplicity, and additional parameters, are deferred to future work. The objective is thus to infer a complete and consistent set of seam correspondences that accurately encode inter-panel stitching relationships, enabling correct and unambiguous garment assembly.

\subsection{Graph-based Two-level Feature Learning}


To manage the inherent complexity of fine-grained seam modeling, we adopt a two-stage framework that progressively refines seam correspondences from coarse to fine. In the first stage, we estimate coarse panel connectivity to reconstruct a global panel graph, which serves as the topological backbone of the sewing pattern, while simultaneously predicting panel semantic labels corresponding to anatomical garment regions (e.g., sleeves, collar, torso), providing a design-aware prior over inter-panel relationships and guiding subsequent fine-grained seam inference. With these conditions, the second stage models detailed seam correspondences within a structured graph representation. This hierarchical design jointly captures global structural dependencies and local geometric detail, enabling robust and accurate seam correspondence prediction under complex garment topologies and various stitching configurations.


\paragraph{Panel-based Graph}

A naive approach to seam correspondence prediction adopts a curve-centric graph, where boundary curve segments are nodes and correspondences are inferred between them. Although this enables fine-grained modeling, it scales poorly: the number of nodes and candidate matches grows rapidly with garment complexity, seams may span multiple segments, and incorporating high-level garment semantics becomes difficult. To address these limitations, we adopt a panel-centric formulation, representing each panel as a graph node. Panel geometry encodes both local boundary detail and global shape cues for semantic reasoning, while graph edges represent potential stitching relationships between panel pairs. This abstraction captures high-level garment structure and enables efficient inference of inter-panel connectivity.

\section{Our Method}

By taking as input only the 2D geometries of individual panels in a sewing pattern, as illustrated in Figure~\ref{fig:process}, our method produces three principal outputs: (1) panel semantic labels; (2) panel connectivity; and (3) seam correspondence between panel pairs. The first two outputs establish a structured prior over panel relationships that informs the final seam correspondence prediction.

\begin{figure}[t]
    \centering
    \subfigure[Panel]{\includegraphics[height=0.24\linewidth]{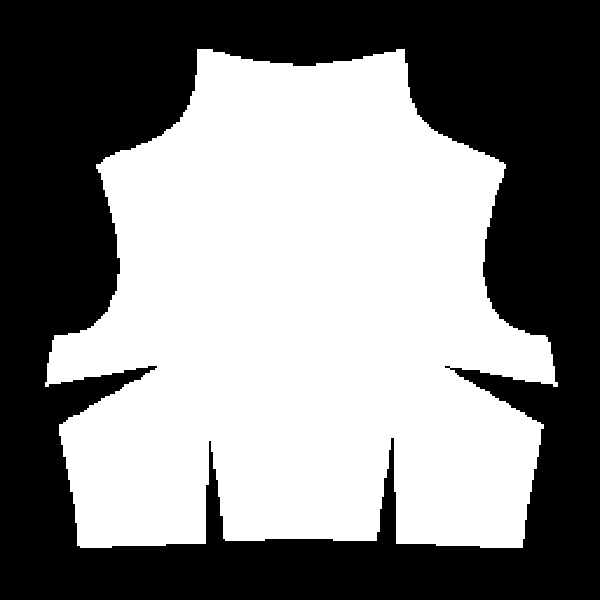}}
    \subfigure[Boundary]{\includegraphics[height=0.24\linewidth]{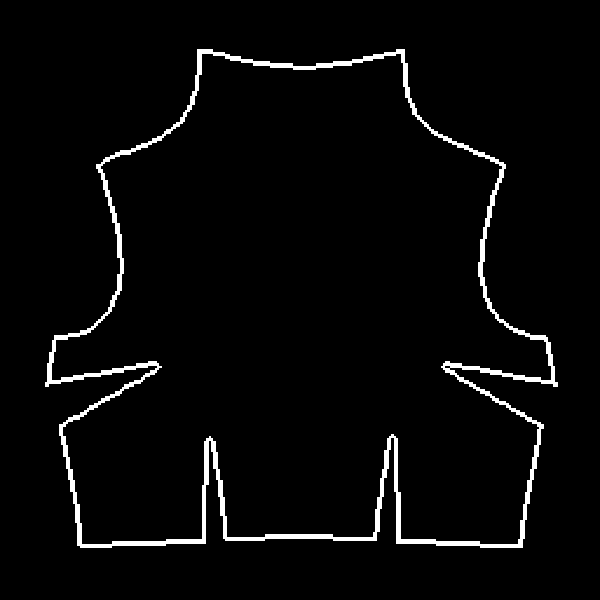}}
    \subfigure[Tangent Vectors]{\includegraphics[height=0.24\linewidth]{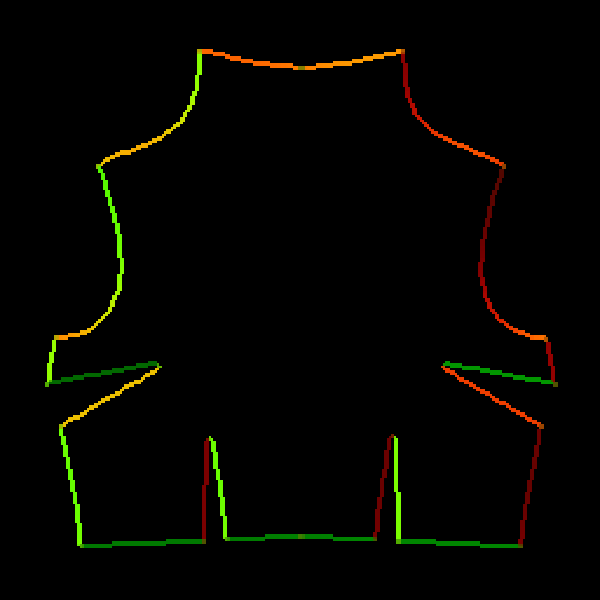}}
    \subfigure[Distance Field]{\includegraphics[height=0.24\linewidth]{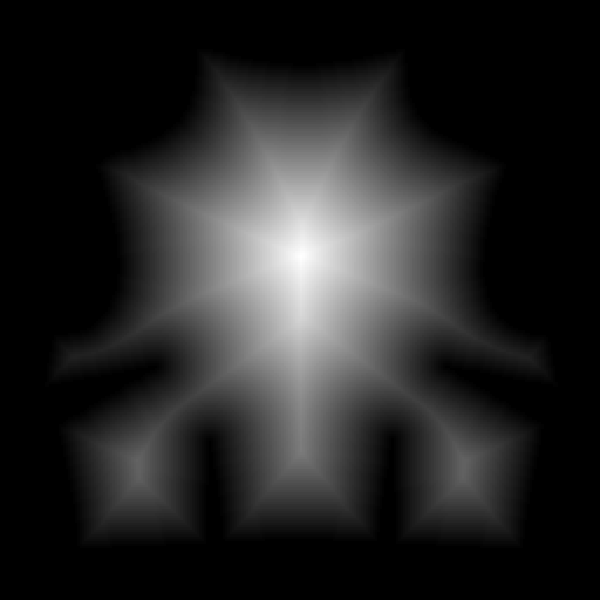}}
    \vspace{-0.1in}
    \caption{Panel representation. (a) and (b) are single-channel binary mask images for encoding the panel shape and boundary; (c) two-channel clockwise boundary tangent vectors; (d) single-channel distance field image.}
    \label{fig:panel-representation}
    \vspace{-0.2in}
\end{figure}

\subsection{Panel Representation}

In digital sewing patterns, panels are typically represented as non-self-intersecting boundary curves, which capture local shape detail but are less expressive for global shape characterization. To address this limitation, we adopt an image-based representation that encodes panel geometry explicitly and enables robust feature extraction via convolutional neural network (CNN) encoders.

As illustrated in Figure~\ref{fig:panel-representation}, each panel is represented by 4 images with resolution of $256\times256$ and pixel width of $1 mm$: (a) a single-channel binary mask encoding the global panel shape; (b) a single-channel boundary mask capturing boundary geometry; (c) a two-channel image encoding clockwise boundary tangent vectors that provide differential geometric cues; and (d) a single-channel distance field image representing the distance to the nearest boundary. For each panel, the maximum dimension is scaled to 0.85 times of the target image width while preserving aspect ratio. Together, these representations support robust, geometry-aware feature extraction. To preserve absolute scale information, we additionally compute explicit geometric descriptors for each panel—including perimeter, area, bounding-box dimensions, and the rasterization scaling factor—which are incorporated as auxiliary panel-level features.

\subsubsection{Explicit Edge Feature Computation}
\begin{figure}[t]
    \centering
    \subfigure[Panels symmetry metric]{\includegraphics[height=0.225\linewidth]{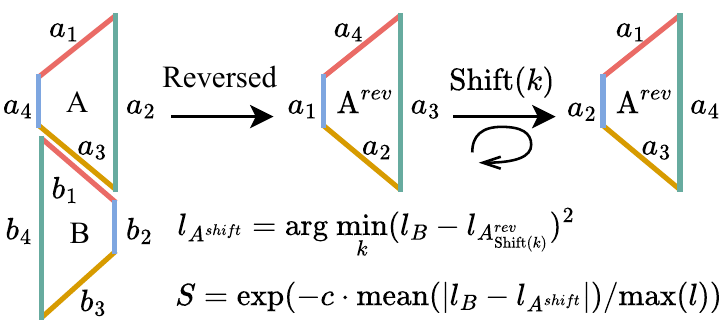}}
    \subfigure[Segments similarity]{\includegraphics[height=0.225\linewidth]{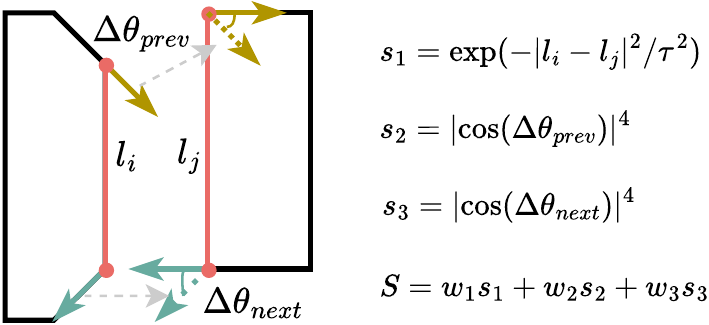}}
    \vspace{-0.1in}
    \caption{Edge feature computation. (a) Symmetry metric for a panel pair based on their most similar segment pair. (b) Segment similarity between two panel edges, with $\tau_l=10$, $w_1=0.7$, $w_2=w_3=0.15$.}
    \label{fig:edge-feature-computation}
    \vspace{-0.2in}
\end{figure}

At this stage, only panel-level node features are available, with no explicit edge features. Nevertheless, panel geometry provides implicit cues about potential seam relationships, some of which can be explicitly derived for candidate pairs. As shown in \autoref{fig:edge-feature-computation}(a), we compute a symmetry indicator by decomposing panel boundaries into smooth, clockwise-oriented segments defined by endpoints. Using the segment similarity metric in \autoref{fig:edge-feature-computation}(b), we identify the most similar segment pair and compute their relative orientation angle $\theta$, represented by $\cos\theta$ and $\sin\theta$. An orientation consistency score, $0.5(1-\cos\theta)\in[0,1]$, quantifies directional alignment, with values near 0 denoting consistent directions and values near 1 indicating opposite directions. These geometric cues are computed for all panel pairs and used as auxiliary edge features for panel connectivity prediction.

\subsection{Topology Reconstruction}

The first task is to predict panel adjacency for reconstructing the sewing-pattern graph topology and to infer semantic labels for enforcing anatomical consistency. This stage establishes a coarse, design-aware prior over inter-panel relationships, guiding subsequent seam correspondence prediction.

\subsubsection{Panel Geometric Feature Encoder}
\label{sec:panel-encoder-a}

In the GNN input layer (Figure~\ref{fig:networks}(a)), we use a multilayer CNN encoder based on EfficientNet~\cite{tan2019efficientnet} to map image-based panel representations to compact latent embeddings. Features from all convolutional layers are concatenated \cite{ghosh2025multi,huang2025multiview,lueangwitchajaroen2024multi} and projected to a fixed 256-dimensional vector via an MLP. Five non-image geometric attributes (e.g., perimeter and area) are then concatenated with the CNN embedding to form enriched panel descriptors. This representation captures both local geometric detail and global shape cues, providing a robust feature space for subsequent graph-based reasoning over panel relationships and semantics.

\subsubsection{Node-based Graph Attention}

In the hidden layers of the GNN (Figure~\ref{fig:networks}(a)), we employ stacked graph attention layers to aggregate neighborhood information and capture global garment structure. Each layer updates panel embeddings by adaptively attending to other nodes and associated edges, with explicit edge features injected directly into every attention layer without additional encoding. Stacking multiple layers enables the modeling of higher-order dependencies and global context, which are critical for accurate panel classification and connectivity prediction.

As no graph topology is available at this stage, standard message passing is inapplicable. We therefore initialize a fully connected panel graph, allowing each node to attend to all others. The attention mechanism learns to selectively weight relevant panels using node features and edge attributes, effectively inferring inter-panel relationships. This design enables recovery of the topological structure of sewing patterns via panel adjacency prediction.

\subsubsection{Prediction Heads}

\begin{figure*}[t]
    \centering
    \includegraphics[width=\linewidth]{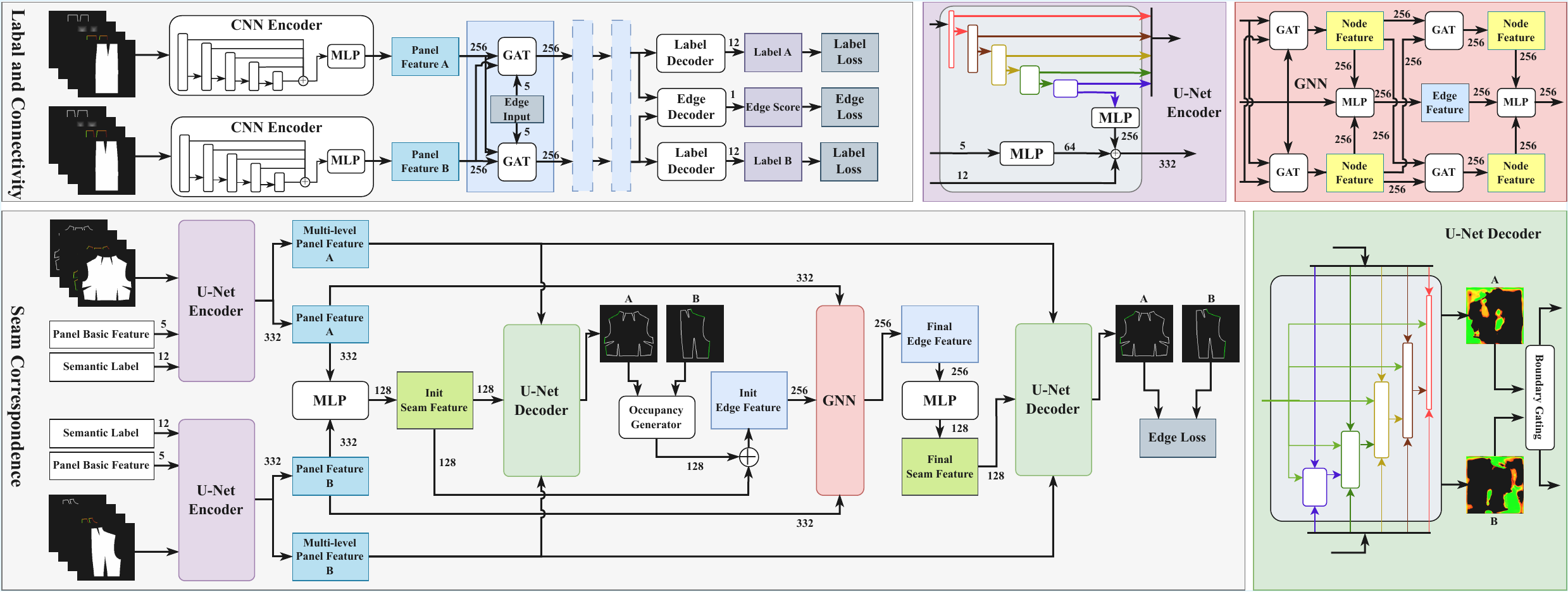}
    \vspace{-0.2in}
    \caption{Network architectures. The first network (top left) predicts panel semantics and panel connectivity. It comprises a EfficientNet-based CNN panel feature extractor followed by three node GAT layers for message-passing. The second network (bottom left) consists of three modules: (i) a U-Net CNN encoder, (ii) two node GAT layers for latent edge feature generation, and (iii) a U-Net decoder for seam correspondence reconstruction. The occupancy field generator is illustrated in \autoref{fig:occupancy}. Specifically, the initial seam feature $f_s^{0}$ is decoded by a pre-trained U-Net decoder $d_0$ and concatenated with its decoded output to form the initial edge feature $f_e^{0}=\{f_s^{0}, d_0(f_s^{0})\}$. After GNN processing, a refined edge feature $f_e^{1}=\{h(f_s^{1}), k(f_s^{1})\}$ is obtained, decoded to $f_s^{1}$ and finally mapped by a U-Net decoder $d_1$ to yield the seam correspondence reconstruction.}
    \label{fig:networks}
    \vspace{-0.1in}
\end{figure*}

As illustrated in Figure~\ref{fig:networks}(a), two prediction heads operate on learned panel and edge embeddings. A panel-level classifier predicts semantic labels from latent geometric embeddings corresponding to anatomical garment regions (e.g., sleeves, collar, torso). We define 12 categories covering common components: collar; waist; left/right sleeves; front/back bodice; front/back skirt; and left/right front/back pants. Panels sharing the same label are grouped into nodes of an anatomical semantic graph, encoding coarse inter-region connectivity as a design prior for graph construction. In parallel, an edge predictor estimates pairwise panel connectivity from learned embeddings. While semantic and connectivity predictions are not explicitly coupled, they share the same node representations. In addition to a supervised topology loss, we introduce a consistency loss enforcing agreement between predicted connectivity and the anatomical semantic graph, promoting structurally and semantically coherent sewing-pattern reconstruction.

\subsection{Seam correspondence reconstruction}

The second task is to predict fine-grained seam correspondence between panel pairs, conditioned on the reconstructed panel graph and enriched panel features. 

\subsubsection{U-Net CNN encoder}

\begin{wrapfigure}{r}{0.25\linewidth}
    \centering
    \includegraphics[width=\linewidth]{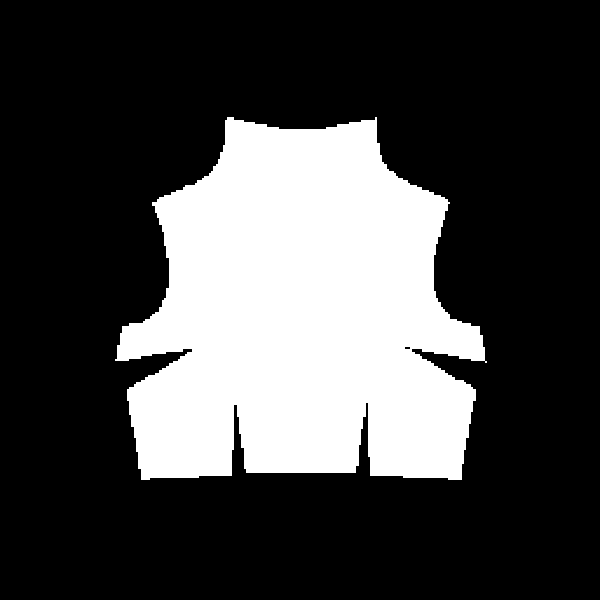}
    \caption{Uniformly scaled panel mask.}
    \label{fig:uniform-scaling-panel}
\end{wrapfigure}
The node inputs to the GNN in Figure~\ref{fig:networks}(b) reuse the panel representations from Figure~\ref{fig:panel-representation}, except that the distance field image is replaced by the original binary mask rendered using a uniform scaling determined by the largest panel dimension (Figure~\ref{fig:uniform-scaling-panel}). This ensures that relative panel sizes are preserved across the graph. Accurate seam correspondence prediction requires joint modeling of local panel boundary geometry and global panel context. Therefore, in this stage, we use a U-Net-style ~\cite{ronneberger2015unet} CNN encoder to explicitly capture multi-scale geometric features without aggregating features from all convolution layers. The 256-dimensional output from the lowest layer is then concatenated with the five non-image geometric attributes (mapped to a 64-dimensional latent vector) and the 12-dimensional predicted semantic label to form an enriched panel feature $f_p\in\mathbb{R}^{332}$, which conditions latent edge feature generation for seam correspondence prediction.

\subsubsection{Edge feature initialization}

\begin{figure}[t]
    \centering
    \includegraphics[width=\linewidth]{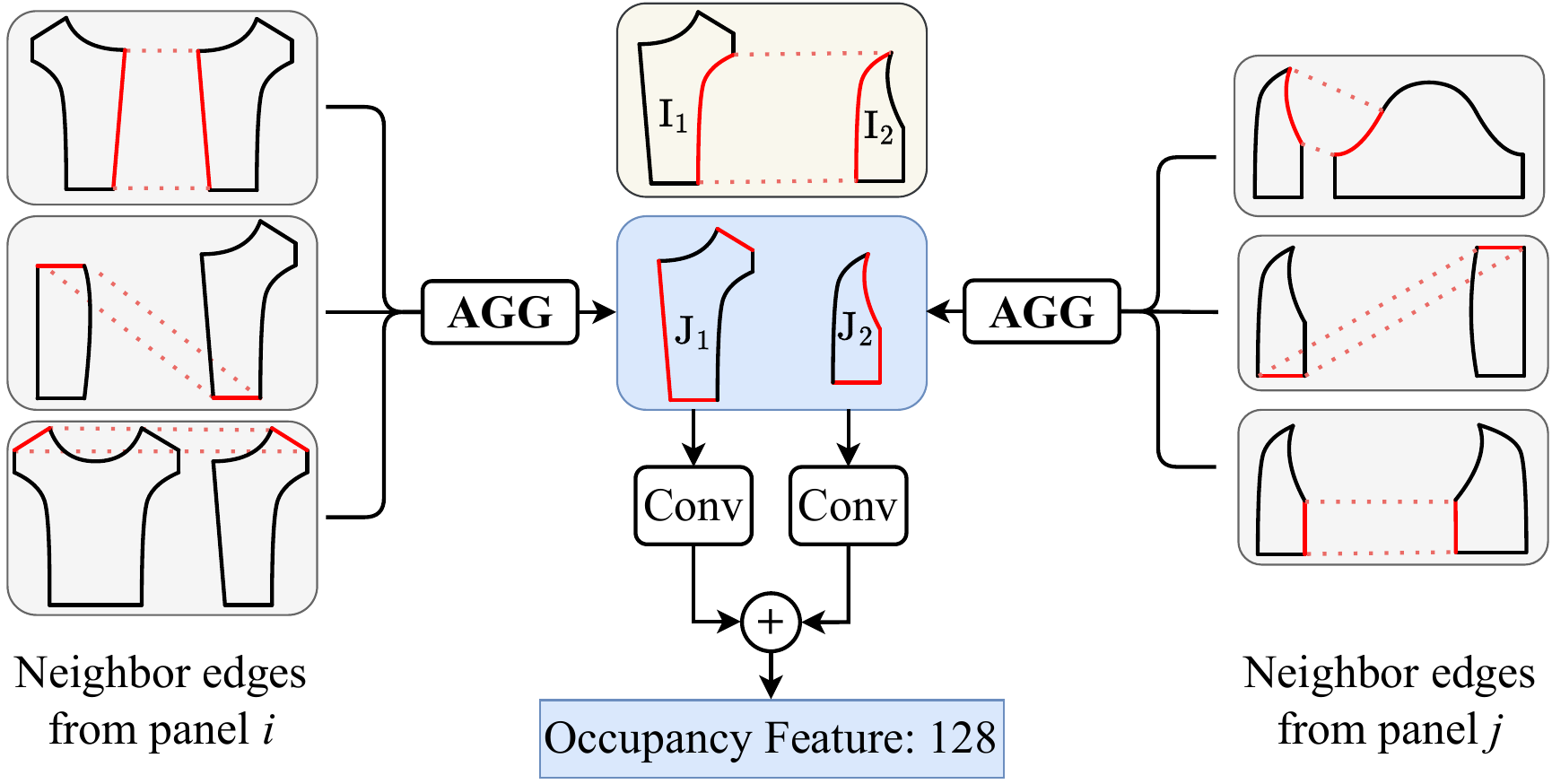}

    \vspace{-0.1in}

    \caption{Occupancy features computation by aggregating seam information from neighboring pairs.}
    \label{fig:occupancy}

    \vspace{-0.2in}
\end{figure}

For each panel pair $(A, B)$, panel features ${f}_{p_A}$ and ${f}_{p_B}$ are first projected via an MLP to an init seam feature $f_{s_{AB}}^0=\text{MLP}(f_{p_A}, f_{p_B})\in\mathbb{R}^{128}$, encoding local inter-panel (fine-grained seam) relationships. A \emph{U-Net decoder} (see section \ref{sec:seam-decoding}) then takes $f_{s_{AB}}^0$ together with multi-level features ${q}_A$ and ${q}_B$ to predict panel-aligned seam maps ${I}_A$ and ${I}_B$ for seam correspondence reconstruction. However, the seam feature $f_{s_{AB}}^0$ encodes only local pairwise information and is insufficient for subsequent GNN edge feature reasoning. To incorporate global context, we explicitly aggregate seam information from neighboring pairs on each panel to form complementary context maps. As illustrated in \autoref{fig:occupancy}, ${I}_A$ encodes local seam on panel $A$, while its complement ${J}_A$ captures seams associated with other neighboring pairs on $A$. $({J}_A,{J}_B)$ are encoded into a global context feature $f_{s_{nb}}^0\in\mathbb{R}^{128}$, yielding the global edge initialization $f_{e_{AB}}^0=(f_{s_{AB}}^0, f_{s_{nb}}^0)\in\mathbb{R}^{256}$.

\begin{figure}[t]
    \centering
    \subfigure[Seam field A]{\includegraphics[width=0.24\linewidth]{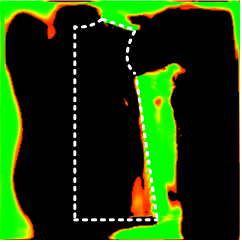}}
    \subfigure[Seam field B]{\includegraphics[width=0.24\linewidth]{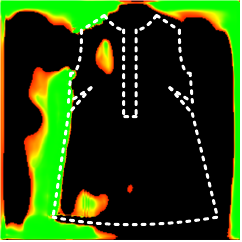}}
    \subfigure[Seam on A]{\includegraphics[width=0.24\linewidth]{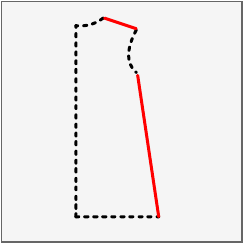}}
    \subfigure[Seam on B]{\includegraphics[width=0.24\linewidth]{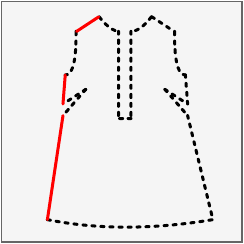}}

    \vspace{-0.1in}

    \caption{Boundary clipping. (a) and (b) are seam fields of two panels. Red segments in (c) and (d) are seam segments marked by the seam fields.}
    \label{fig:gate}

    \vspace{-0.2in}
\end{figure}

\subsubsection{Edge feature refinement}

Panel embeddings together with initialized edge embeddings are processed by a GNN with two node-attention layers. Each layer aggregates neighborhood information by adaptively attending to adjacent panels and associated edges, capturing complex inter-panel interactions, higher-order dependencies, and global garment structure. The network outputs a refined edge embedding $f_{e_{AB}}^1=\text{GNN}(f_{e_{AB}}^0, f_{p_A}, f_{p_B},...)\in\mathbb{R}^{256}$, which serves as the latent seam representation. 

\subsubsection{Seam correspondence decoding}
\label{sec:seam-decoding}


To reconstruct seam correspondence, the final edge embedding $f_{e_{AB}}^1$ should be first projected via a MLP to the final seam feature $f_{s_{AB}}^1\in\mathbb{R}^{128}$. We employ a U-Net decoder, as illustrated in \autoref{fig:networks}, which takes as input $f_{s_{AB}}$ together with multi-level features ${q}_A$ and ${q}_B$, to predict panel-specific seam field images for the two panels. As shown in \autoref{fig:gate}, these fields are intersected with the panel boundary masks to extract seam-edge pixels, yielding panel-aligned seam maps ${I}_A$ and ${I}_B$ at $256\times256$ resolution consistent with the ground truth. Their difference is used to define a seam loss for supervising seam feature learning.



\subsubsection{Seam correspondence reconstruction}

\begin{figure}[t]
    \centering

    \subfigure[Clustering]{\includegraphics[width=0.24\linewidth]{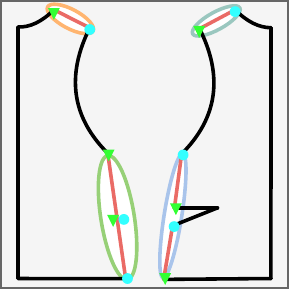}} \hspace{0.02\linewidth}
    \subfigure[Cluster matching]{\includegraphics[width=0.24\linewidth]{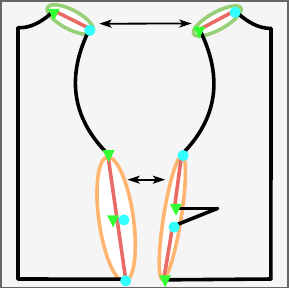}} \hspace{0.02\linewidth}
    \subfigure[Intra-cluster correspondence]{\includegraphics[width=0.24\linewidth]{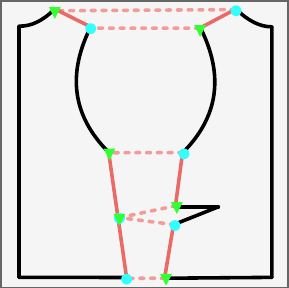}}

    \subfigure[One-to-one]{\includegraphics[width=0.24\linewidth]{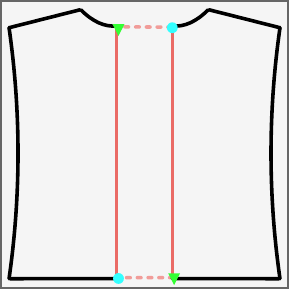}} \hspace{0.02\linewidth}
    \subfigure[Many-to-one]{\includegraphics[width=0.24\linewidth]{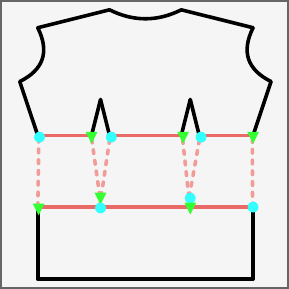}} \hspace{0.02\linewidth}
    \subfigure[Many-to-many]{\includegraphics[width=0.24\linewidth]{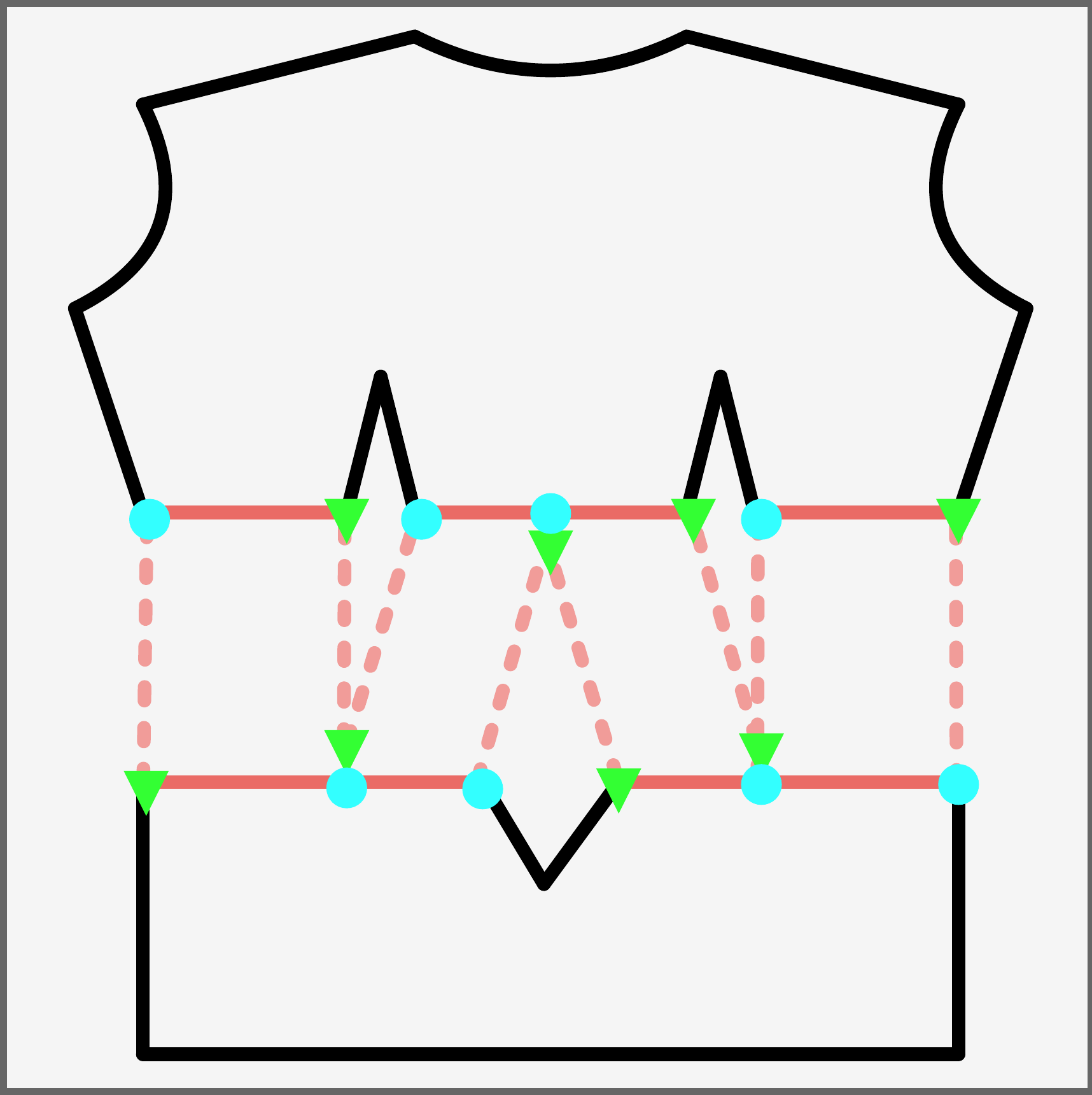}}

    \vspace{-0.1in}

    \caption{Green triangles and cyan circles denote seam segment start and end points, respectively. Three steps for seam correspondence reconstruction: (a) seam segment clustering, (b) cluster matching, (c) intra-cluster correspondence prediction. (d-f) Different seam correspondence types.}
    \label{fig:seam-correspondence}

    \vspace{-0.2in}
\end{figure}

To enable unambiguous seam reconstruction, we enforce a consistent clockwise orientation for all panel boundary curves, thereby uniquely defining the start and end points of each seam segment, denoted by green triangles and cyan circles, respectively, as shown in Figure~\ref{fig:seam-correspondence}. Seam correspondence reconstruction proceeds in three steps: (a) seam segment clustering, (b) cluster matching, and (c) intra-cluster correspondence inference. First, seam segments on each panel are clustered based on spatial proximity and geometric similarity. Multiple seam segments on a panel are clustered only when they are connected by darts. Second, clusters across the two panels are matched using a learned matching network operating on cluster-level features. For each matched cluster pair, new start and end points are defined on the panel with fewer seam segments via proportional length segmentation. Seam segments on the panel with more segments are then aligned to these reference points based on their relative boundary positions. This strategy robustly resolves complex seam correspondences, including many-to-one and many-to-many configurations. Finally, within each matched cluster pair, intra-cluster seam correspondences are inferred by considering three correspondence types: one-to-one, many-to-one, and many-to-many (Figures~\ref{fig:seam-correspondence}(d–f)).

\subsection{Training}


Specifically, the two models in our method are optimized using AdamW with a learning rate of 1e-4, cosine learning rate scheduling, and a weight decay of 0.01. Mixed-precision training is employed to reduce GPU memory usage, with a batch size of 2. At the first stage, panel nodes in the sewing-pattern graph are initially independent without connectivity, preventing effective graph message passing. To address this issue, we employ randomly sampled graph topologies during training, including fully connected edges (20\%), noisy edges (50\%), and top-k edges selected using cosine similarity of CNN-extracted shape features (30\%). During inference, a fully connected graph is used to enable complete message passing. 


\subsubsection{Dataset}


To train our models, we curate a sewing-pattern dataset with detailed stitch annotations covering diverse garment styles. Each sample consists of disjoint 2D panel geometries and corresponding stitch specifications, including 2,596 dresses, 393 pairs of pants, and 1,910 tops. The dataset is randomly split into training (80\%), validation (10\%), and test (10\%) sets for model optimization, evaluation, and inference. For generalization assessment, we additionally construct an out-of-distribution test set of 90 garments with unseen styles, including coats and jackets, to evaluate robustness to novel designs.

\subsubsection{Training Strategy}

Fine-grained seam correspondence learning is substantially more challenging than panel connectivity reconstruction, as it requires modeling intricate inter-panel relationships. To address this, we adopt a two-stage training strategy that balances local feature learning at the panel and seam levels with global garment-level context modeling.


Given a panel pair with ground-truth seam correspondence, we extract image-based representations as described in \autoref{fig:panel-representation} and input them into the U-Net encoder to obtain panel embeddings. These embeddings are used to generate seam features, which are decoded by the U-Net decoder to produce panel-wise seam maps for edge loss computation. This pre-training stage excludes all GNN components and focuses solely on pairwise learning. It enables the U-Net to learn robust, geometry-aware representations that capture local seam-edge characteristics, thereby facilitating subsequent seam correspondence reconstruction.



We integrate the pre-trained U-Net modules into the full GNN-based seam correspondence pipeline and fine-tune the entire network end-to-end. This stage incorporates global garment context and inter-panel dependencies into the seam features, enabling more accurate seam correspondence reconstruction.


\begin{figure}[t]
    \centering
    \subfigure[]{\includegraphics[width=0.3\linewidth]{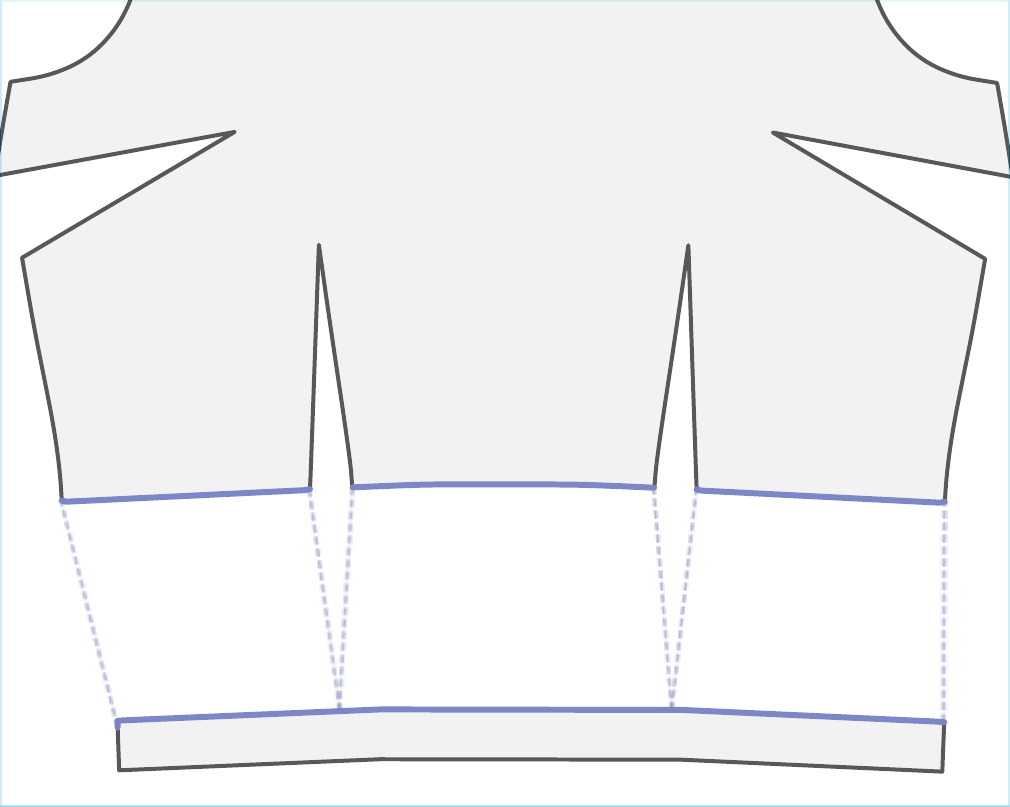}} \hspace{0.03\linewidth}
    \subfigure[]{\includegraphics[width=0.3\linewidth]{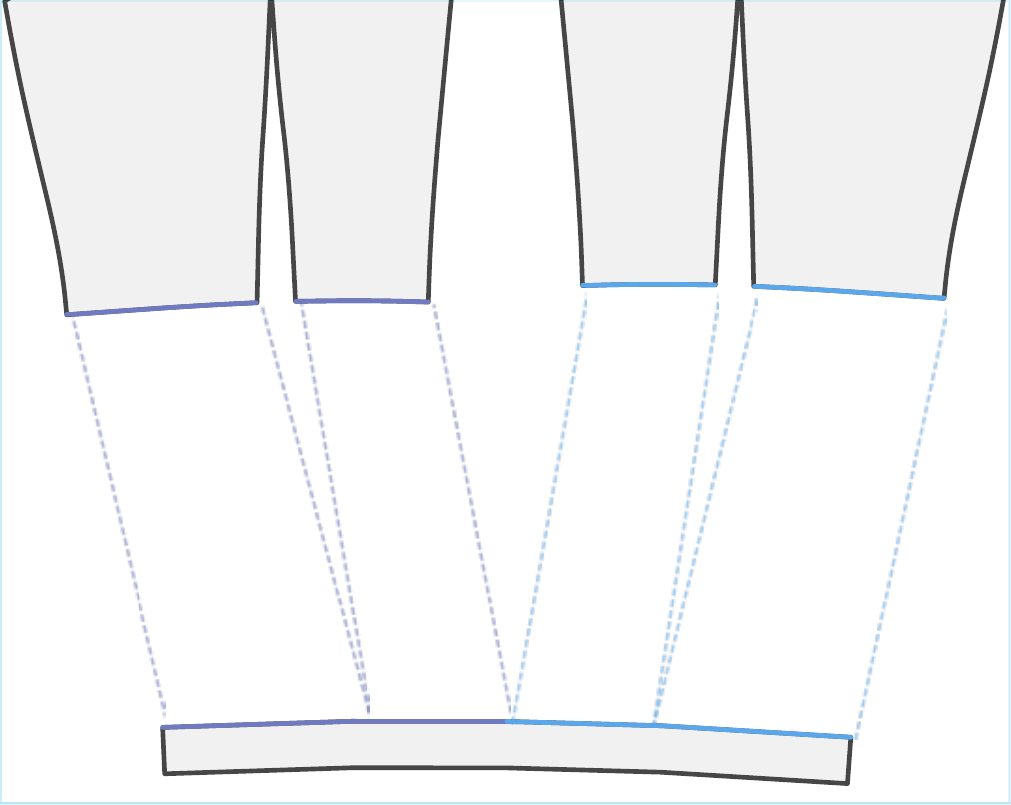}} \hspace{0.03\linewidth}
    \subfigure[]{\includegraphics[width=0.3\linewidth]{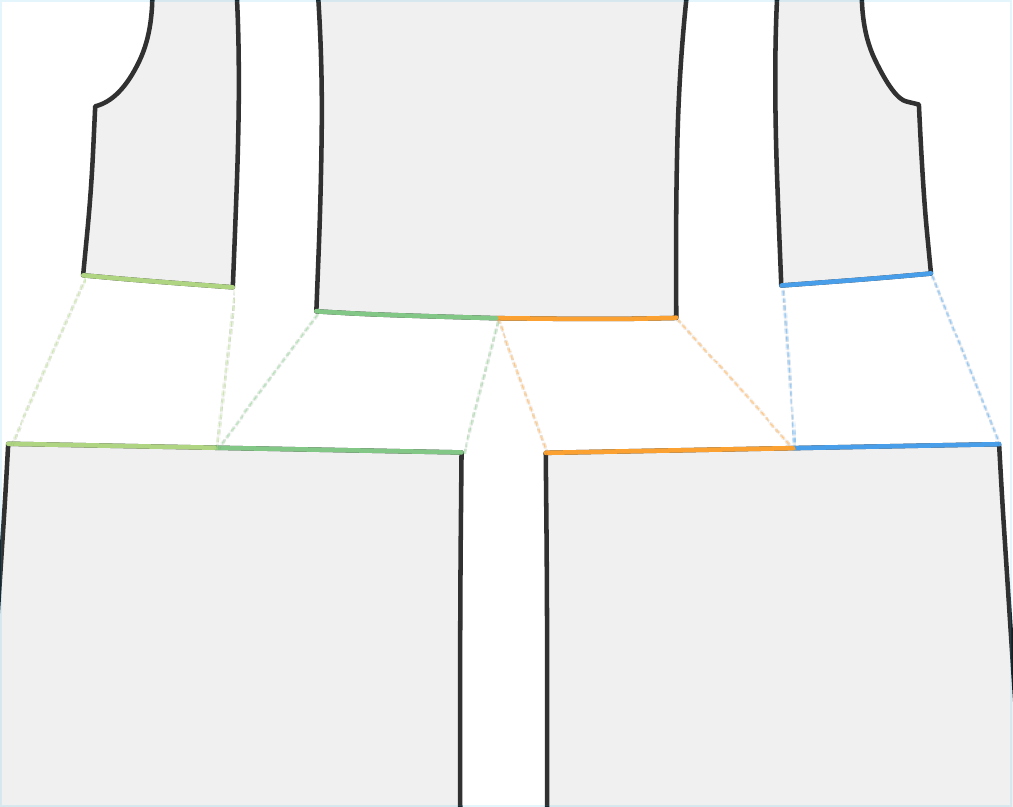}}
    
    \vspace{-0.1in}
    
    \caption{Examples of complex stitch types correctly predicted by our method.}
    \label{fig:stitch-types-ture}

    \vspace{-0.2in}
\end{figure}

\section{Results}

We evaluate our method on the curated stitched-pattern dataset using stitching accuracy, recall under complex topologies, and cross-style generalization as evaluation metrics.

\subsection{Evaluation and Ablation}


Table \ref{tab:first-level} evaluates the model’s performance on panel semantic classification and panel connectivity reconstruction. It reports label accuracy (LA) for semantic prediction and panel-pair connectivity metrics including F1-score (E-F1), optimal decision threshold (THR), precision (E-P), recall (E-R), and micro average precision precision (MAP). The full model achieves the best overall performance, indicating that the combination of learned panel descriptors, explicit pairwise edge features, and graph-based context aggregation is most effective. Removing the GNN results in a substantial drop in both LA and F-1 score, highlighting the critical role of message passing in capturing global garment structure. Excluding explicit edge features degrades connectivity metrics (E-F1/MAP), confirming that boundary-derived symmetry and orientation cues provide complementary information beyond node embeddings. Input ablations demonstrate that boundary-related channels significantly contribute to both semantic discrimination and connectivity inference.

\begin{table}[t]
    \caption{Ablation study on semantic label prediction and panel connectivity prediction. \emph{EF} denotes edge features; \emph{BK} the boundary mask image; and \emph{TD} the boundary tangent vector field and distance field. \emph{E-F1} is the F1-score for edge connectivity prediction; \emph{THR} the optimal decision threshold; \emph{E-P} and \emph{E-R} the precision and recall, respectively; and \emph{MAP} the micro average precision for edge connectivity.}
    \begin{tabular}{c c c c c c c}
        \hline
        Models & LA (\%) & E-F1 & THR & E-P & E-R & MAP \\
        \hline
        Full Model & 90.9 & 0.9073 & 0.74 & 0.8995 & 0.9162 & 0.9719 \\
        w/o GNN  & 84.9 & 0.8449 & 0.80 & 0.8191 & 0.8723 & 0.9245 \\
        w/o EF   & 86.4 & 0.8256 & 0.75 & 0.8277 & 0.8235 & 0.9151 \\
        w/o BK   & 89.6 & 0.8815 & 0.78 & 0.8834 & 0.8797 & 0.9691 \\
        w/o TD   & 88.0 & 0.8608 & 0.72 & 0.8410 & 0.8816 & 0.9473 \\
        \hline
    \end{tabular}
    \label{tab:first-level}
    \vspace{-0.15in}
\end{table}

Table~\ref{tab:second-level} evaluates the fine-level seam correspondence module using pixel-wise agreement between predicted and ground-truth seam indicator images via the Dice coefficient (Pixel Dice), geometric fidelity through relative length error (RLE), and structural consistency using the overlap ratio to measure the incidence of overlapping predicted seams. The full model achieves the highest Dice score while maintaining low RLE and minimal overlap, indicating accurate and topologically consistent seam recovery. Removing the GNN results in a pronounced degradation in Dice and substantial increases in RLE and overlap, demonstrating that global graph context is essential for resolving ambiguous and many-to-one correspondences. Disabling local module pre-training consistently degrades performance, confirming that encoder–decoder initialization with pairwise seam supervision stabilizes and improves subsequent global fine-tuning. Replacing the U-Net with a simpler CNN encoder further reduces accuracy, underscoring the importance of multi-scale feature aggregation and decoder-side reconstruction for pixel-level seam prediction. Finally, input ablations—including image-only inputs, removal of boundary cues, and removal of tangent or uniform-scale information—verify that boundary-aware representations are critical for precise localization during seam decoding.

\begin{table}[t]
    \caption{Ablation study on seam correspondence prediction. \emph{PT} denotes local module pre-training; w/o \emph{U-Net} indicates replacing the U-Net with a plain CNN encoder; \emph{BK} is the boundary mask image; and \emph{TF-UPM} are the tangent vector field and the uniformly scaled panel mask image. \emph{Pixel Dice} measures pixel-wise agreement between predicted and ground-truth seam indicator images; \emph{RLE} denotes the relative length error of predicted seams; and \emph{Overlap Ratio} is the ratio of overlapping seams to total seams.}
    \begin{tabular}{c c c c }
        \hline
        Models & Pixel Dice & RLE& Overlap Ratio \\
        \hline
        Full Model & 0.8493 & 0.1829 & 0.0072 \\
        w/o GNN  & 0.6340 & 1.3039 & 0.2678 \\
        w/o PT   & 0.8408 & 0.2008 & 0.0142 \\
        w/o U-Net   & 0.7871 & 0.2907 & 0.0103 \\
        only image  & 0.7820 & 0.2351 & 0.0092 \\
        w/o BK  & 0.7998 & 0.2144 & 0.0063 \\
        w/o TF-UPM  & 0.7542 & 0.2591 & 0.0081 \\
        \hline
    \end{tabular}
    \label{tab:second-level}
    \vspace{-0.1in}
\end{table}

\subsection{Complex Stitch Types}
As shown in \autoref{fig:stitch-types-ture}, our method accurately predicts complex stitch types, including one-to-one, many-to-one, and many-to-many correspondences. It robustly handles these challenging configurations and demonstrates particularly strong performance on many-to-one seams, which are known to be difficult for prior approaches.

\subsection{Garment Style Generalization}
\begin{figure*}[t]
    \centering

    \subfigure[]{\includegraphics[width=0.32\linewidth]{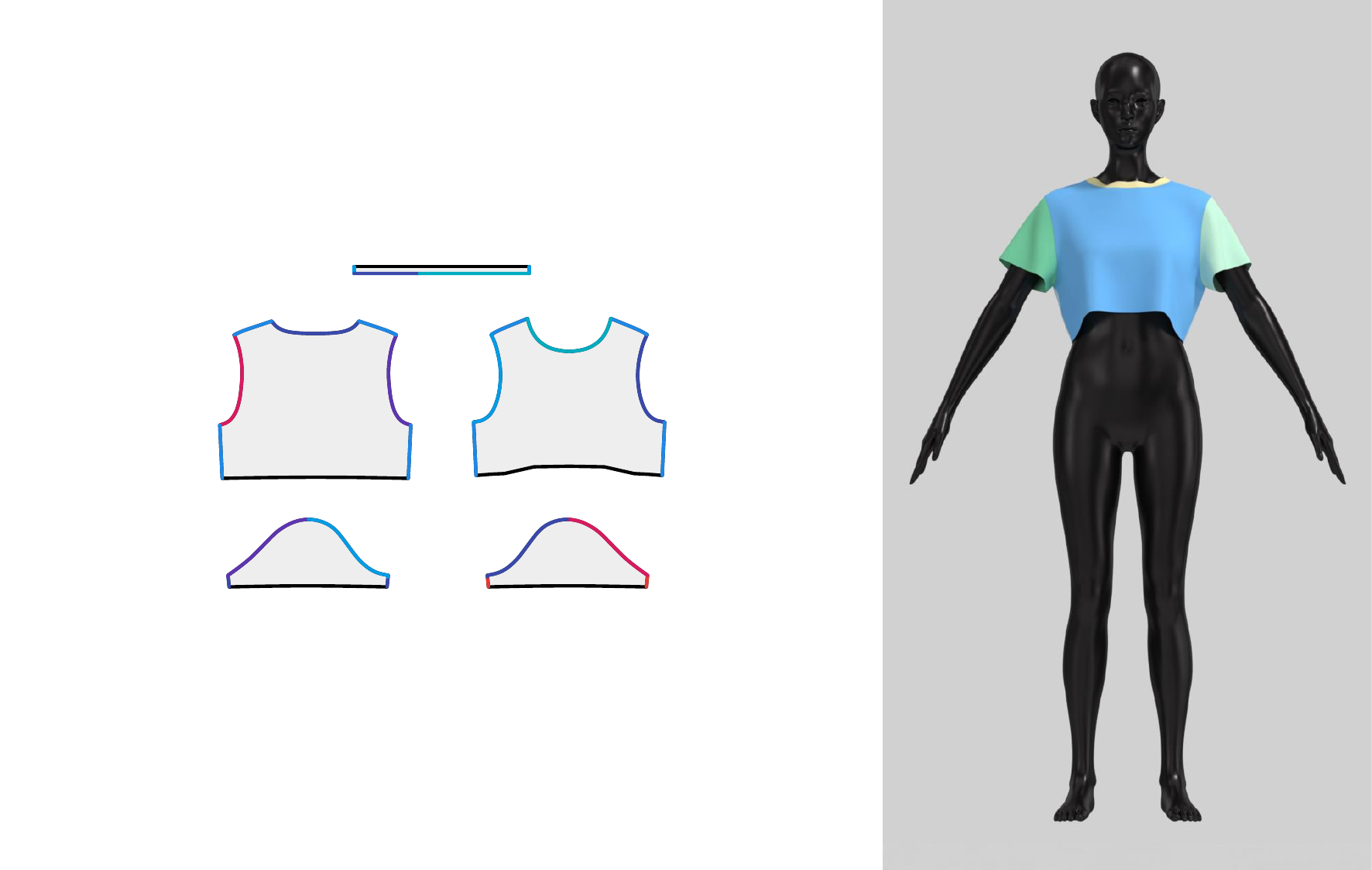}}
    \subfigure[]{\includegraphics[width=0.32\linewidth]{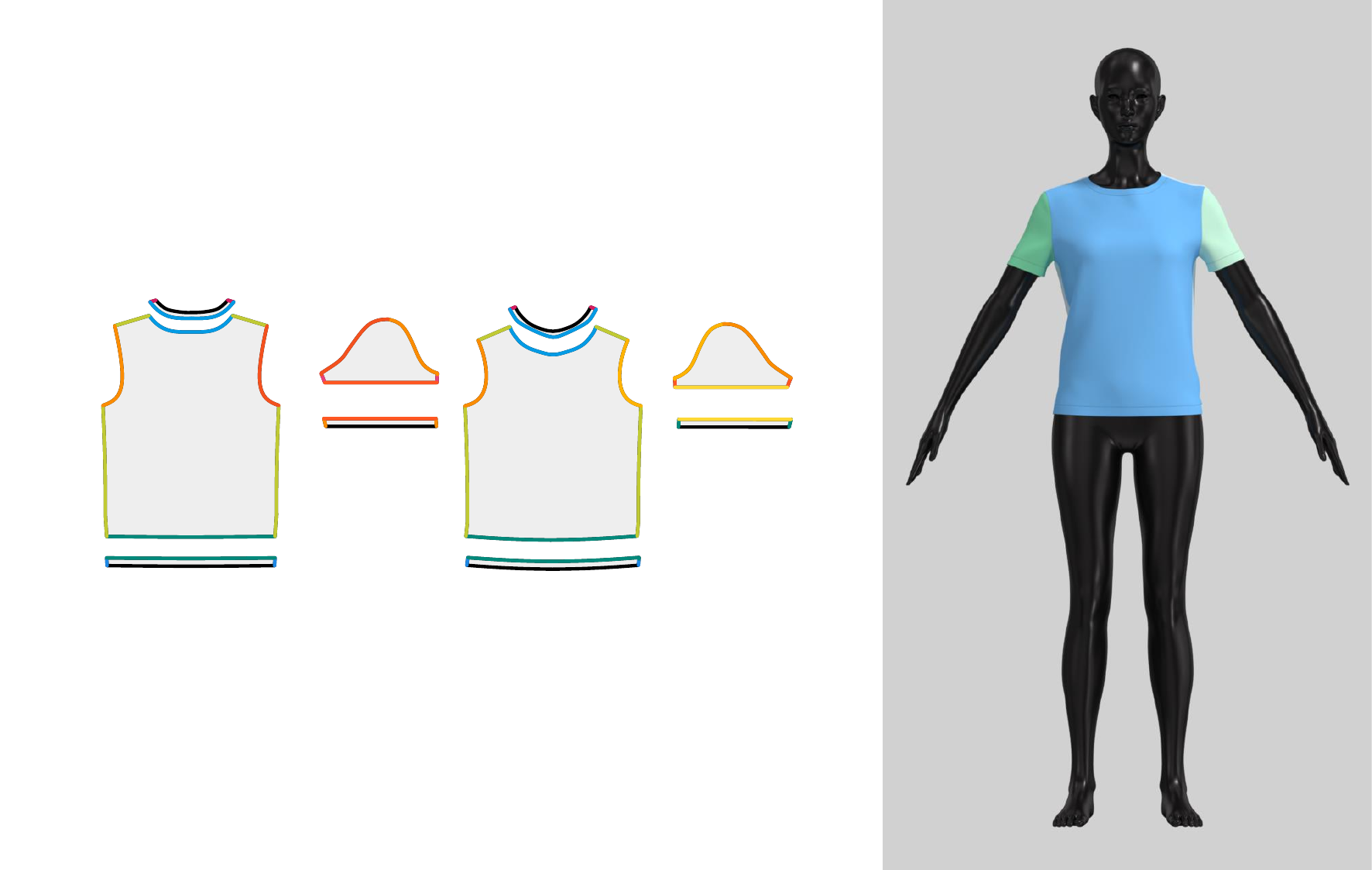}}
    \subfigure[]{\includegraphics[width=0.32\linewidth]{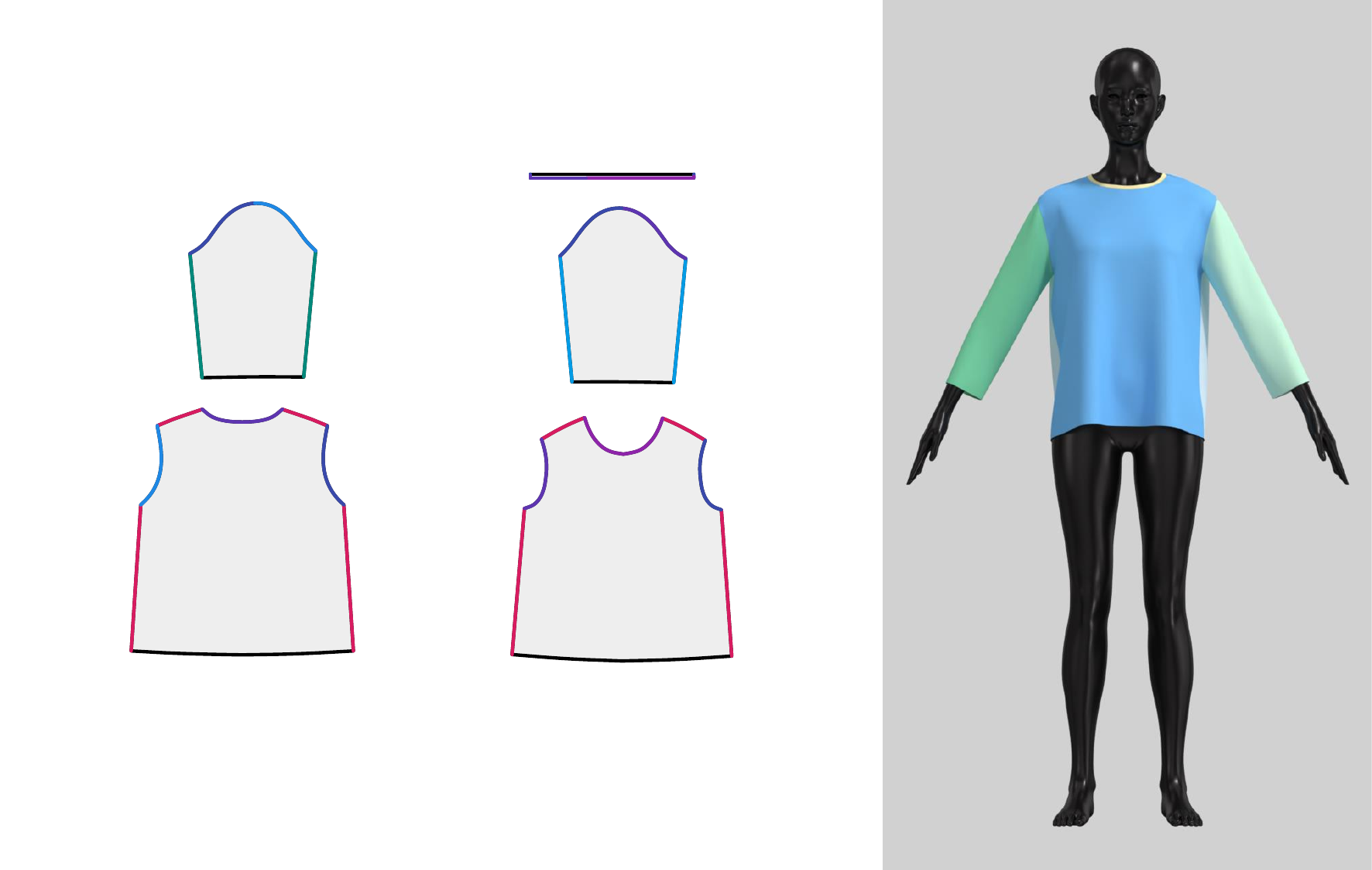}}

    \vspace{-0.1in}

    \subfigure[]{\includegraphics[width=0.32\linewidth]{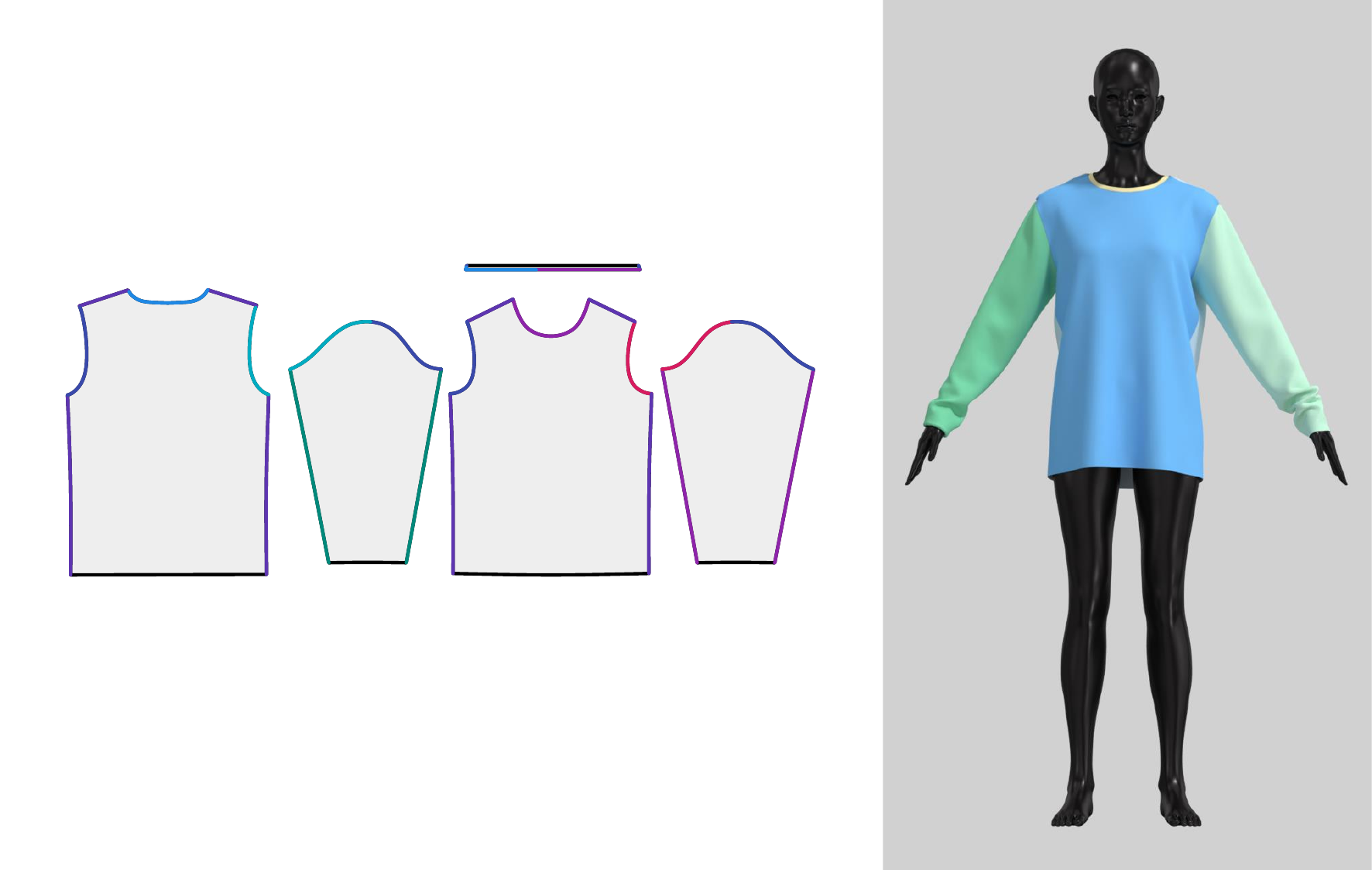}}
    \subfigure[]{\includegraphics[width=0.32\linewidth]{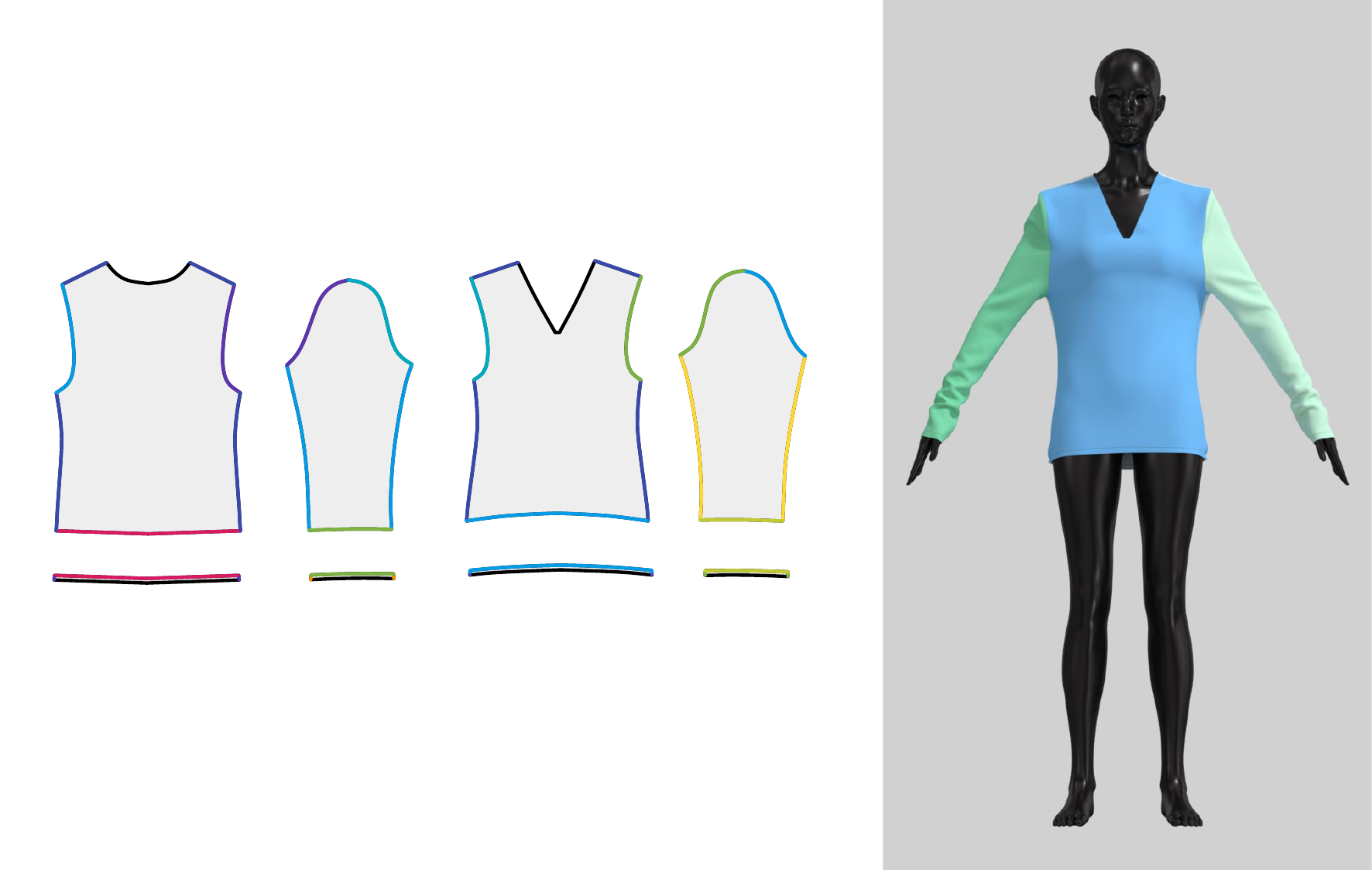}}
    \subfigure[]{\includegraphics[width=0.32\linewidth]{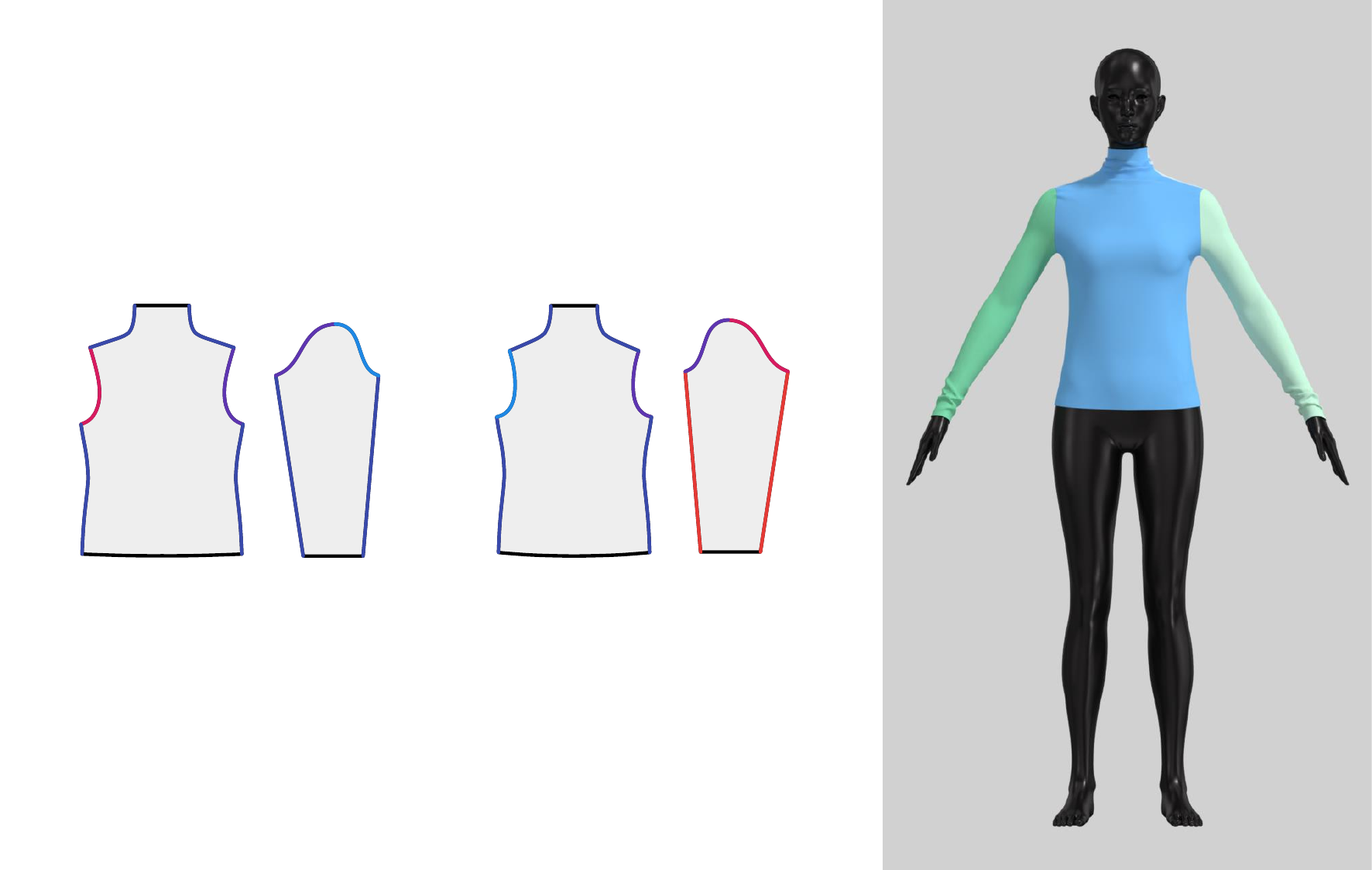}}
    
    \vspace{-0.1in}

    \subfigure[]{\includegraphics[width=0.32\linewidth]{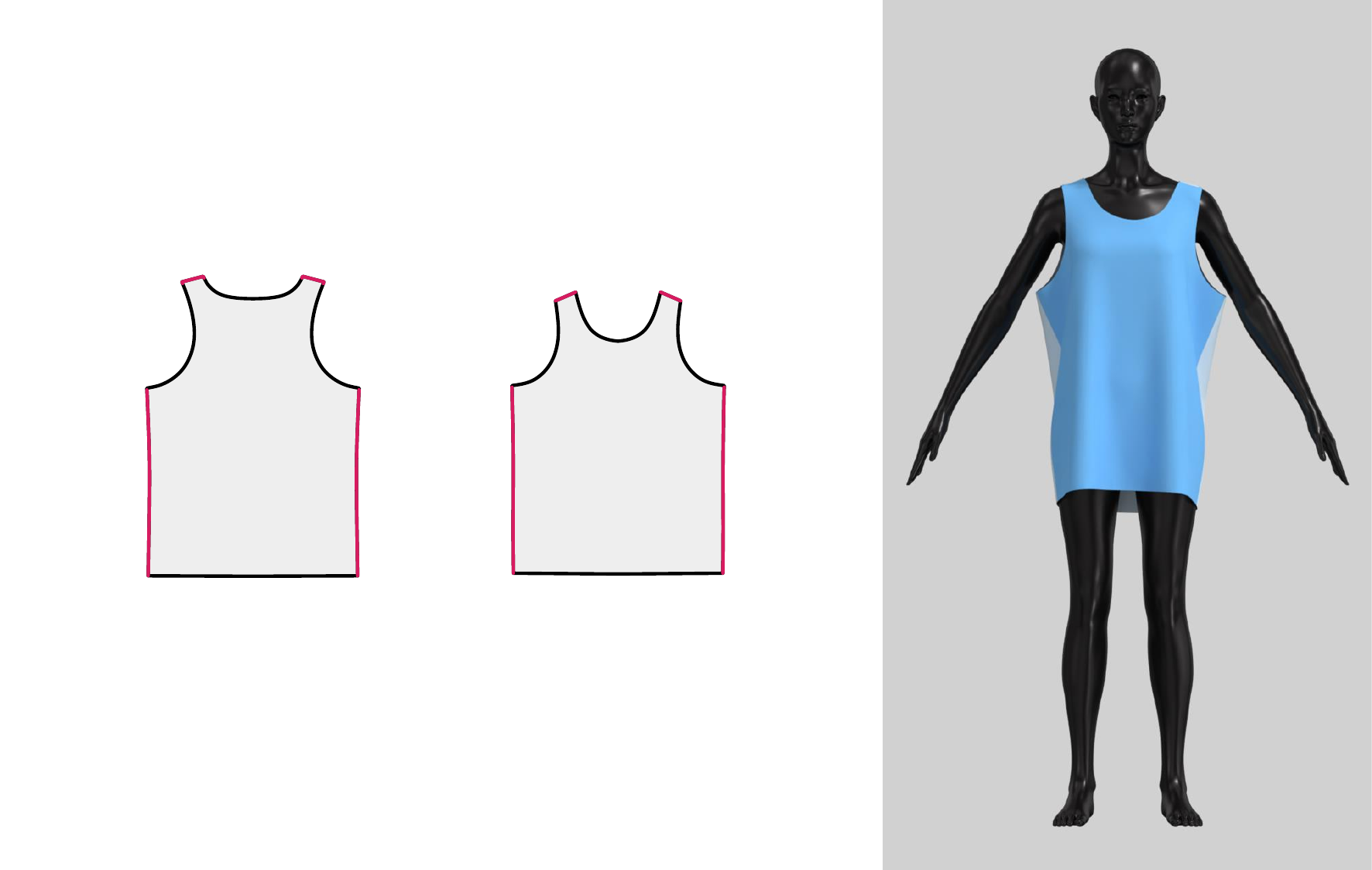}}
    \subfigure[]{\includegraphics[width=0.32\linewidth]{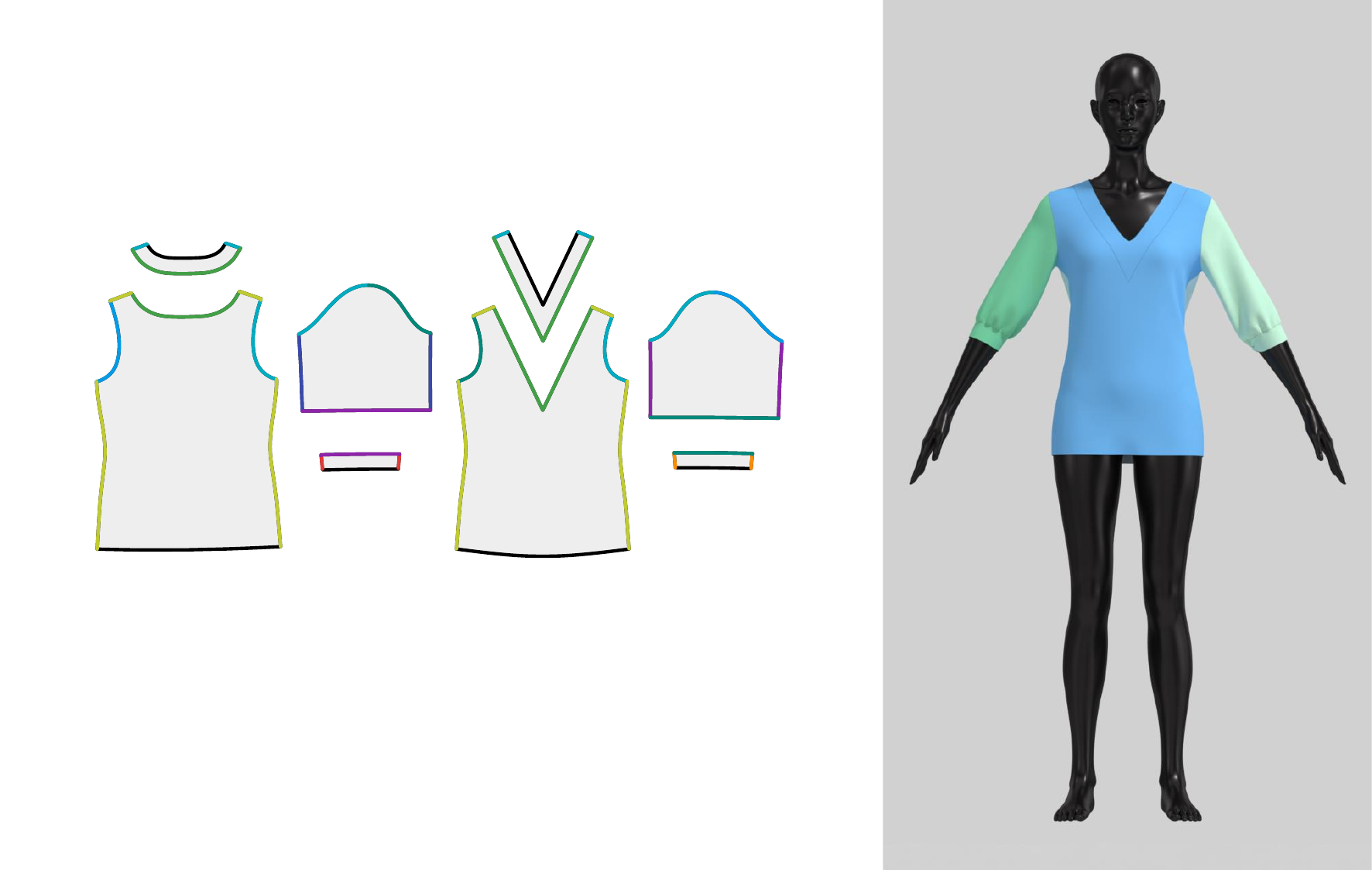}}
    \subfigure[]{\includegraphics[width=0.32\linewidth]{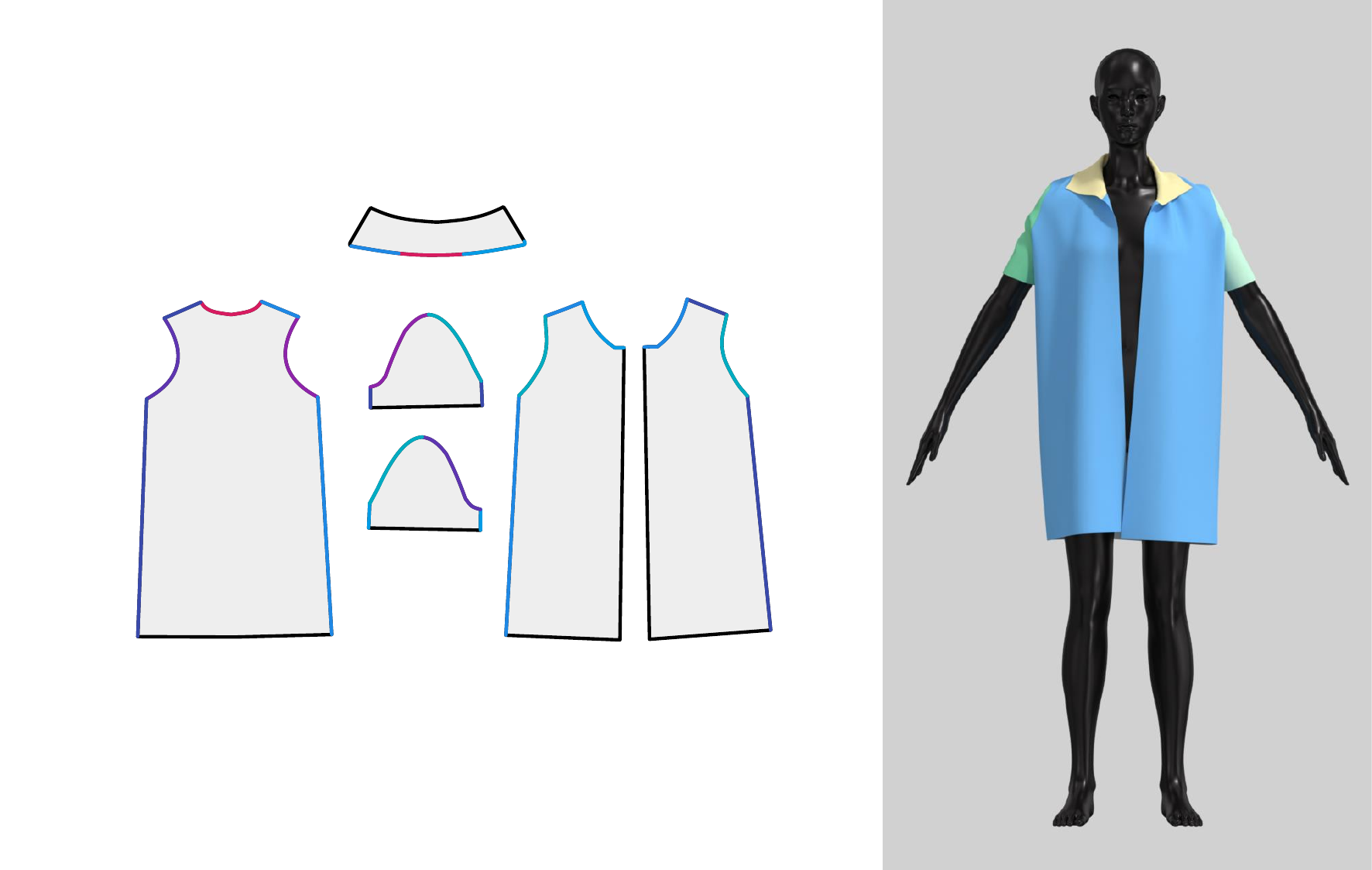}}

    \vspace{-0.1in}

    \subfigure[]{\includegraphics[width=0.32\linewidth]{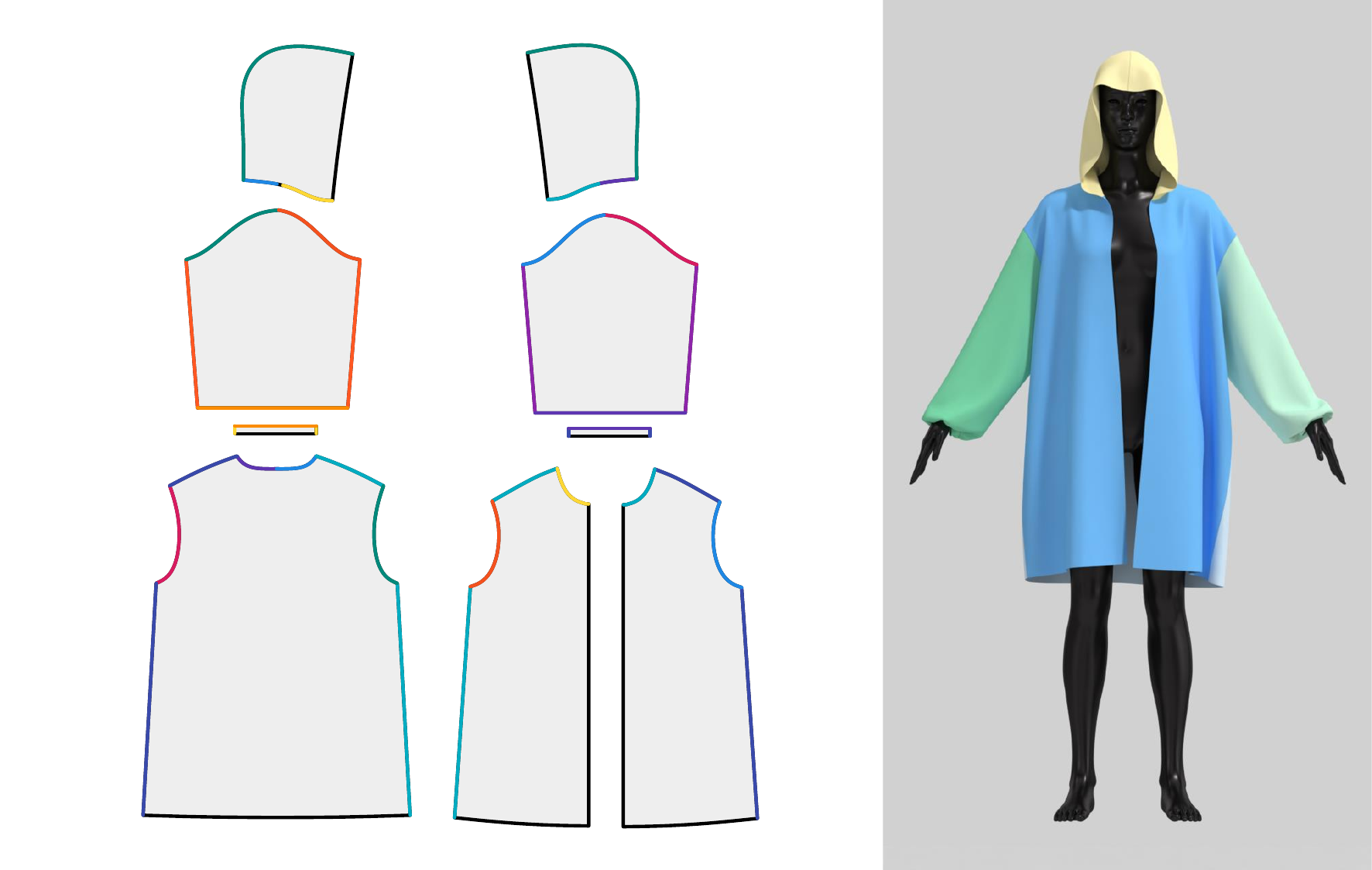}}
    \subfigure[]{\includegraphics[width=0.32\linewidth]{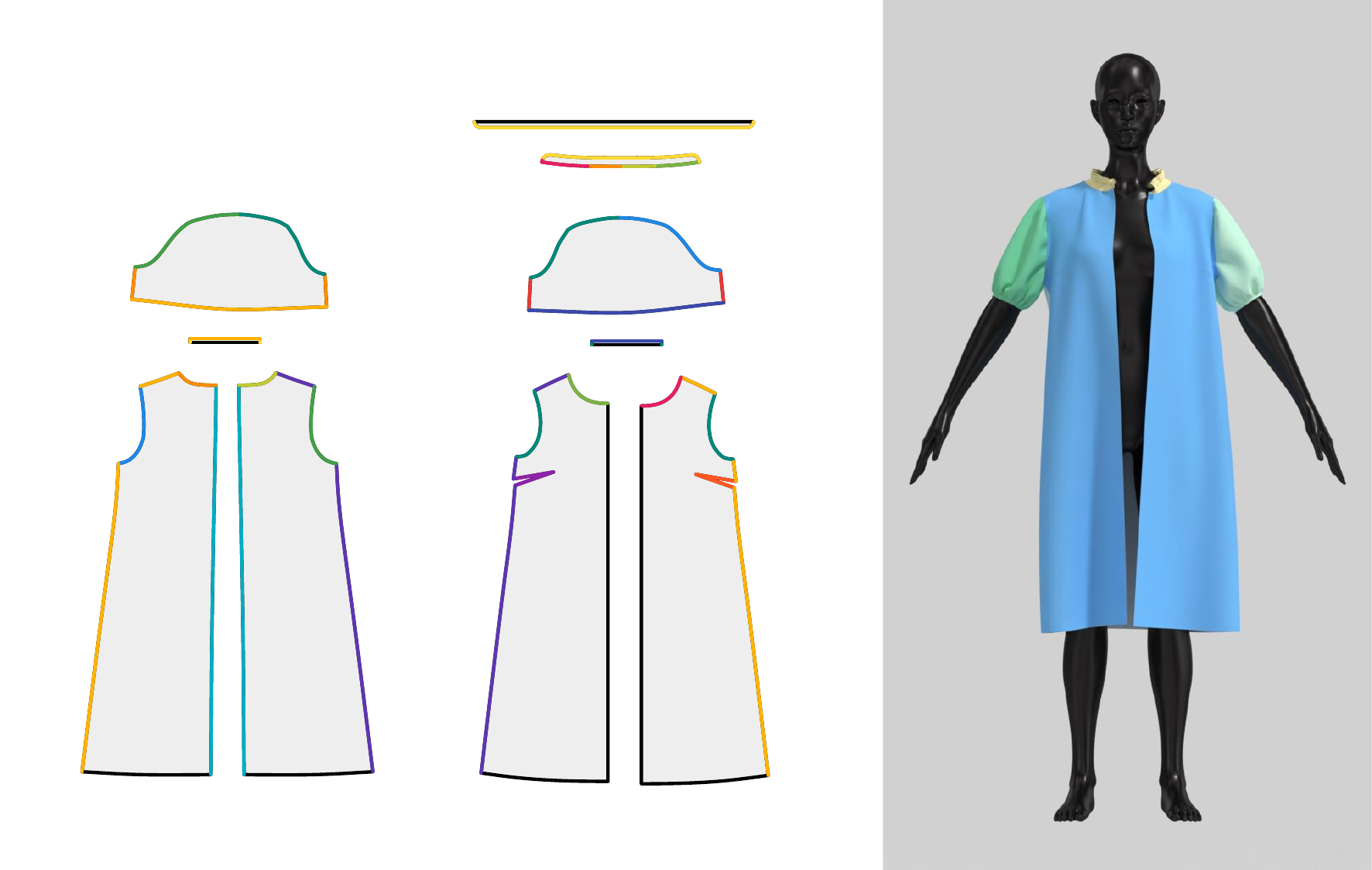}}
    \subfigure[]{\includegraphics[width=0.32\linewidth]{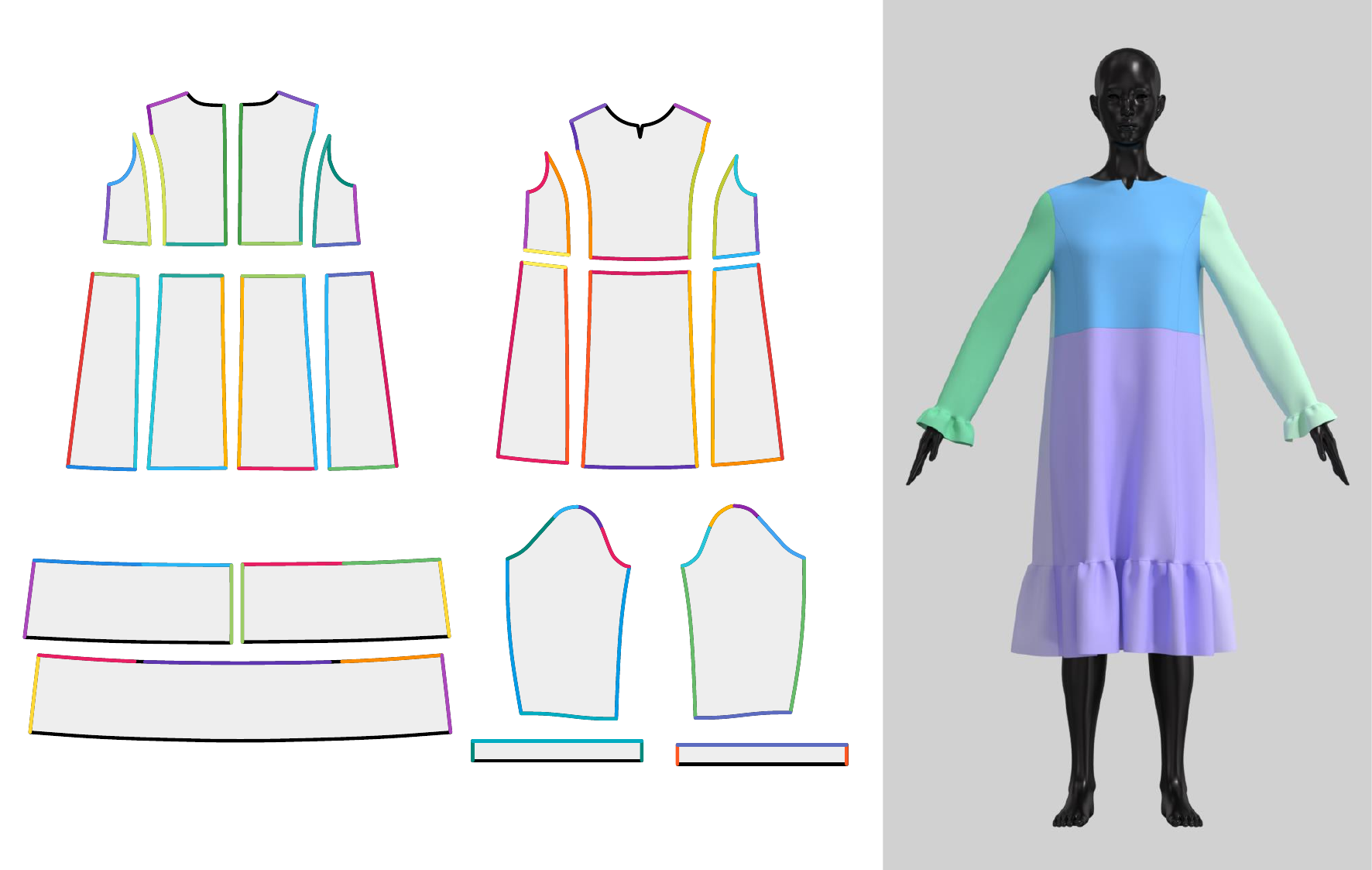}}

    \vspace{-0.1in}

    \subfigure[]{\includegraphics[width=0.32\linewidth]{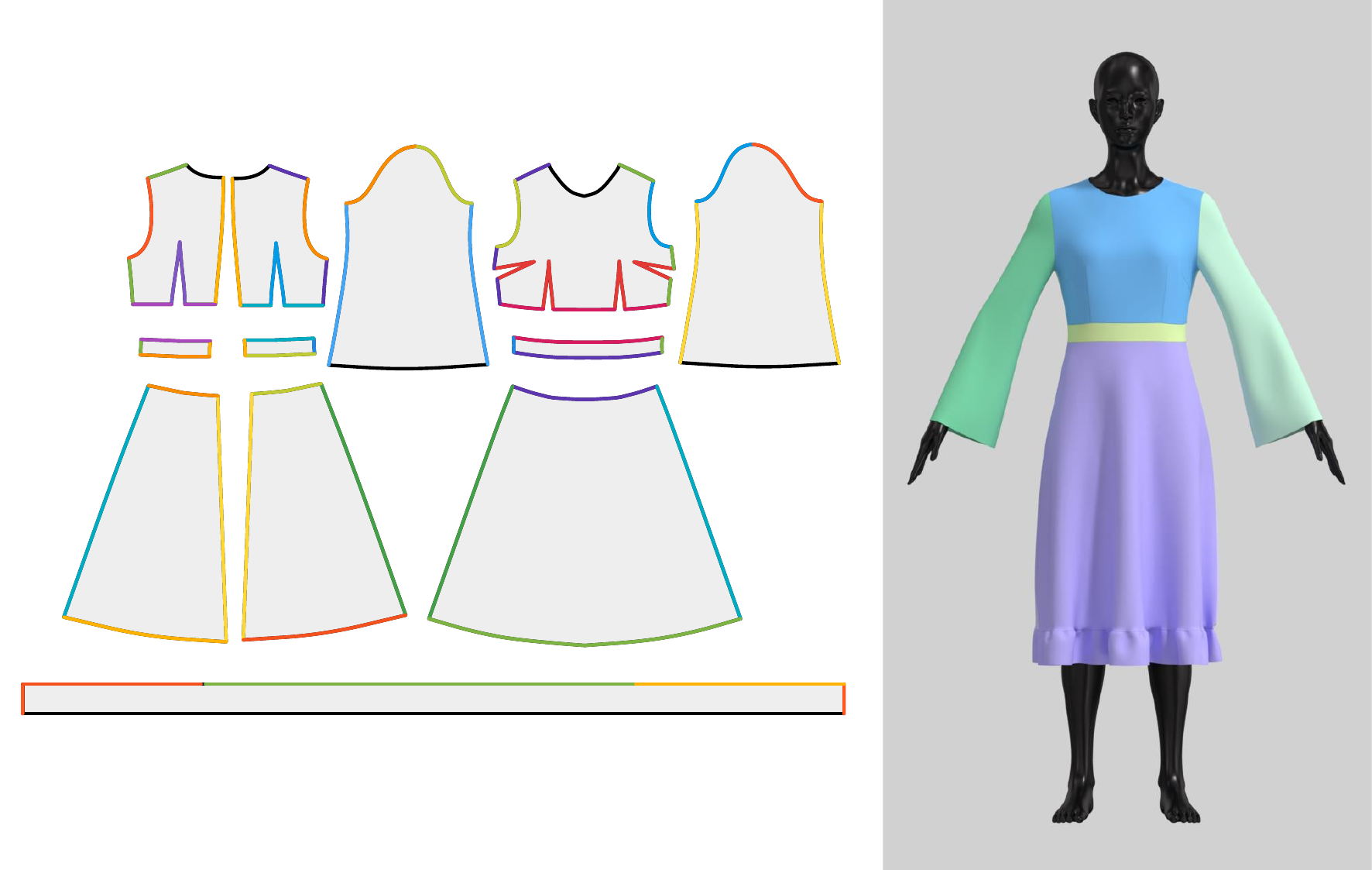}}
    \subfigure[]{\includegraphics[width=0.32\linewidth]{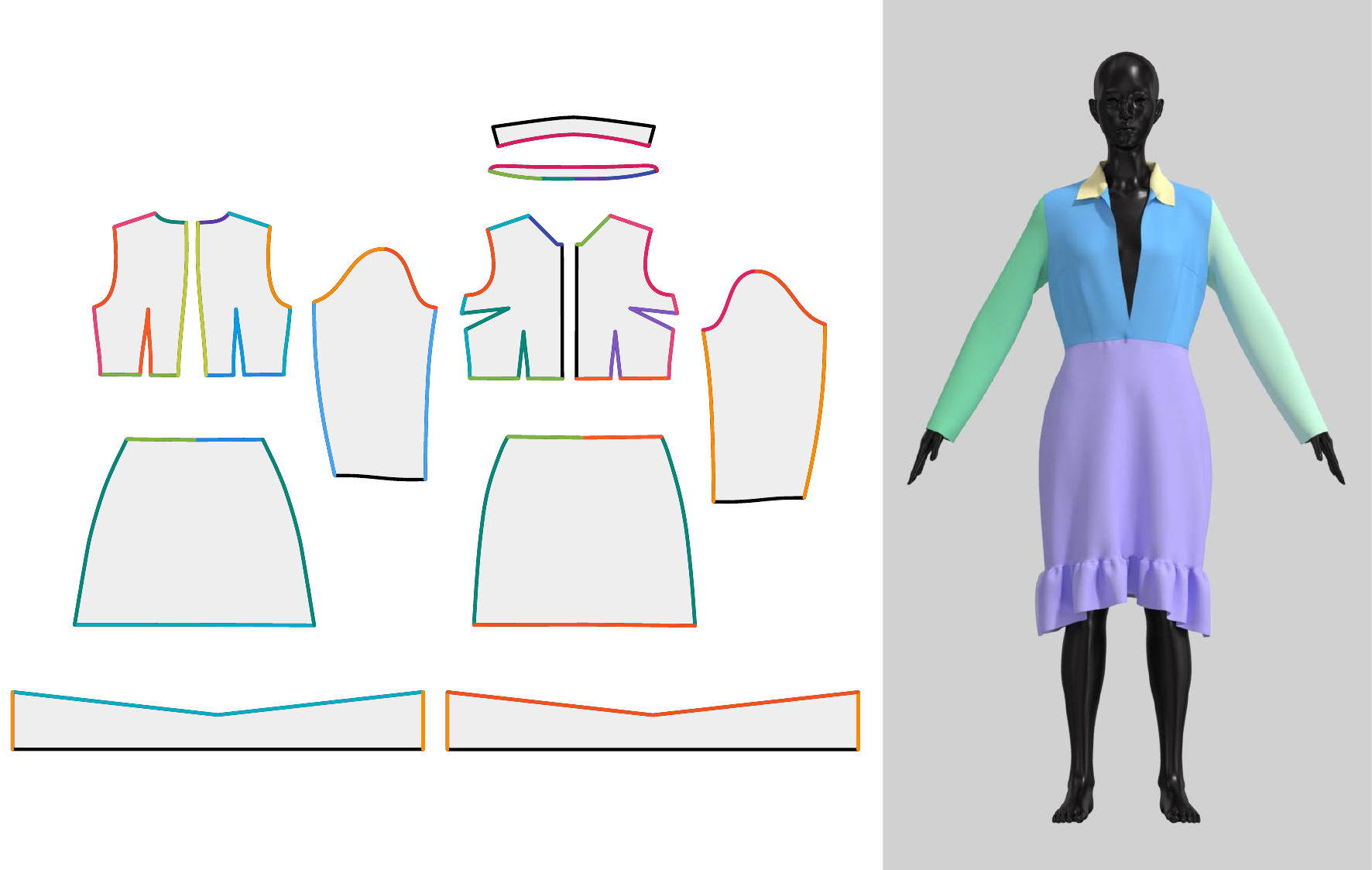}}
    \subfigure[]{\includegraphics[width=0.32\linewidth]{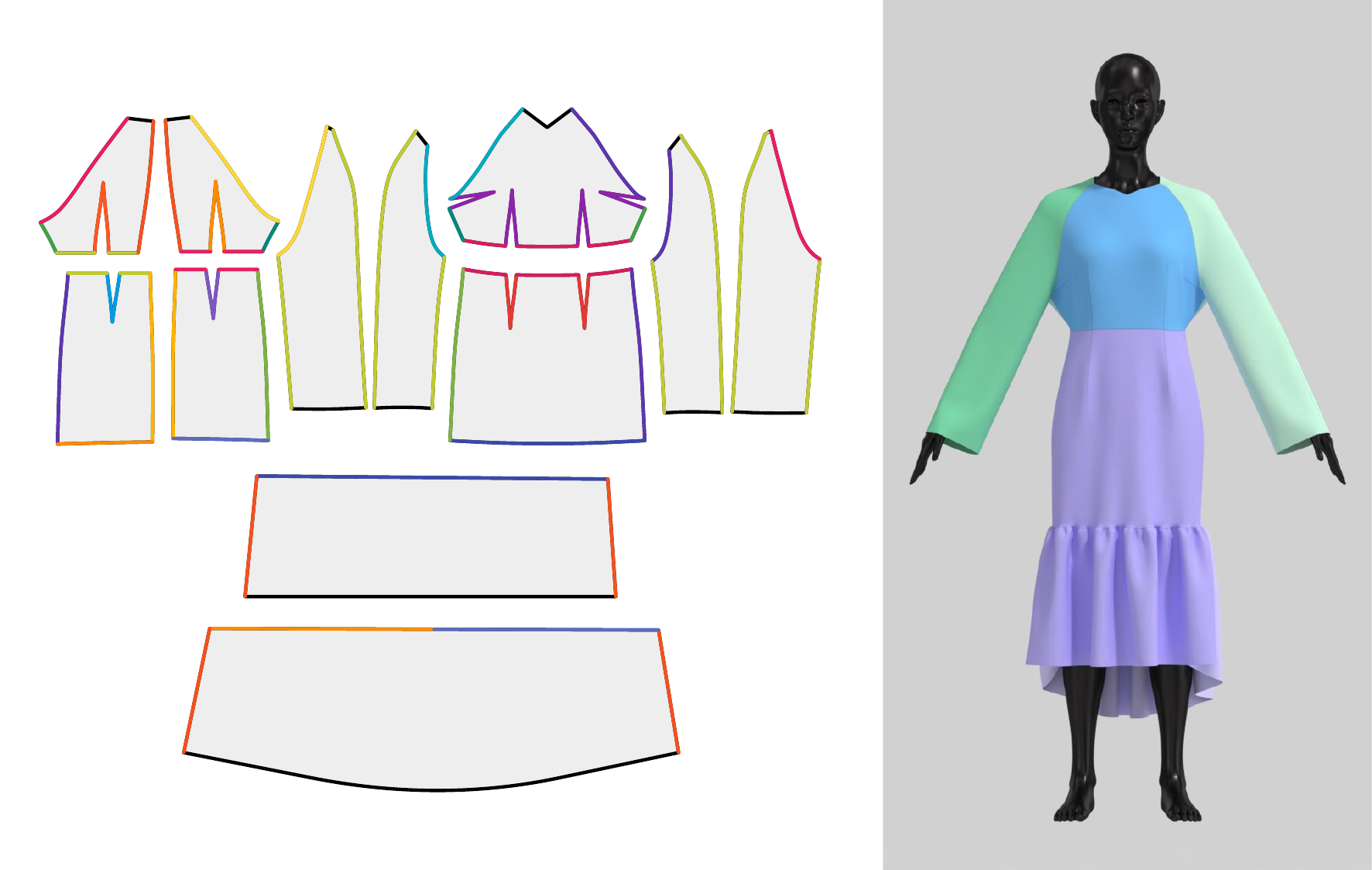}}

    \vspace{-0.1in}

    \caption{Additional examples. For each example, we visualize the input sewing pattern with predicted seam correspondences indicated by segment color matching, and the reconstructed 3D garment with panel semantics represented by panel-wise color coding.}

    \label{fig:good-examples}
\end{figure*}

We evaluate our method on an test set comprising unseen garment styles. As shown in \autoref{fig:good-examples} and supplementary material, the model generalizes well to these novel designs, accurately recovering seam correspondences despite the absence of similar patterns in the training data, demonstrating the robustness of the our graph-based framework and its ability to capture fundamental garment construction principles beyond the training distribution.

\subsection{Conclusion}
In conclusion, we propose a graph-based, two-level learning framework for automatically inferring seam correspondences in sewing patterns from 2D geometry. By combining CNN- and U-Net–based panel feature extractors with an attention-based GNN for efficient message passing, the method jointly predicts panel semantics, panel connectivity, and fine-grained seam correspondences. This design enables robust handling of complex garment topologies, including curved seams and many-to-one relations, while exhibiting strong generalization across diverse garment styles. Experiments on a curated sewing-pattern dataset demonstrate significant improvements in accuracy and generalization. The proposed framework has the potential to streamline digital garment design workflows and support rapid prototyping in fashion applications.

\section{Limitations and Future Work}
While the proposed method demonstrates robust performance across a wide range of garment styles, several limitations remain. First, its reliance on accurate panel semantic classification can lead to suboptimal seam predictions when semantic labels are misclassified. Second, the current formulation supports only five common stitch types (see \autoref{fig:stitch-types}), leaving more complex configurations—such as zigzag self-stitches and interior-attached seams—for future investigation. Addressing these issues may benefit from more robust semantic inference or explicit uncertainty modeling. In addition, the dataset covers a limited variety of garment categories; expanding it to include more complex designs would further improve generalization. Finally, incorporating user-in-the-loop feedback could enable interactive refinement of seam predictions, enhancing practicality in real-world design workflows.

\begin{acks}
We thank reviewers for valuable comments. This work was supported by Key R\&D Program of Zhejiang (No. 2024C01069).
\end{acks}


\bibliographystyle{ACM-Reference-Format}
\bibliography{reb}

\clearpage
\newpage

\clearpage
\newpage

\clearpage
\newpage

\begin{figure*}[t]
    \centering

    \subfigure[]{\includegraphics[width=0.32\linewidth]{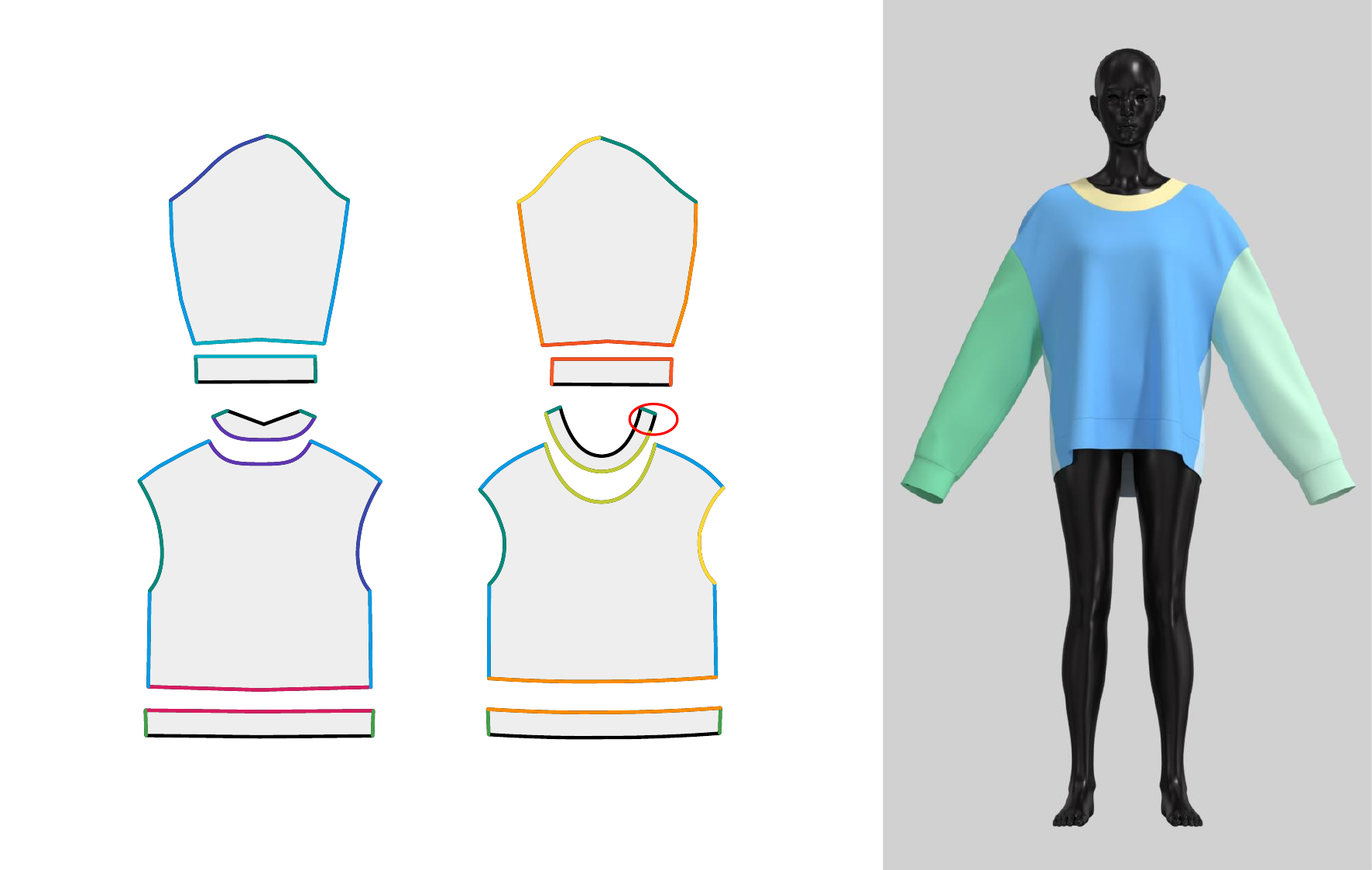}}
    \subfigure[]{\includegraphics[width=0.32\linewidth]{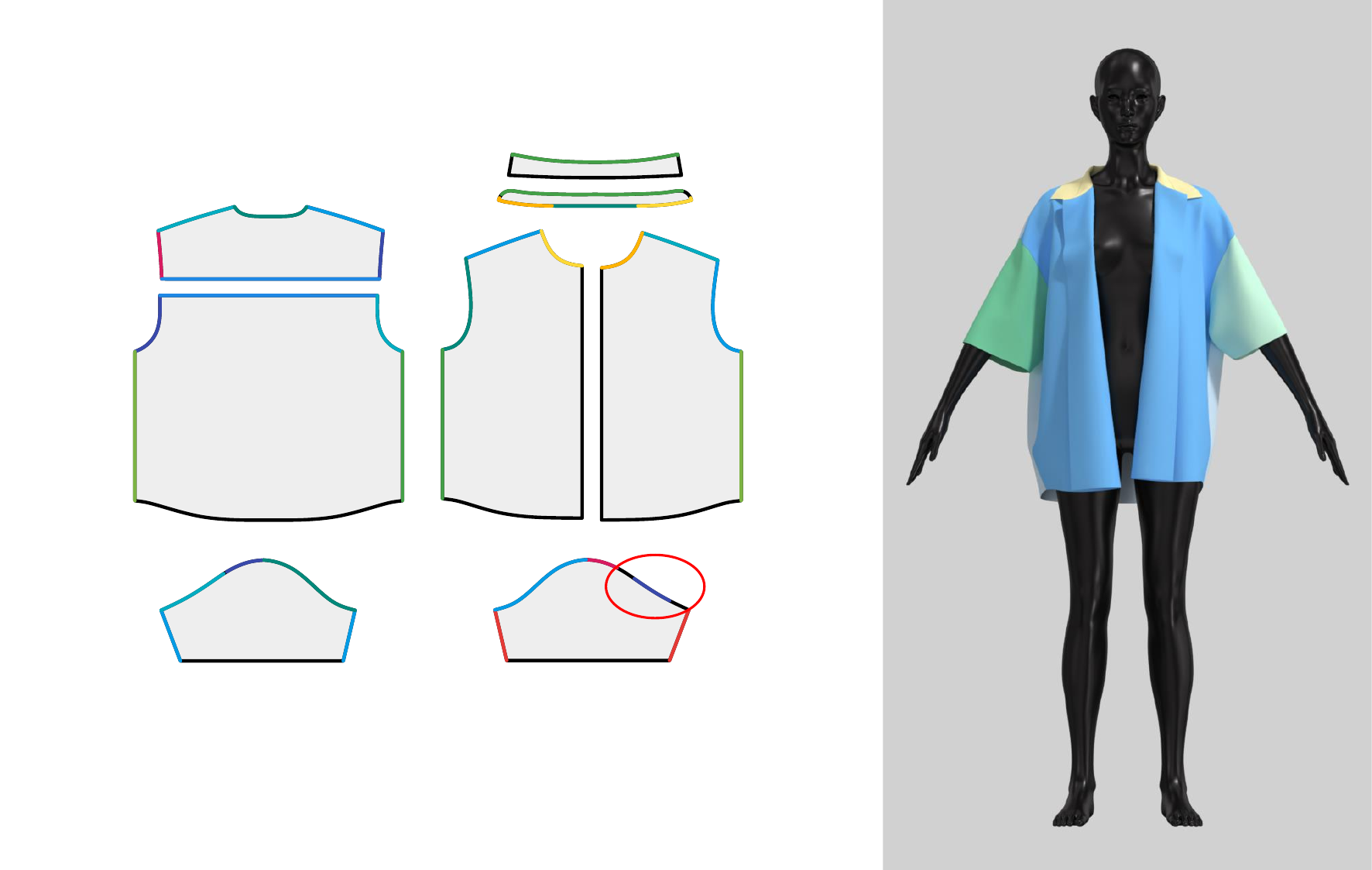}}
    \subfigure[]{\includegraphics[width=0.32\linewidth]{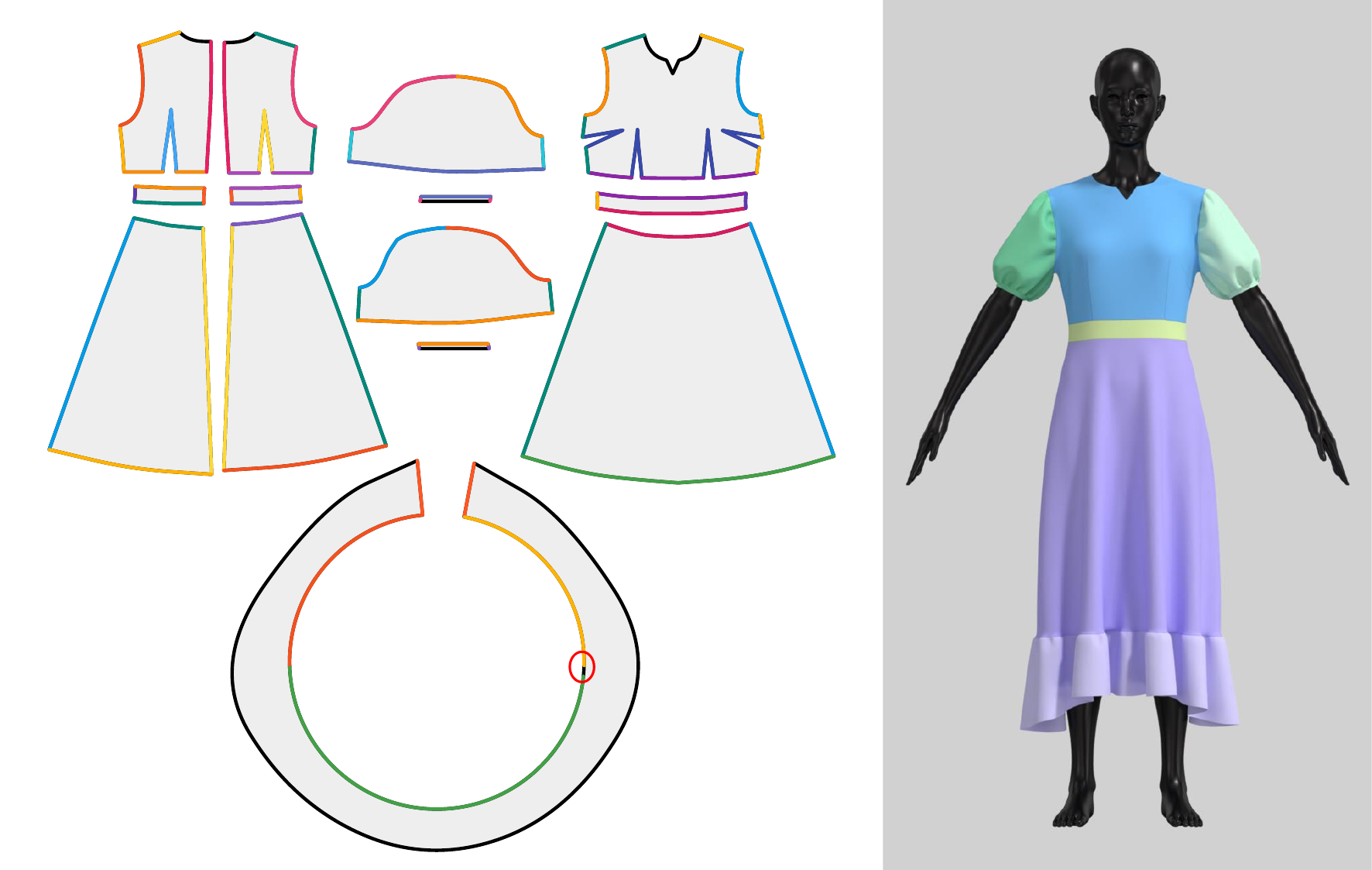}}
    
    \vspace{-0.1in}

    \subfigure[]{\includegraphics[width=0.32\linewidth]{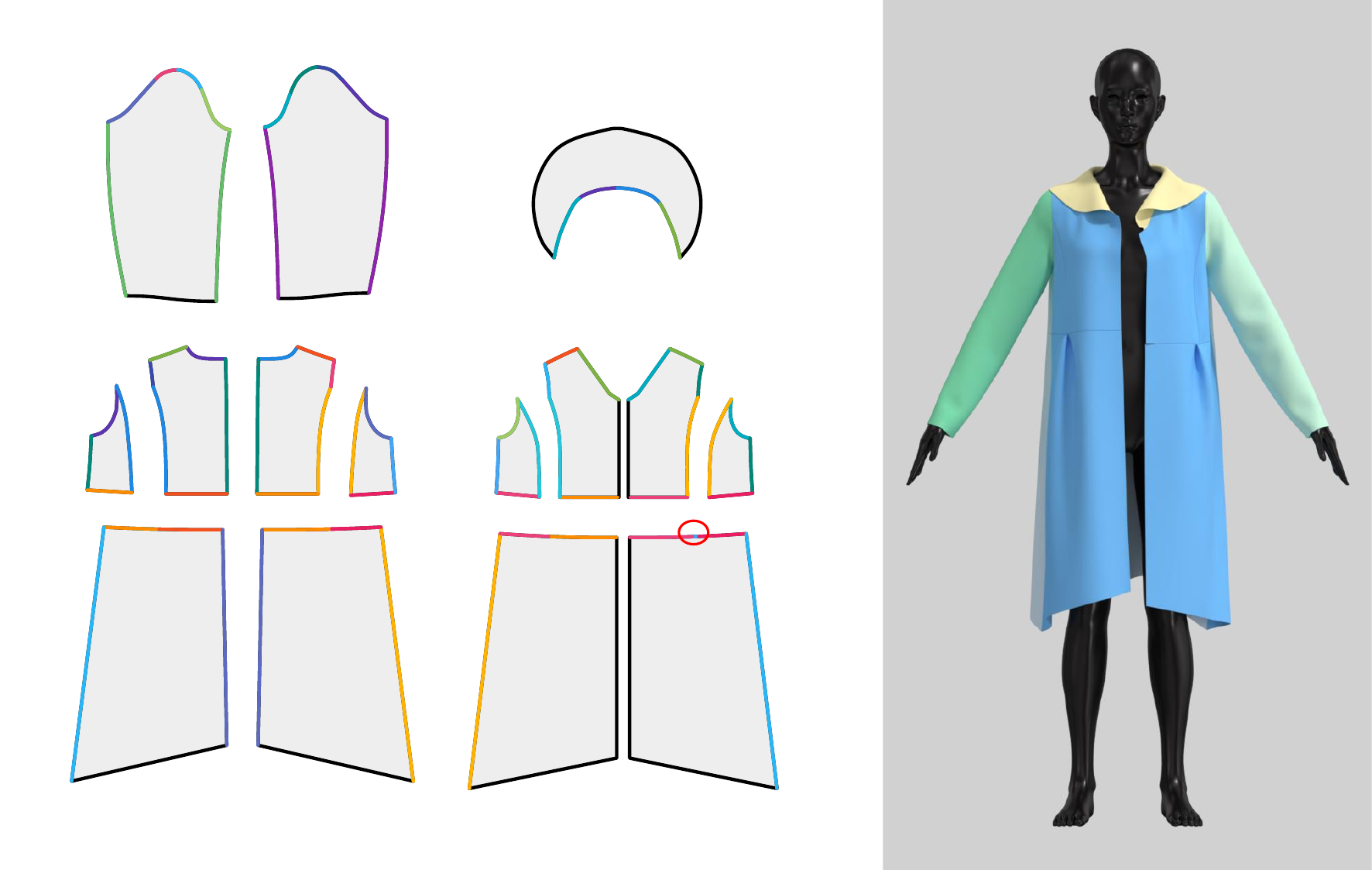}}
    \subfigure[]{\includegraphics[width=0.32\linewidth]{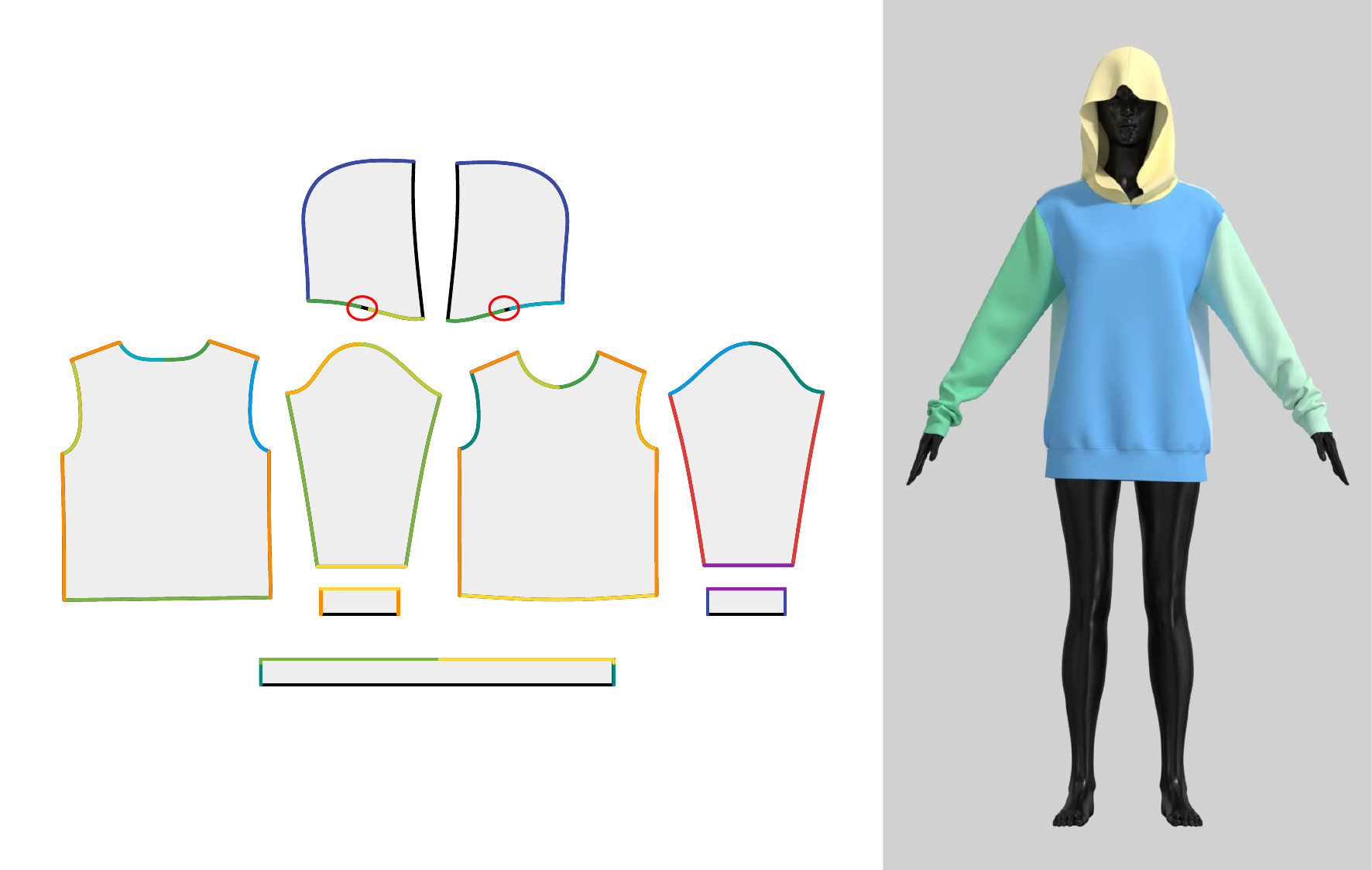}}
    \subfigure[]{\includegraphics[width=0.32\linewidth]{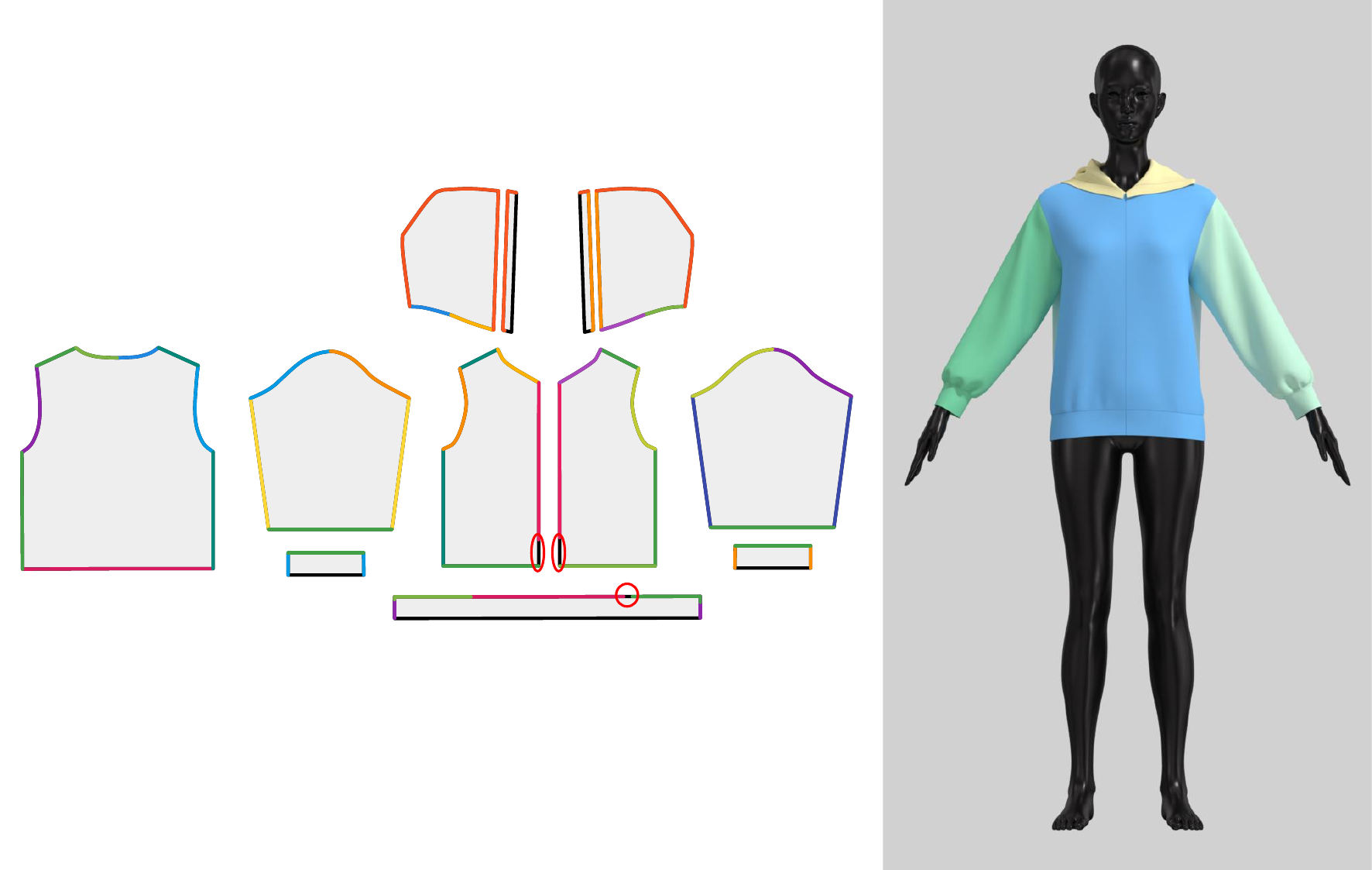}}

    \vspace{-0.1in}
    
    \subfigure[]{\includegraphics[width=0.32\linewidth]{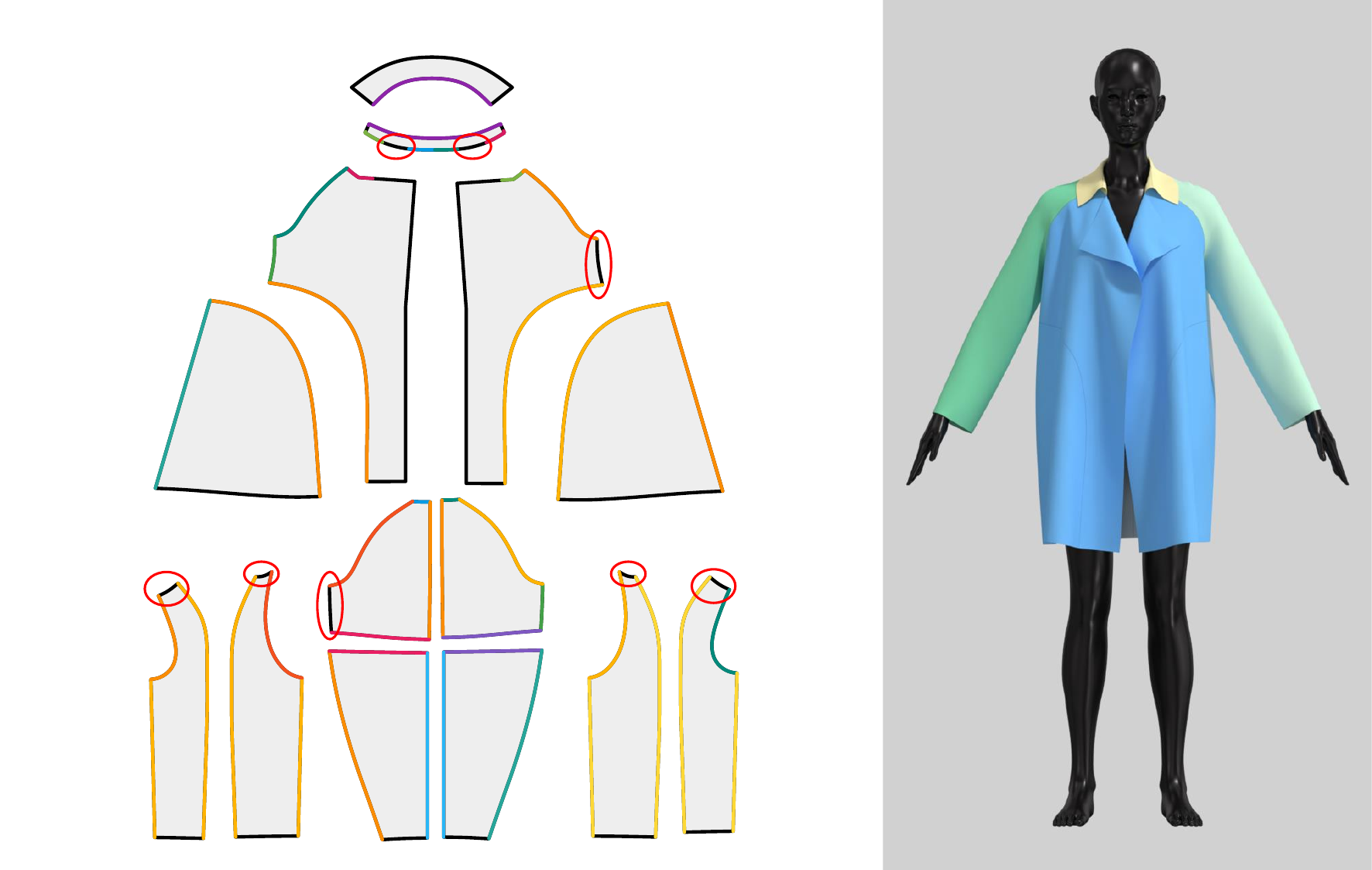}}
    \subfigure[]{\includegraphics[width=0.32\linewidth]{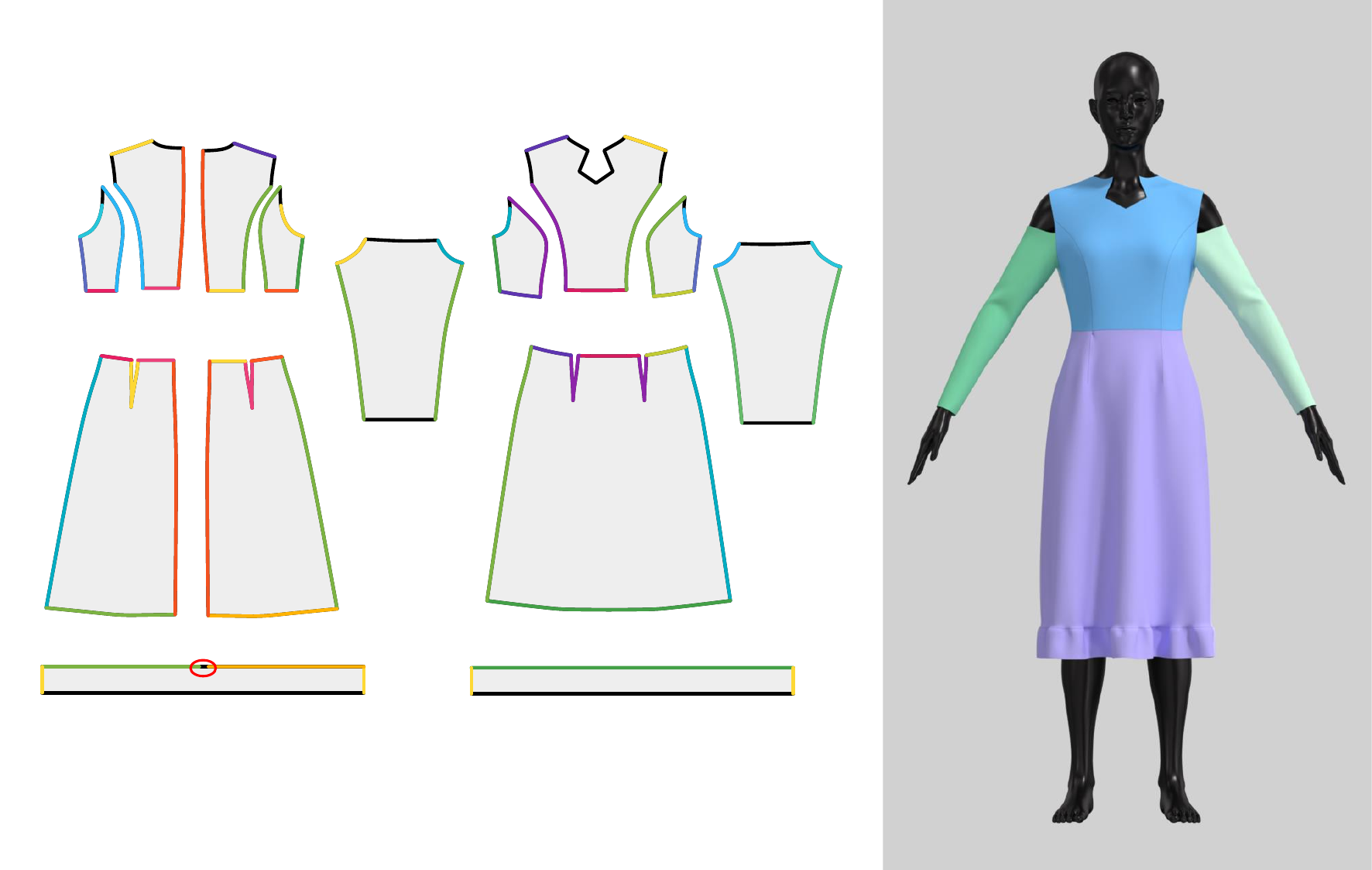}}
    \subfigure[]{\includegraphics[width=0.32\linewidth]{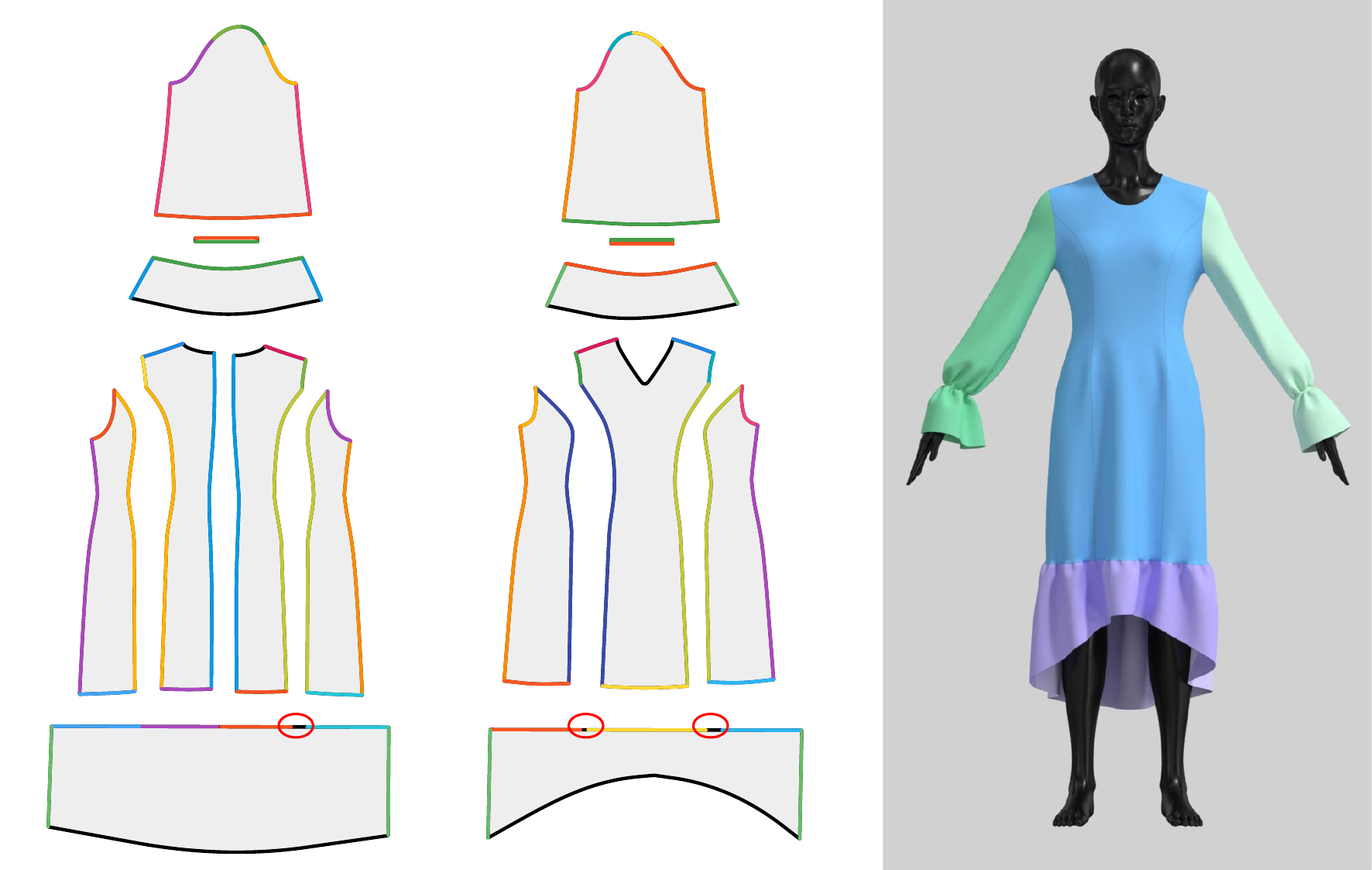}}

    \vspace{-0.1in}

    \subfigure[]{\includegraphics[width=0.32\linewidth]{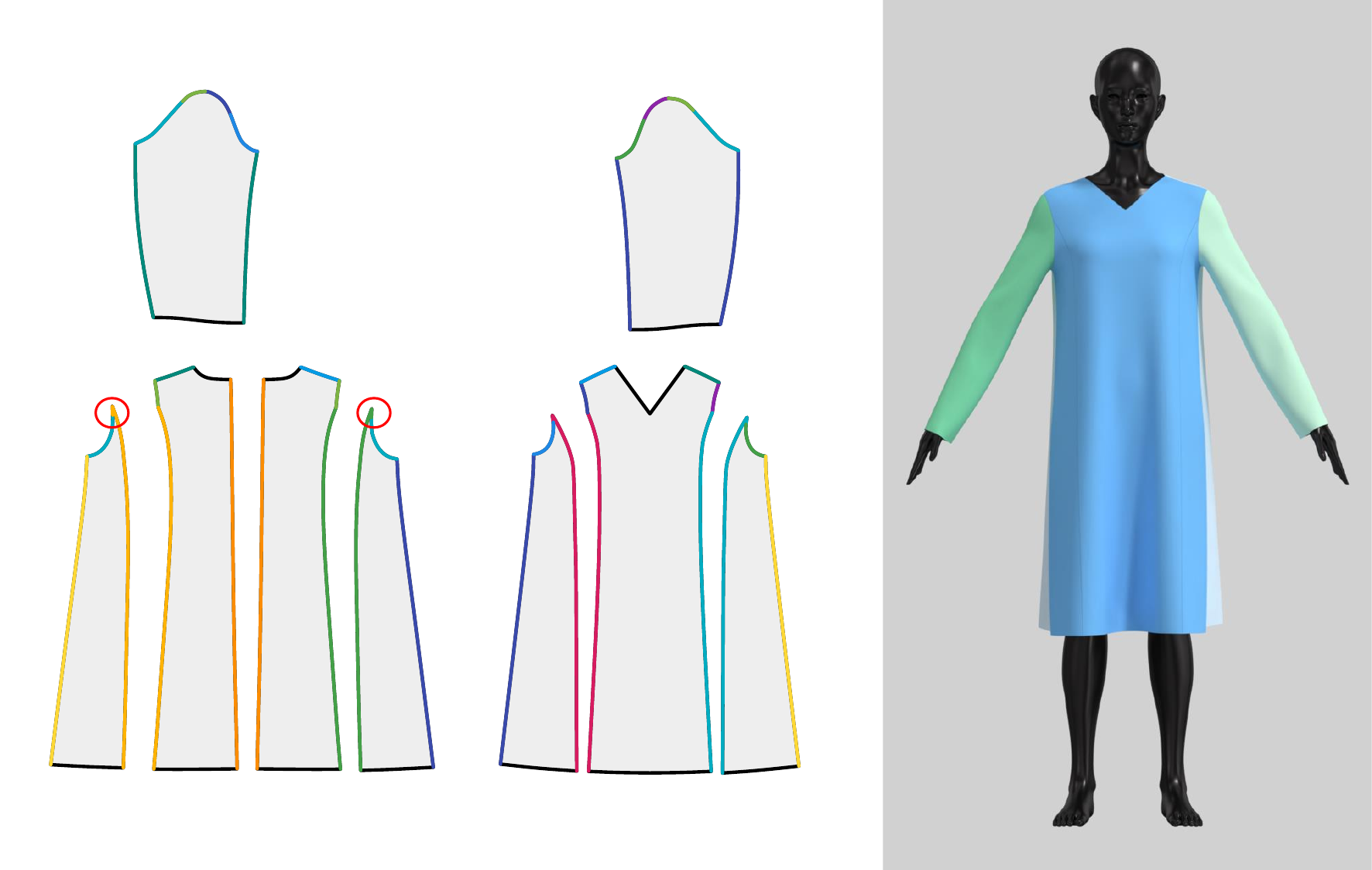}}
    \subfigure[]{\includegraphics[width=0.32\linewidth]{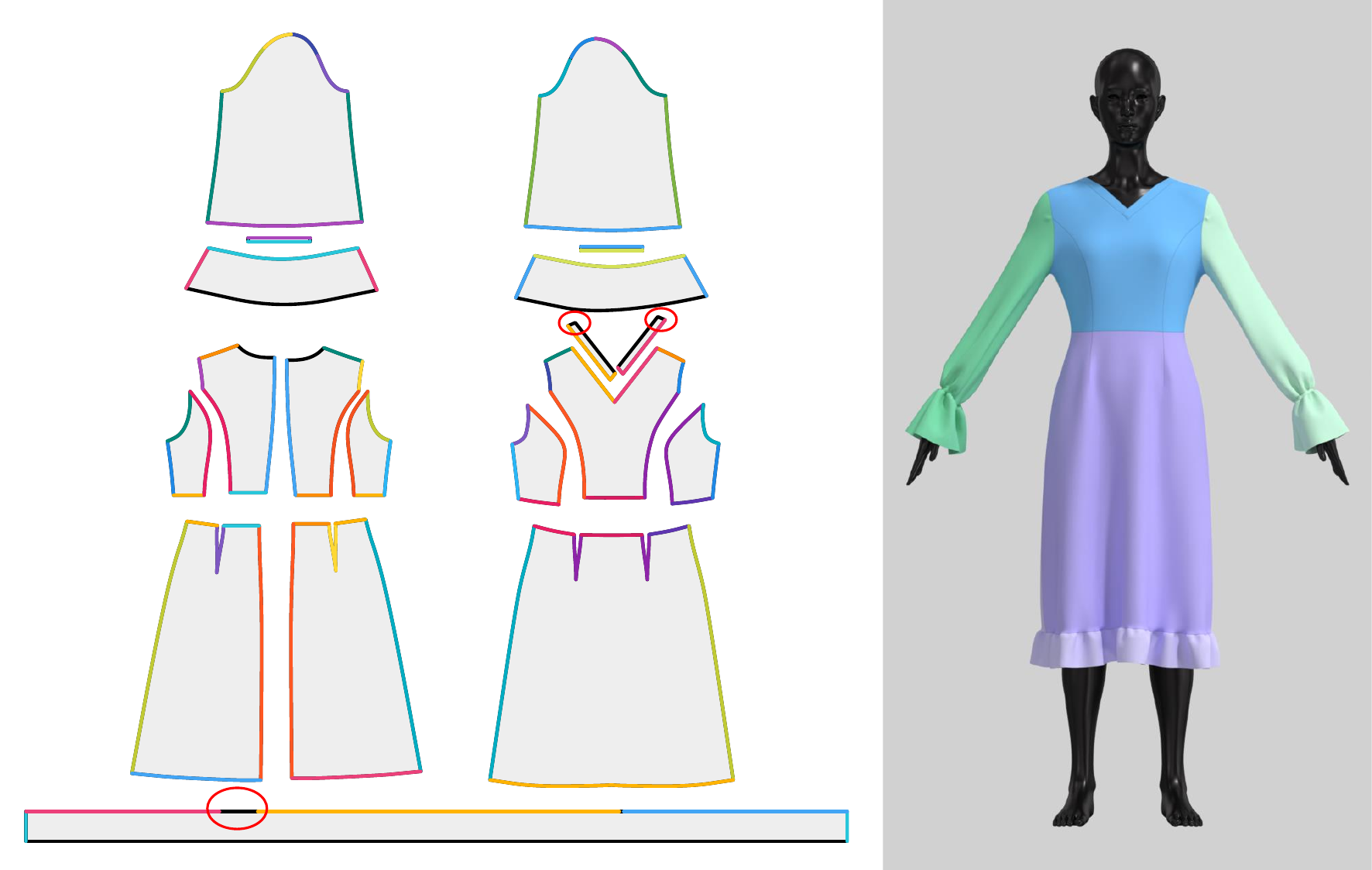}}
    \subfigure[]{\includegraphics[width=0.32\linewidth]{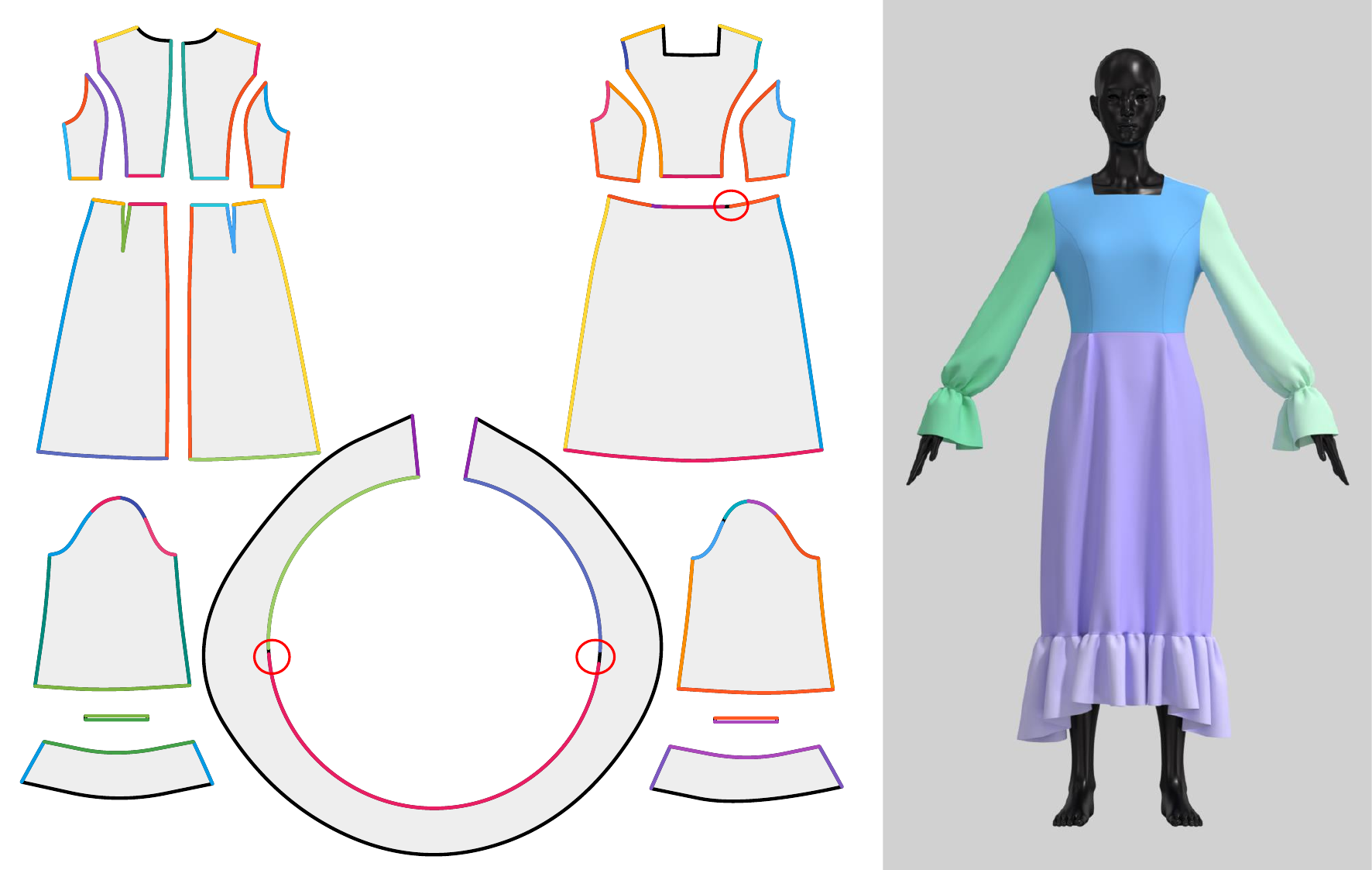}}

    \vspace{-0.1in}

    \subfigure[]{\includegraphics[width=0.32\linewidth]{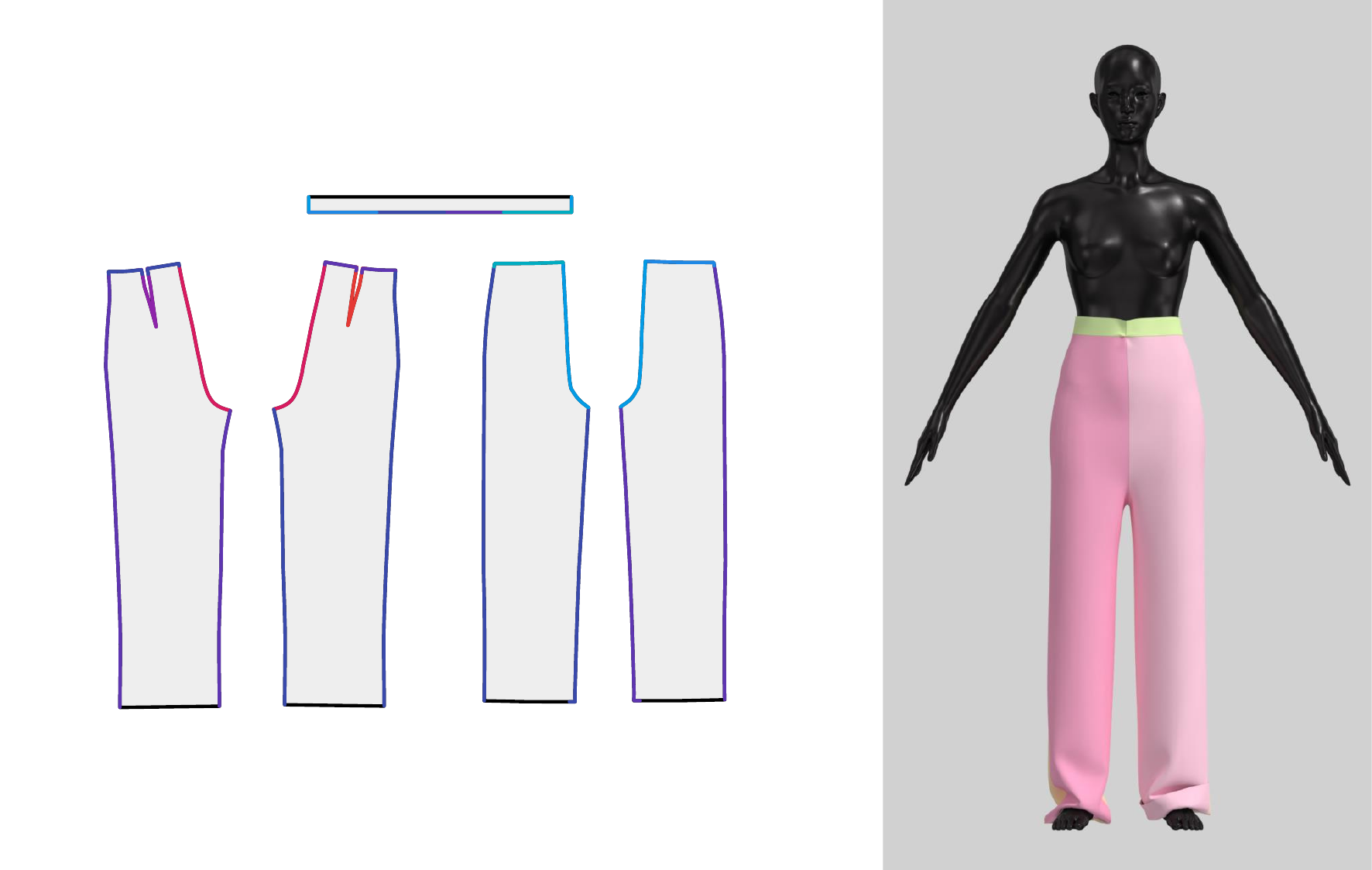}}
    \subfigure[]{\includegraphics[width=0.32\linewidth]{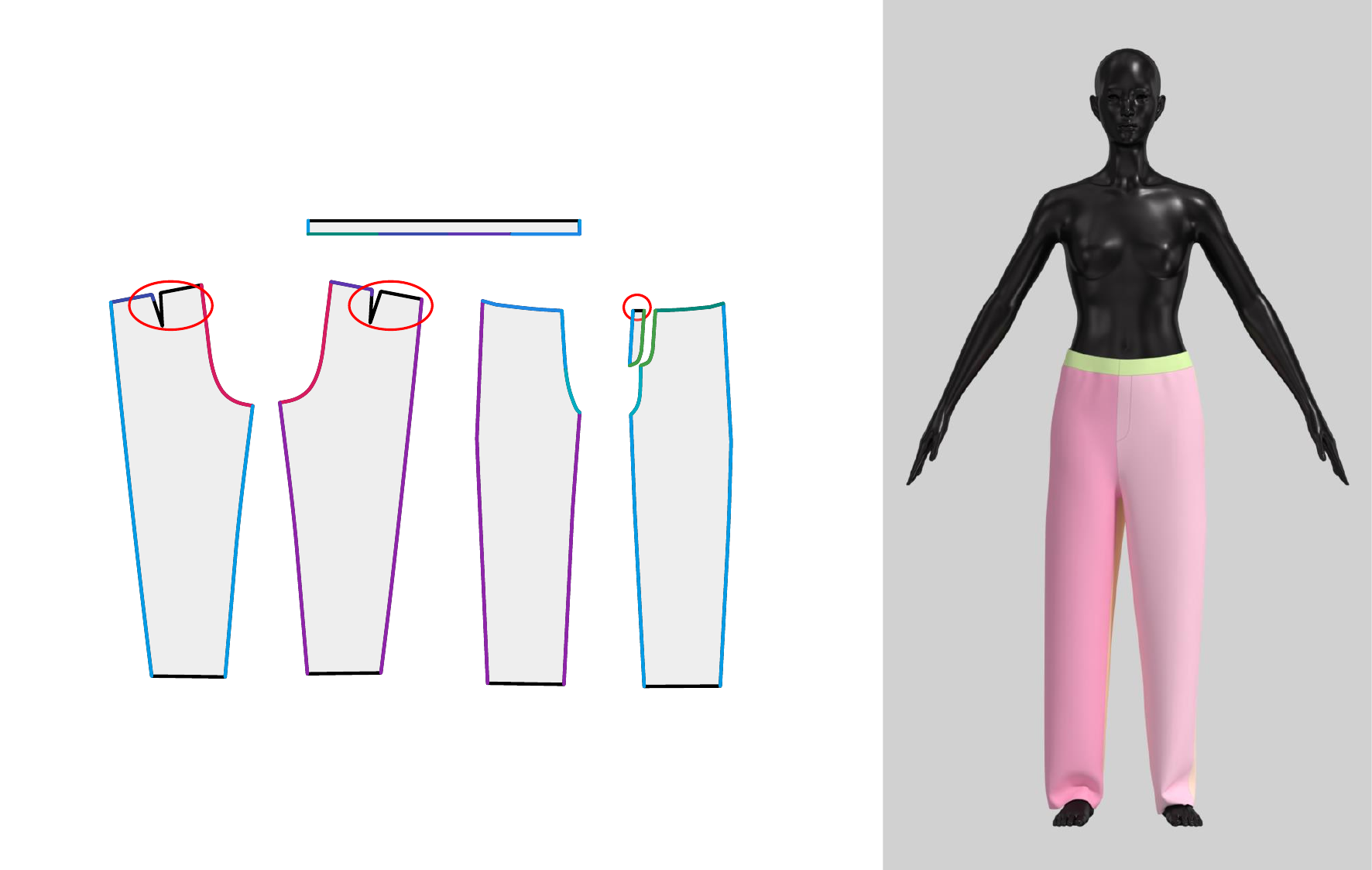}}    
    \subfigure[]{\includegraphics[width=0.32\linewidth]{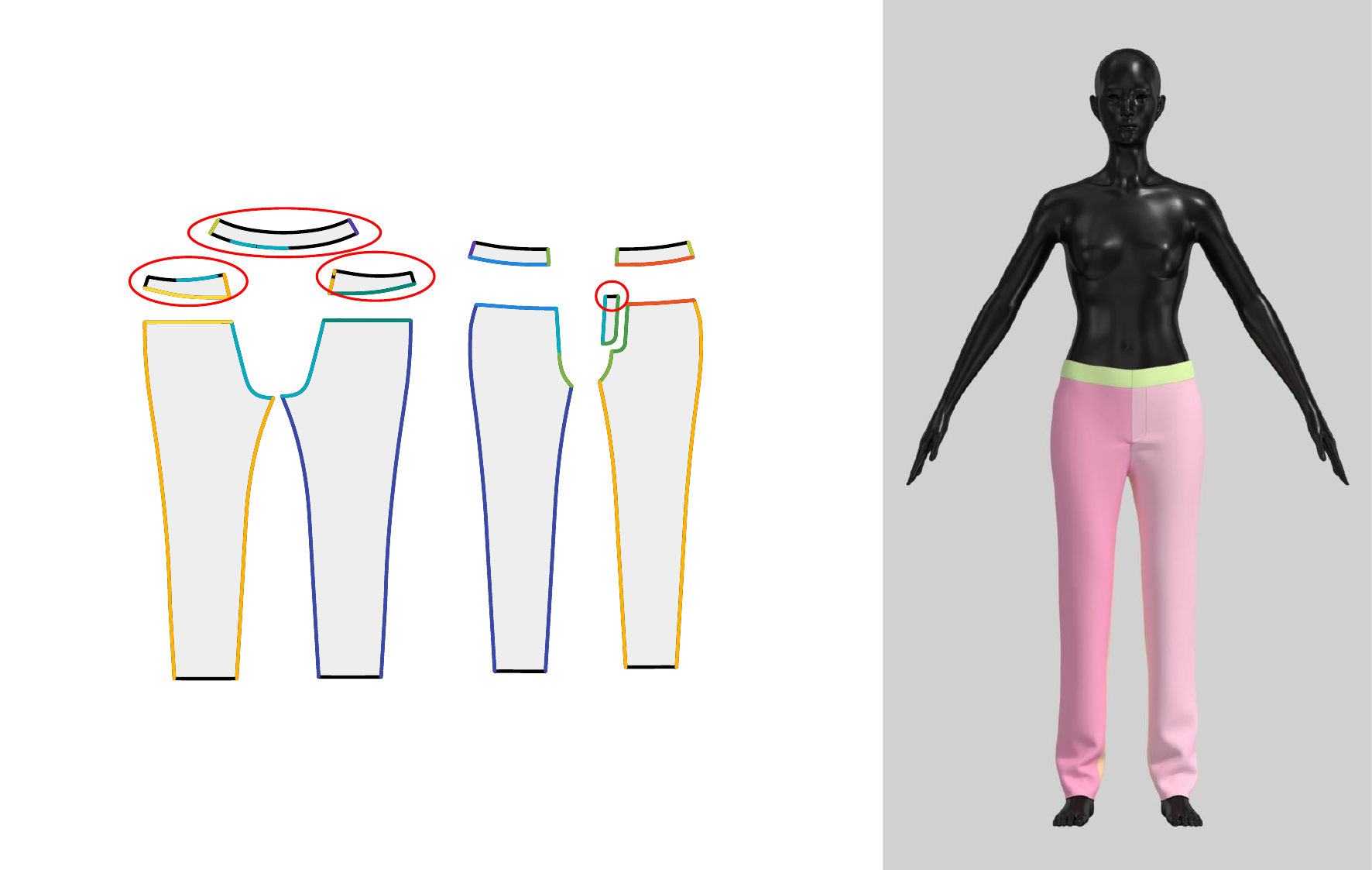}}

    \vspace{-0.1in}

    \caption{Additional examples. For each example, we visualize the input sewing pattern with predicted seam correspondences indicated by segment color matching, and the reconstructed 3D garment with panel semantics represented by panel-wise color coding. Red ellipses indicate minor artifacts.}

    \label{fig:less-good-examples}
\end{figure*}

\clearpage
\newpage

\begin{figure*}[t]
    \centering
    
    \subfigure[]{\includegraphics[width=0.32\linewidth]{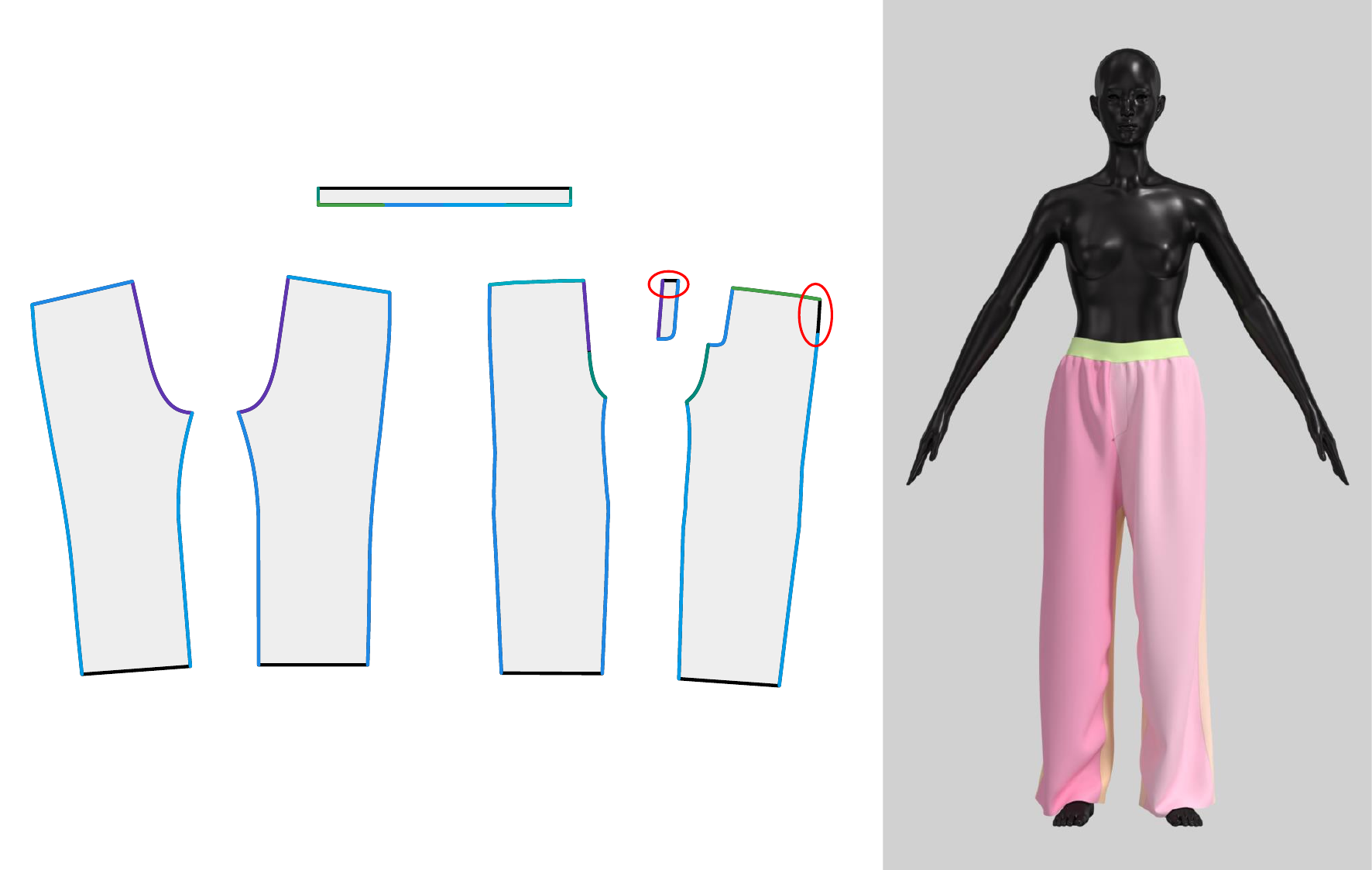}}
    \subfigure[]{\includegraphics[width=0.32\linewidth]{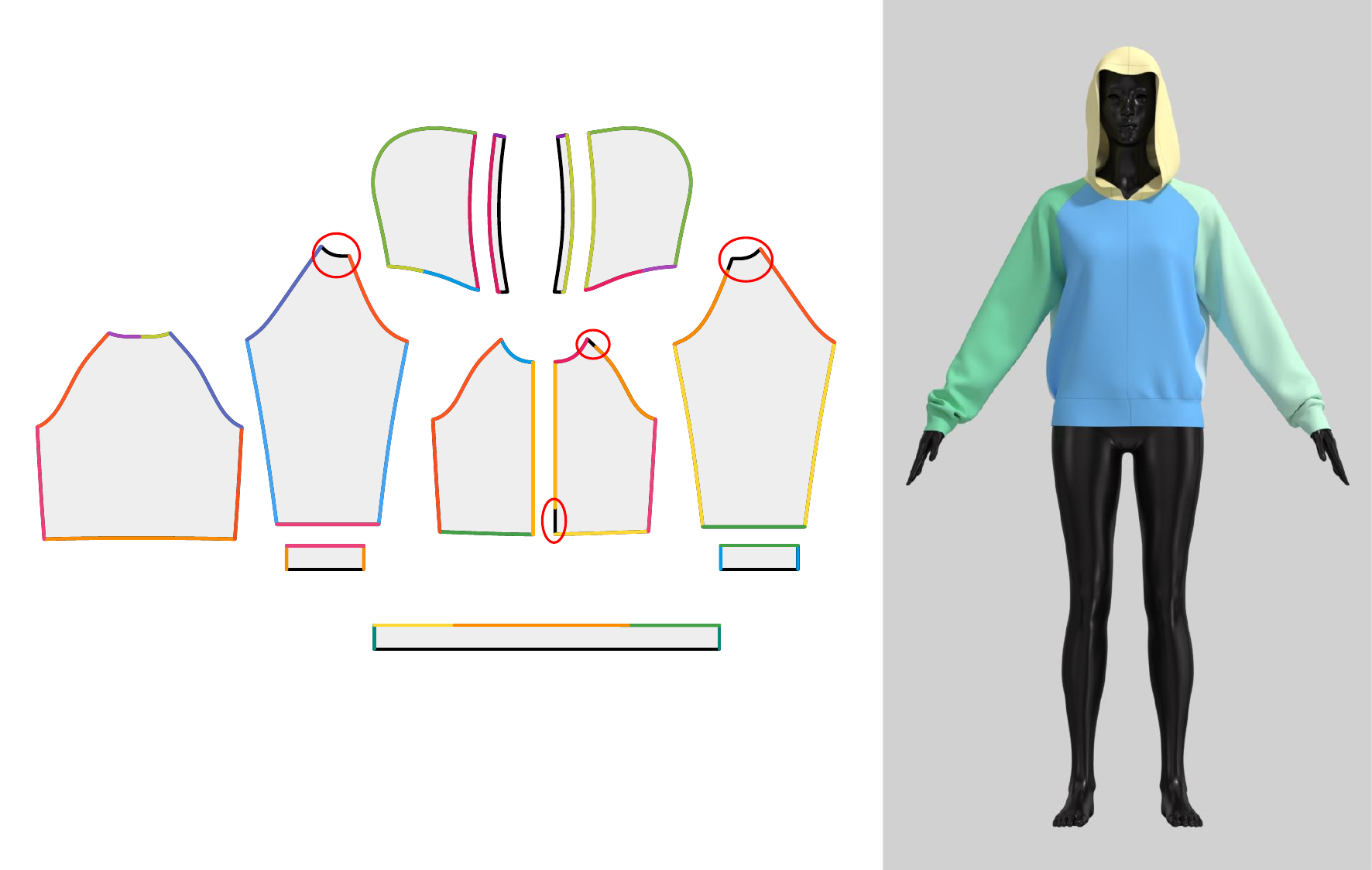}}
    \subfigure[]{\includegraphics[width=0.32\linewidth]{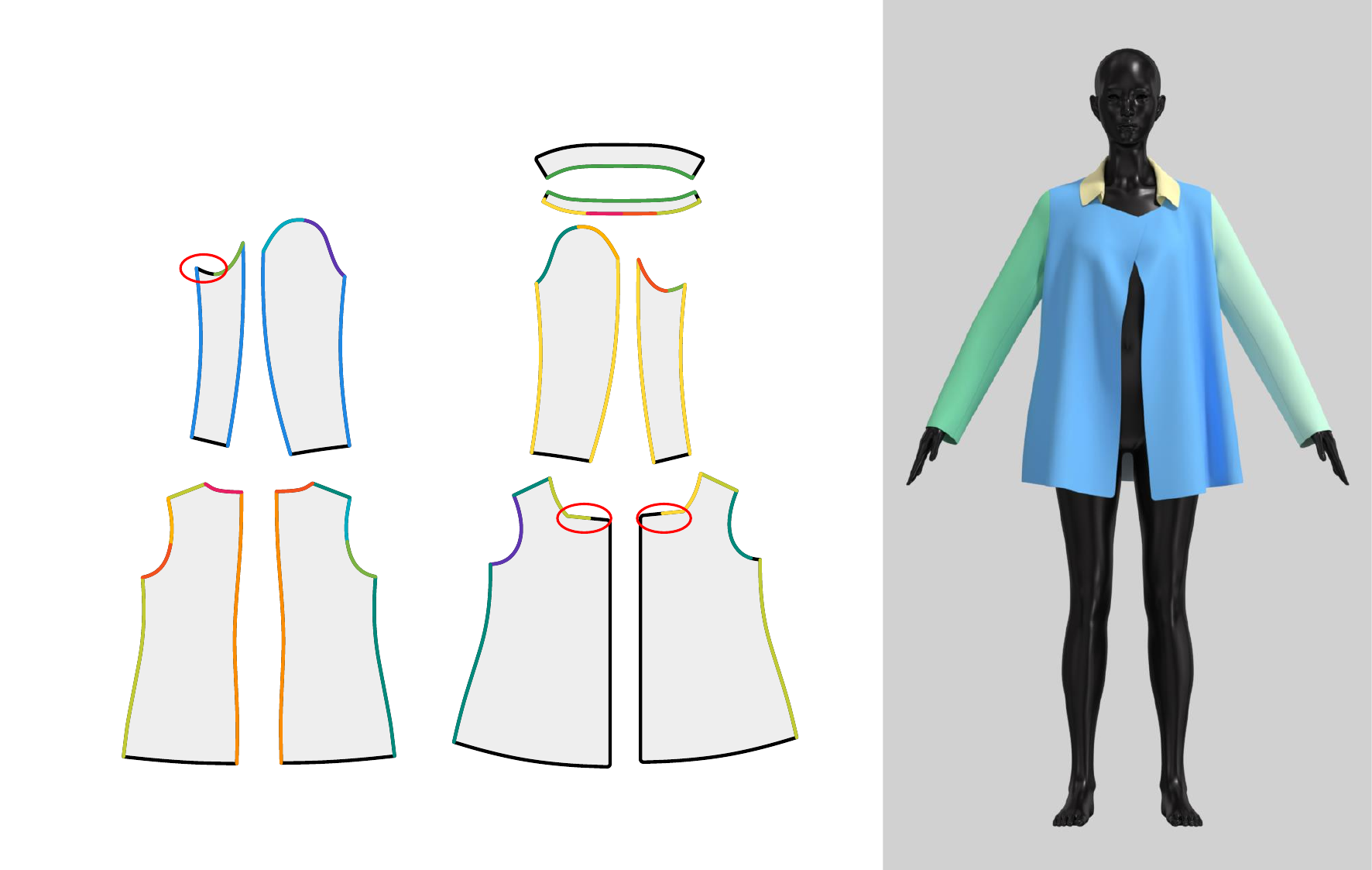}}

    \vspace{-0.1in}

    \subfigure[]{\includegraphics[width=0.32\linewidth]{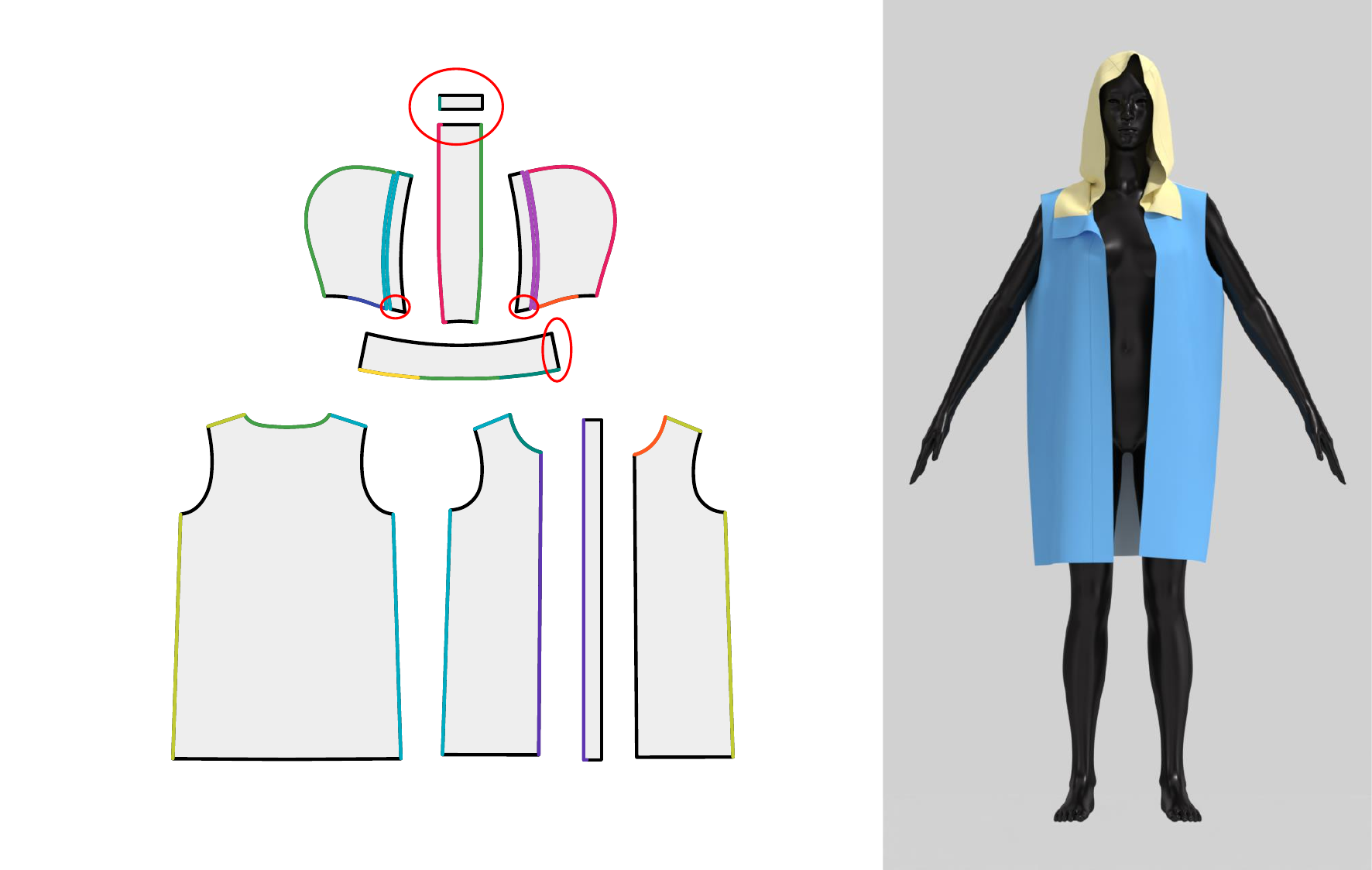}}
    \subfigure[]{\includegraphics[width=0.32\linewidth]{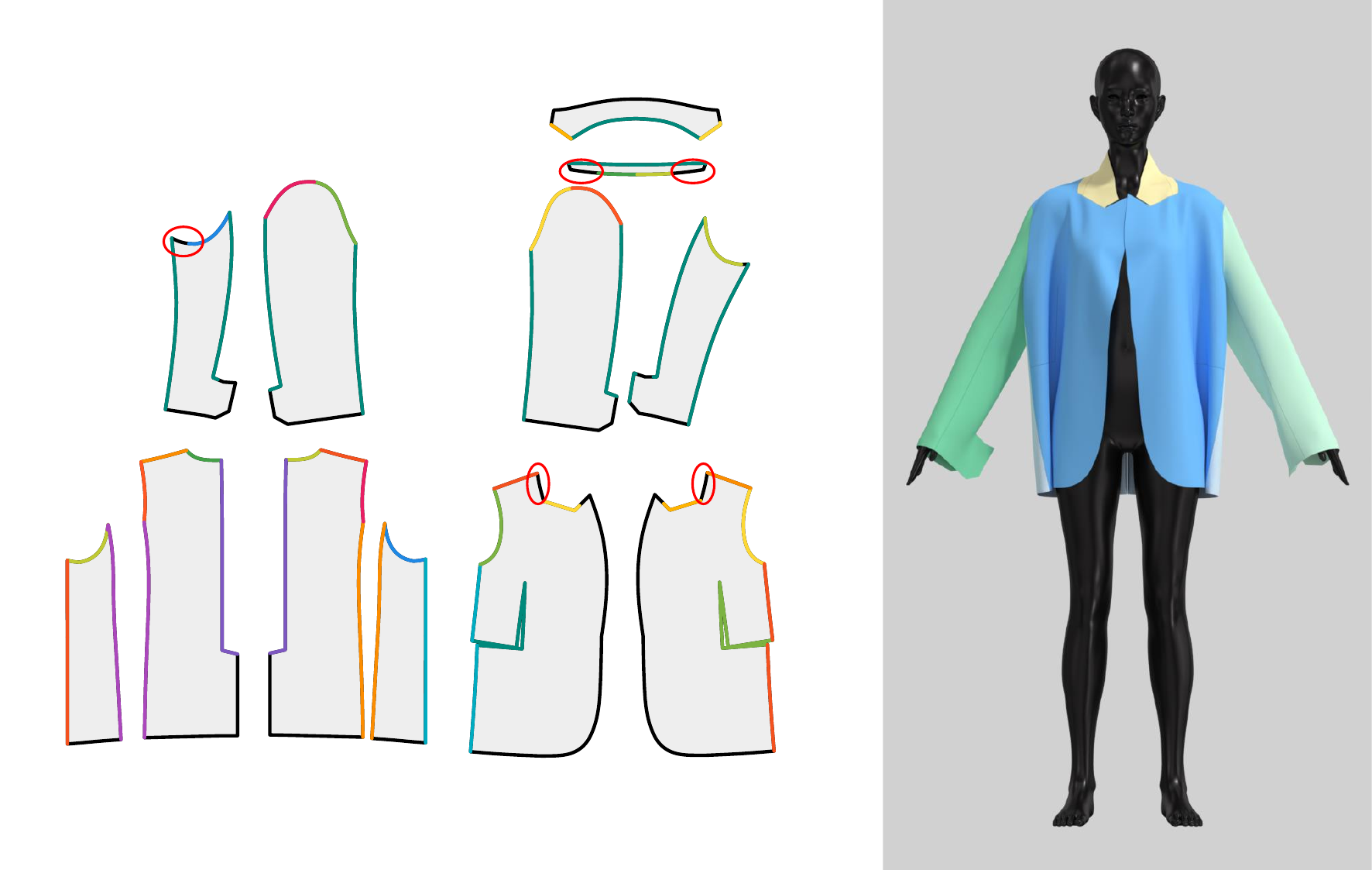}}

    \vspace{-0.1in}

    \caption{Additional examples. For each example, we visualize the input sewing pattern with predicted seam correspondences indicated by segment color matching, and the reconstructed 3D garment with panel semantics represented by panel-wise color coding. Red ellipses indicate minor artifacts.}
    
    \label{fig:last-good-examples}
\end{figure*}

\clearpage

\end{document}